\documentclass{article}

\usepackage[utf8]{inputenc} 
\usepackage[T1]{fontenc}    
\usepackage{url}            
\usepackage{booktabs}       
\usepackage{amsfonts}       
\usepackage{nicefrac}       
\usepackage{microtype}      
\usepackage{lipsum}		
\usepackage{microtype}
\usepackage{amsfonts}
\usepackage{url}
\usepackage{booktabs} 
\usepackage{microtype}
\usepackage{graphicx}
\usepackage{booktabs} 
\usepackage{amsmath}
\usepackage{lipsum}
\usepackage{mathtools}
\usepackage{cuted}
\DeclareMathOperator*{\argmax}{arg\,max}

\usepackage{pdfpages}
\usepackage{amsthm}
\usepackage{amsthm}
\newtheorem{theorem}{Theorem}
\newtheorem{lemma}{Lemma}
\newtheorem{assumption}{Assumption}
\usepackage{subfig}
\usepackage{natbib}
\usepackage{doi}
\usepackage{microtype}
\usepackage{amsfonts}
\usepackage{booktabs} 
\usepackage{microtype}
\usepackage{bbm}
\usepackage{booktabs} 
\usepackage{amsmath}
\usepackage{lipsum}
\usepackage{mathtools}
\usepackage{sidecap}
\usepackage{cuted}
\usepackage{pdfpages}
\usepackage{floatrow}
\usepackage{amsthm}
\usepackage{algorithm}
\usepackage{algorithmic}

\newcommand{\TOPTC}{\operatorname{2-OPT-C}}
\newcommand{\Xtilde}{\widetilde{X}}
\usepackage{subfig}

\usepackage{arxiv}

\title{Two-step lookahead Bayesian optimization with inequality constraints}

\author{%
  Yunxiang Zhang\\
  Cornell University\\
  \texttt{yz2547@cornell.edu} \\
  \And
  Xiangyu Zhang\\
  Cornell University\\
  \texttt{xz556@cornell.edu} \\
  \And
  Peter I. Frazier \\
  Cornell University\\
  \texttt{pf98@cornell.edu} \\
}

\begin{document}

\maketitle

\begin{abstract}
Recent advances in computationally efficient non-myopic Bayesian optimization (BO) improve query efficiency over traditional myopic methods like expected improvement while only modestly increasing computational cost. These advances have been largely limited, however, to unconstrained optimization. For constrained optimization, the few existing non-myopic BO methods require heavy computation. For instance, one existing non-myopic constrained BO method \citep{Willcox} relies on computationally expensive unreliable brute-force derivative-free optimization of a Monte Carlo rollout acquisition function. Methods that use the reparameterization trick for more efficient derivative-based optimization of non-myopic acquisition functions in the unconstrained setting, like sample average approximation and infinitesimal perturbation analysis, do not extend: constraints introduce discontinuities in the sampled acquisition function surface that hinder its optimization. Moreover, we argue here that being non-myopic is even more important in constrained problems because fear of violating constraints pushes myopic methods away from sampling the boundary between feasible and infeasible regions, slowing the discovery of optimal solutions with tight constraints. In this paper, we propose a computationally efficient two-step lookahead constrained Bayesian optimization acquisition function (2-OPT-C) supporting both sequential and batch settings. To enable fast acquisition function optimization, we develop a novel likelihood-ratio-based unbiased estimator of the gradient of the two-step optimal acquisition function that does not use the reparameterization trick. In numerical experiments, 2-OPT-C typically improves query efficiency by 2x or more over previous methods, and in some cases by 10x or more.
\end{abstract}

%

\hypersetup{
pdftitle={A template for the arxiv style},
pdfsubject={q-bio.NC, q-bio.QM},
pdfauthor={David S.~Hippocampus, Elias D.~Striatum},
pdfkeywords={First keyword, Second keyword, More},
}


\section{Introduction}

We consider constrained optimization of a continuous black-box function $f$ under continuous black-box constraints $g_i$,  $\min_{x \in \mathbbm{A}} f(x)$ 
    subject to $g_i(x) \leq 0$, $i = 1,\ldots,I$,
within the compact design space $\mathbbm{A} \subseteq \mathbbm{R}^d$.
We suppose both $f(x)$ and $g_i(x)$ are derivative-free, time-consuming-to-evaluate and also noise-free. Such problems arise, for example, in tuning hyperparameters of machine learning models subject to runtime or fairness constraints and policy optimization in reinforcement learning with safety constraints. For instance, neural networks deployed on mobile phones must be accurate but may also have limited computation available while needing to respond to users in real time, creating a constraint on how long the model takes to predict at test time \citep{sigopt}. Another example, from \citep{fair}, is predicting recidivism risk in the criminal justice system with a fairness constraint ensuring that false positive rates are equal across racial and ethnic groups. Other applications arise in drug discovery \citep{drug} and aircraft design \citep{aircraft}.

Bayesian optimization (BO) has proven successful at solving black-box optimization problems with expensive objectives \citep{BO_ml, BO_tutorial}, including constrained problems of the form above. BO methods for constrained problems include constrained expected improvement (EIC as in \citep{EIC_1998}, rediscovered by \citep{Gardner14}), constrained BO with stepwise uncertainty reduction \citep{stepwise}, predictive entropy search with unknown constraints \citep{pesc}, Alternating
Direction Method of Multipliers Bayesian optimization \citep{ADMMBO}, constrained BO with max-value entropy search \citep{constrained_max_entropy},
and augmented Lagrangian techniques that convert constrained problems into a sequence of unconstrained ones \citep{BO_AL}.
It also includes methods designed specifically for equality and mixed constraints \citep{BO_slack}, for batch observations \citep{NEI} and for problems with high dimensions \citep{scalable}.


All of these existing methods, however, for constrained Bayesian optimization (CBO), are myopic, in the sense that they only consider the immediate improvement in solution quality resulting from a function evaluation and ignore later improvements in solution quality enabled by this evaluation.
(Notable exceptions are \citep{Willcox, multi_info}, discussed below.)
 This greedy behavior may hinder an algorithm's ability to find good solutions efficiently.  While a recent flurry of activity is addressing this issue for unconstrained problems \citep{lookeahead_unconstrained, glasses, practical, binoculars,fb_multi}, and the performance improvements provided by these non-myopic algorithms for unconstrained BO suggest that non-myopic BO is promising for constraints as well, substantial non-myopic development has not reached the constrained setting.

Moreover, we argue in detail below that being non-myopic provides even more value in constrained settings than it does in unconstrained ones. 
To find a global optimum under constraints quickly, an algorithm benefits by efficiently learning the boundary between the feasible region where the constraint is satisfied and the infeasible region where it is not. 
This is facilitated by sampling points likely to be close to this boundary. Myopic methods, however, such as EIC, undervalue sampling such points because they have a substantial probability of being infeasible and because infeasible points do not directly improve solution quality. Non-myopic methods, on the other hand, understand that learning about the boundary's location will provide future benefits, allowing them to value this information more appropriately. Localizing the boundary efficiently is especially important when the global optimum lies on this boundary, as it often does in constrained optimization when the objective (e.g., the quality of a product) is negatively correlated with a constraint (e.g., the cost required to produce it). We illustrate this via a simple example in \S3 and our numerical experiments in \S6, which show several-fold improvement over the state-of-the-art in some problems.

One existing non-myopic constrained BO method \citep{Willcox} first formulates CBO as a dynamic program (DP). However, this DP is intractable. To mitigate the issue, rollout, an approximate DP technique, is used. Nevertheless, this approach requires an extremely large amount of computation to approximate the multi-step lookahead policy well, especially in problems with more than a few dimensions, in part because it relies on computationally expensive derivative-free optimization , and because its acquisition function is computed via Monte Carlo, further increasing the computation required. This limits its applicability.

The constrained multi-information source BO method recently proposed by \citep{multi_info} is also non-myopic. It focuses on the multi-information source setting and assumes that the objective and constraint are evaluated in a decoupled fashion. 
In contrast, our applications of interest often compute the objective and constraints simultaneously. For example, when tuning ML hyperparameters to maximize accuracy subject to a model execution time constraint, the marginal cost of evaluating accuracy is negligible once we evaluate model execution time and incur the training cost. Thus, a method that evaluates the objective and constraints in a decoupled way discards information in such settings.


\textbf{Our Contributions.} We provide a novel non-myopic computationally efficient method for batch CBO. It substantially outperforms myopic CBO methods. In relation to \citep{Willcox} it requires substantially less computation to decide where to sample. Its query efficiency is substantially better in relatively higher-dimensional problems and is at least as good in lower-dimensional ones.

The key to our approach is a new method for optimizing stochastic acquisition functions, leveraging the likelihood ratio method \citep{likelihood}. Standard efficient approaches to optimizing stochastic non-myopic acquisition functions, such as infinitesimal perturbation analysis (IPA) \citep{IPA} or the one-shot method \citep{botorch} (also called sample average approximation or SAA), rely on a sampled acquisition function surface created using the reparameterization trick. In CBO, however, this surface is discontinuous, preventing the efficient use of these methods. Our novel approach is potentially generalizable to other settings where such discontinuities prevent the use of IPA and one-shot optimization.

Our work builds on the unconstrained two-step optimal method \citep{practical}, overcoming substantial computational difficulties created by constraints' inherent discontinuities. These difficulties require abandoning the IPA approach used in  \citep{practical} and instead developing a new likelihood-ratio-based approach.
The likelihood ratio method that we use here to estimate the gradient of the acquisition function relies on a change of measure of the same type used within importance sampling. \citep{practical} coincidentally also uses importance sampling, but in a fundamentally different way: as a variance reduction technique, and not for gradient estimation.

\newcommand{\savespace}[1]{}

Through the rest of this paper, \S2 reviews myopic CBO methods, Gaussian processes, and EIC. \S3 demonstrates why being non-myopic is important in the constrained setting via a simple example. Then we define notation and propose our new lookahead acquisition function $\TOPTC$ for batch CBO. \S4 discusses the discontinuity issue introduced by the reparameterization trick in the constrained setting. \S5 shows how to efficiently optimize $\TOPTC$ via a likelihood ratio gradient estimator. \S6 compares the empirical performance of $\TOPTC$ against other CBO algorithms on a collection of synthetic and real-world problems widely used in the CBO literature. \S7 concludes the paper.

\section{Background}
We briefly review the literature on myopic CBO. Then we summarize standard results needed later on Gaussian processes and the widely used myopic method EIC as well as its batch version.

\subsection{Myopic Constrained Bayesian Optimization}
This work builds on the larger literature on myopic CBO. We review this literature here, giving more details than in \S1. \citep{Gardner14} proposes constrained expected improvement, a constraint-weighted expected improvement acquisition function which multiplies the expected improvement with the probability of feasibility associated with each constraint, rediscovering an approach due to \citep{EIC_1998}. \citep{BOUC} proposes an approach in which the point to sample is found by maximizing EIC and then a decision is made whether to evaluate the objective or the constraint based on the information gain. \citep{stepwise} proposes a stepwise uncertainty reduction method in which the acquisition function aims to maximally decrease our uncertainty on the location of the optimizer with a single evaluation. Predictive entropy search with constraints (PESC) \citep{pesc}, an extension of predictive entropy search \citep{pes} is another information-gain based approach. It chooses the point to evaluate by approximating the expected information gain on the value of the constrained minimizer. \citep{BO_AL} proposes a hybrid approach combining the expected improvement with an augmented Lagrangian framework.
\citep{BO_slack} extends this technique by introducing an alternative slack variable formulation that handles equality and mixed constraints. Integrated expected conditional improvement (IECI) \citep{IEIC} proposes a new acquisition function that integrates a conditional improvement with respect to a reference point over the design space. 
\citep{NEI} develops a quasi-Monte Carlo approximation of expected improvement under batch optimization with noisy observations and noisy constraints. 
\citep{constrained_max_entropy} modifies the mutual information criterion of max-value entropy search \citep{max_entropy} and extends to the constrained setting with the ability to handle both continuous and binary constraints. 
\citep{ADMMBO} leverages the ADMM framework to convert constrained problems into multiple unconstrained subproblems by introducing auxiliary variables for each constraint, then solves the subproblems using standard BO. Similar to TuRBO \citep{turbo}, \citep{scalable} proposes an acquisition function for constrained optimization that scales to high dimensions by maintaining and adjusting trust regions.

\subsection{Gaussian Processes}
BO makes productive use of Gaussian processes (GPs) \citep{GP_book}. We put a GP prior on the objective function $f$, which is specified by a mean function $\mu(\cdot)$ and a kernel function $K(\cdot , \cdot)$.  After observing the data points $D = \{x^{(1)}, x^{(2)}, x^{(3)}, \ldots, x^{(n)}\}$ and their corresponding function values $f(D) := \{f(x^{(1)}), f(x^{(2)}), f(x^{(3)}), \ldots, f(x^{(n)})\}$, the GP prior over $f$ is updated by:
\begin{equation*}
    f(x)| D, f(D) \sim \mathcal{N}\left(\mu(x; D), \sigma^2(x; D)\right),
\end{equation*}
where
\begin{align*}
    &\mu(x; D) = \mu(x) + K(x, D)K(D, D)^{-1}(f(D) - \mu(D)),\\[0.25em]
    &\sigma^2(x; D) = K(x,x) - K(x, D)K(D, D)^{-1}K(D, x).
\end{align*}
$\mu(x; D)$ is the posterior mean and $\sigma(x; D)$ is the posterior standard deviation. $\mu(D)$ is the set of values of the prior mean function evaluated at points in D. $K(x, D), K(D,x)$ are column and row vectors respectively of kernel function values evaluated at $x$ and the points in $D$, and $K(D, D)$ is the kernel matrix defined similarly. 

While we support multiple constraints, we focus on a single constraint $g$ for ease of presentation. We use $\mu^c(\cdot)$ and $K^c(\cdot , \cdot)$ to denote the mean function and the kernel of the independent GP prior on $g$ respectively, and $\mu^c(x; D)$ and $\sigma^c(x; D)$ to denote the mean and standard deviation of the posterior. We refer readers to the supplement for multiple constraints. 

\subsection{Constrained Expected Improvement}
\label{sec:EIC_def}
We briefly review constrained expected improvement (EIC) 
\citep{Gardner14,EIC_1998}, introducing notation used later.
Suppose we have independent GP priors on $f$ and $g$ and observations 
$f(D) := \{f(x) : x \in D\}$
and 
$g(D) := \{g(x) : x \in D\}$
at a collection of datapoints 
$D$.
Let $f^*$ be the best point observed so far subject to our constraints, $f^{*} = \min_{x \in D, g(x) \leq 0} f(x)$. 
The constrained expected improvement at $x$ is:
 \begin{align}\label{eq:1}
    \text{EIC}(x) &= \mathbb{E}\left[\left[f^* - f(x)\right]^{+}\cdot \mathbbm{1}\{g(x) \leq 0 \}\right]  = \mathbb{E}[f^* - f(x)]^{+} \cdot \mathbb{E} [ \mathbbm{1}\{g(x) \leq 0 \}] = \text{EI}(x)\cdot\text{PF}(x),
\end{align}
where $\mathbb{E} [\cdot]$ is the expectation taken with respect to the posterior given $D$, $f(D)$, and $g(D)$. $\text{EI}(x)$ and $\text{PF}(x)$ are the expected improvement and the probability of feasibility at $x$ respectively, and both have analytic forms. Here, we assume the prior on $f$ and $g$ are independent as in \citep{Gardner14}. We refer readers to \citep{Mockus} and \citep{Gardner14} for more details.


We now define EIC in the batch setting, building on the first discussion of which we are aware \citep{NEI}. 
Let $\mathbf{X} = \{x^{(n+1)}, x^{(n+2)}, \ldots, x^{(n+q)}\}$ be the batch of $q$ candidate points that we consider evaluating next. Then the  \textit{constrained multi-points expected improvement} $\text{EIC}(\mathbf{X})$ is,
\begin{equation}\label{eq:2}
    \text{EIC}(\mathbf{X}) = \mathbb{E} \left[\max_{x \in \mathbf{X}} \hspace{0.5em}(f^* - f(x))^+ \cdot \mathbbm{1}\{g(x) \leq 0 \} \right].
\end{equation}
The details of the derivation of $\text{EIC}(\mathbf{X})$ are provided in the supplement.

\section{Constrained Two-step Acquisition Function}
\label{sec:two_step_def}
In this section, we first show that why being non-myopic is important in CBO. Then we formally define the novel acquisition function $\TOPTC$ that is at the heart of our method. 

\subsection{The Importance of Being Non-Myopic in CBO}

\begin{figure*}[ht]
\begin{center}
\includegraphics[width=.16\textwidth]{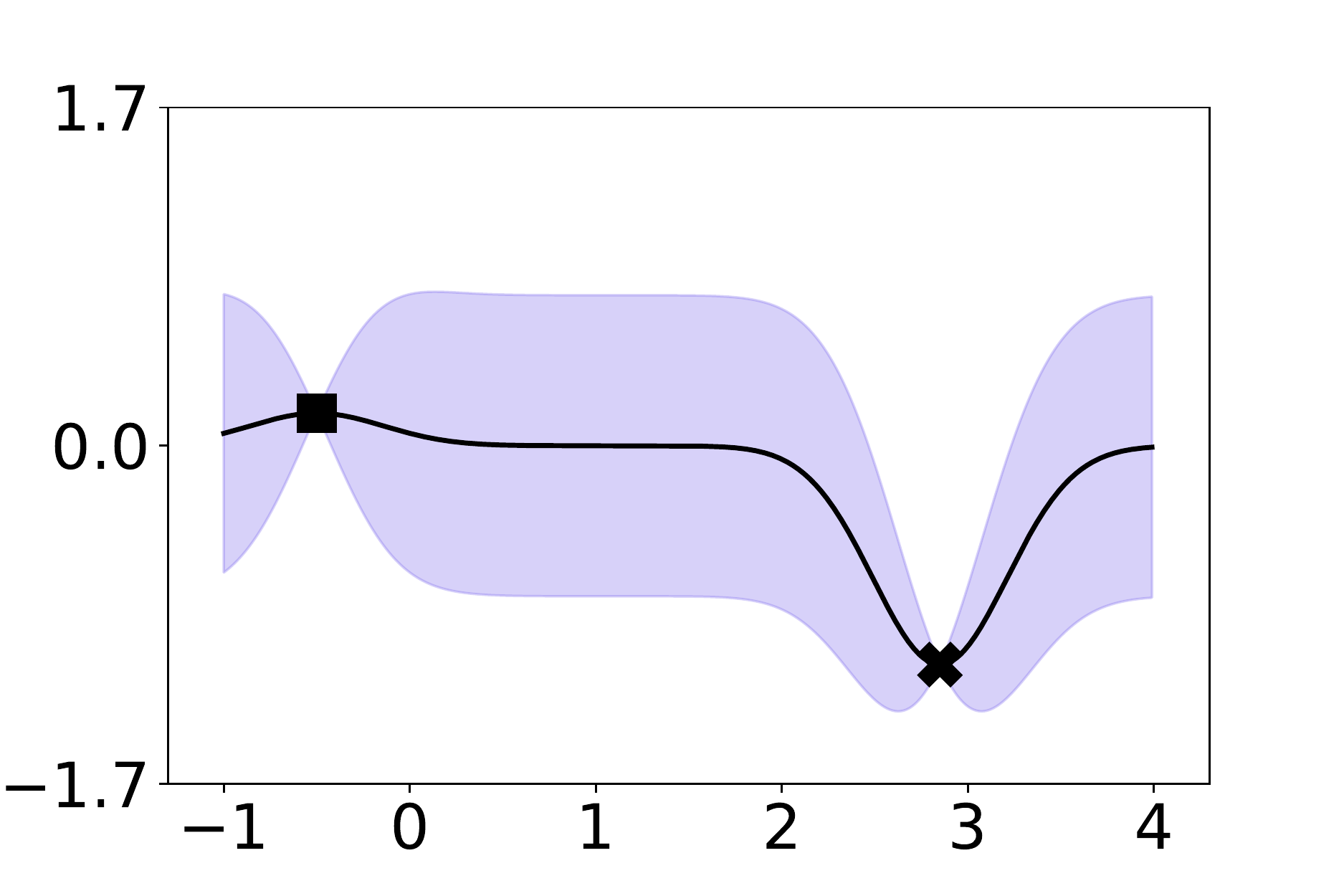}
\includegraphics[width=.16\textwidth]{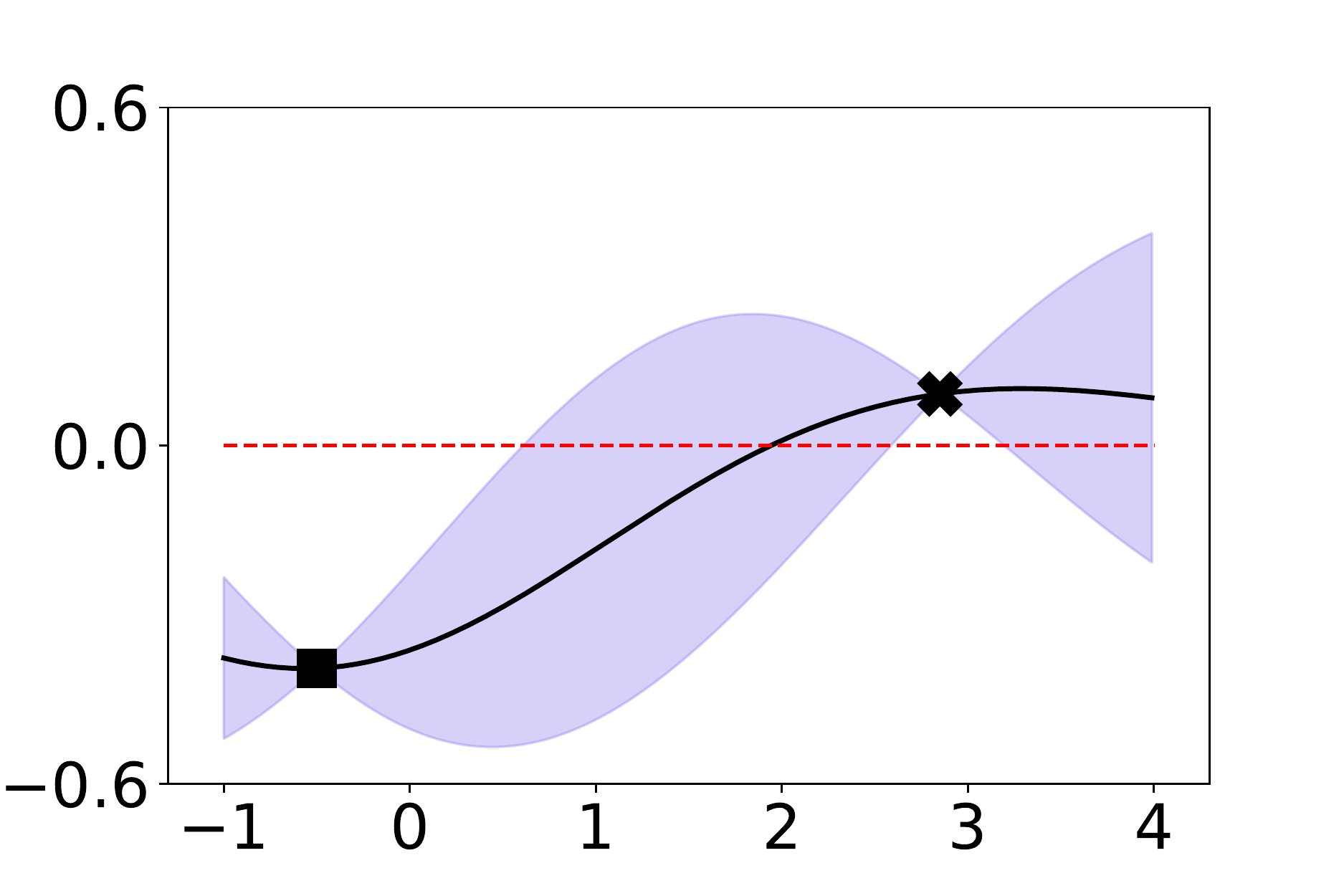}
\includegraphics[width=.16\textwidth]{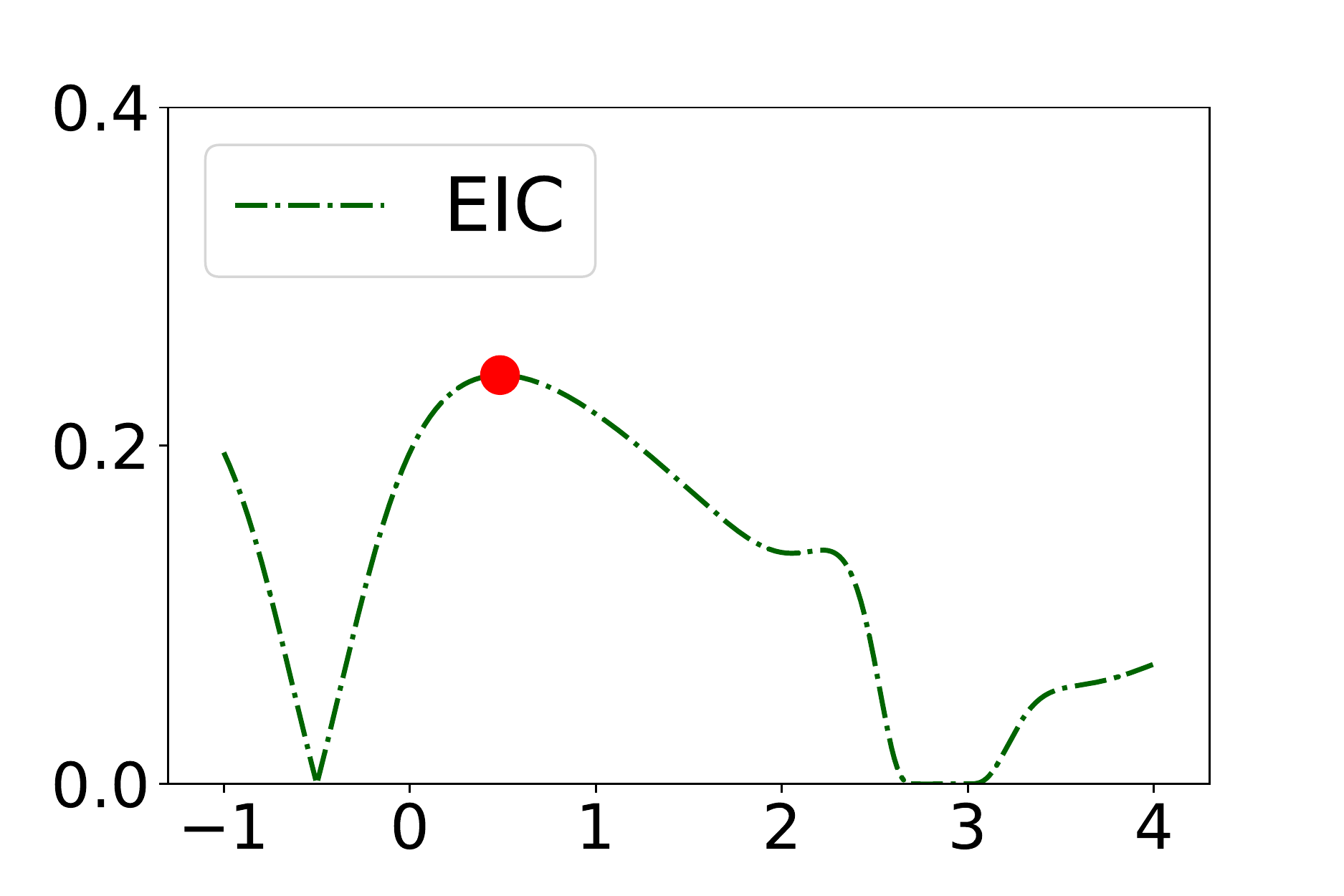}
\includegraphics[width=.16\textwidth]{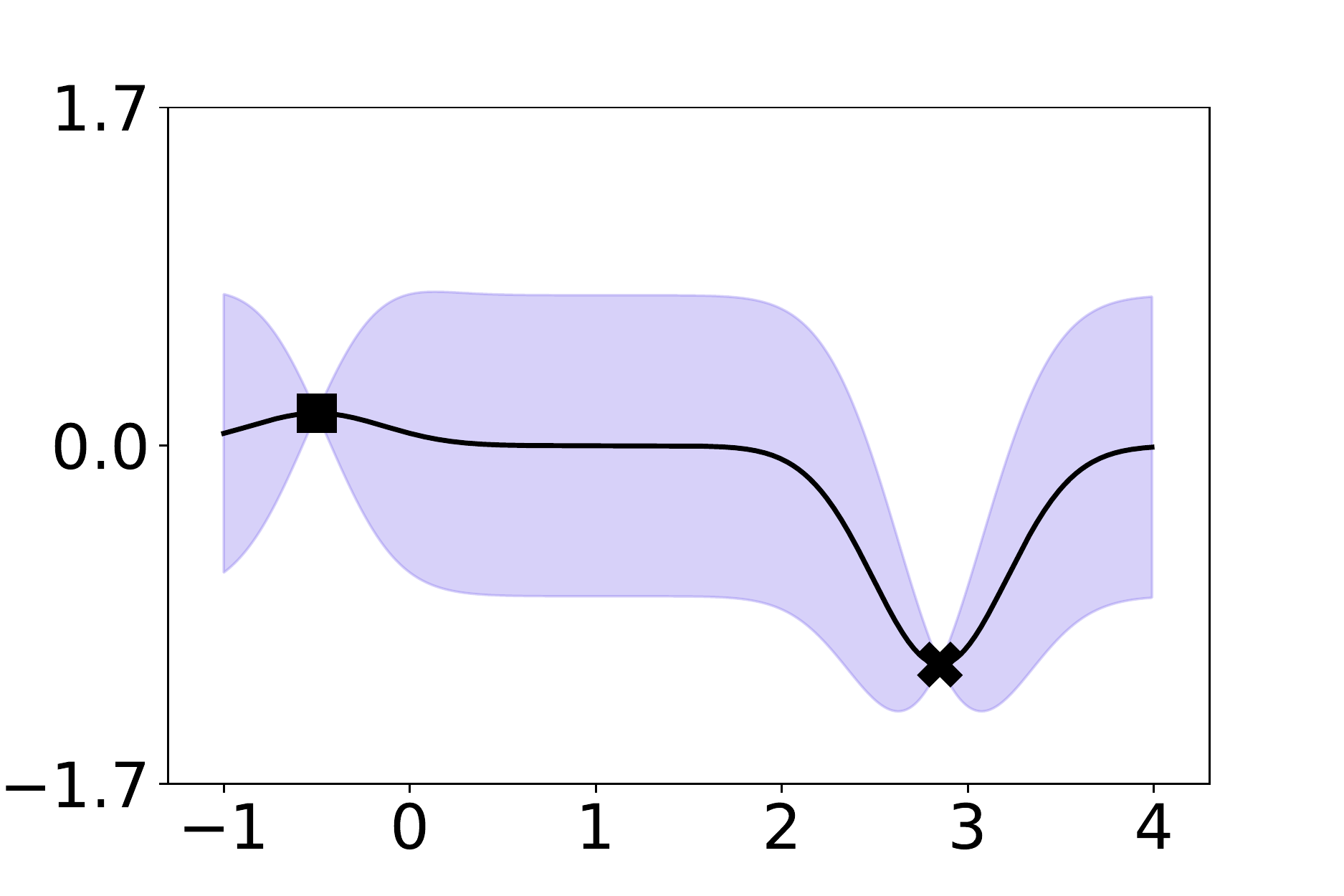}
\includegraphics[width=.16\textwidth]{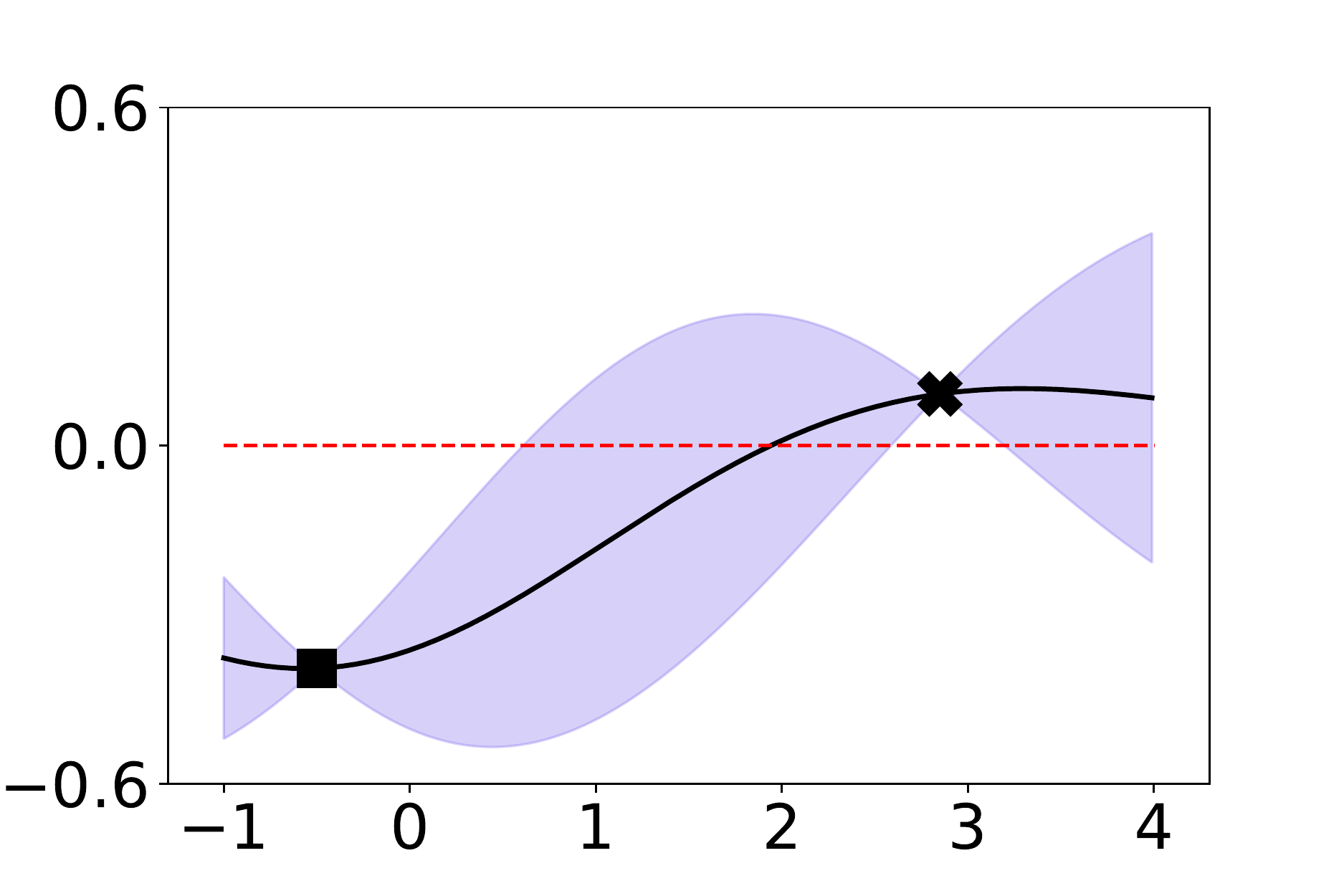}
\includegraphics[width=.16\textwidth]{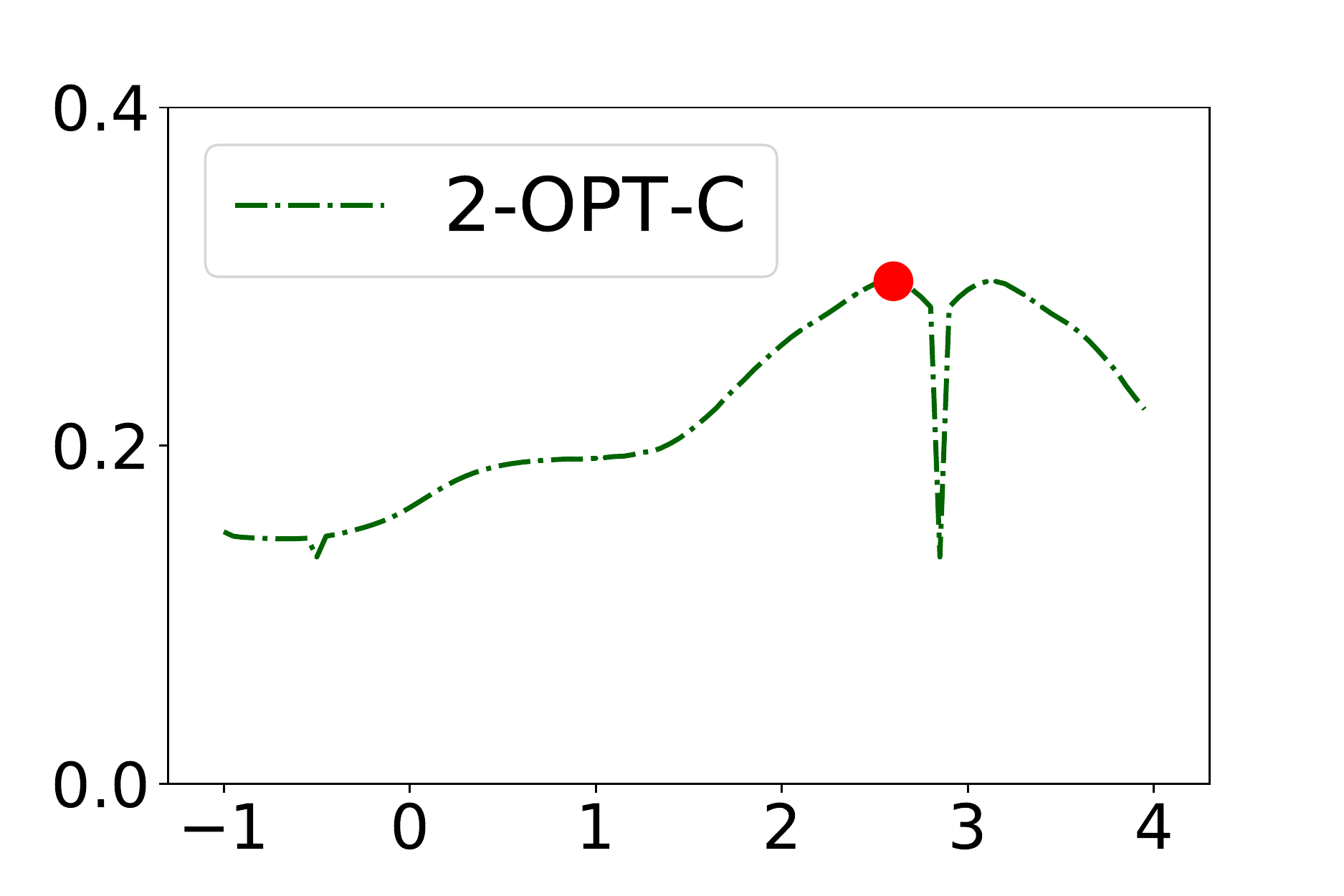}
\includegraphics[width=.16\textwidth]{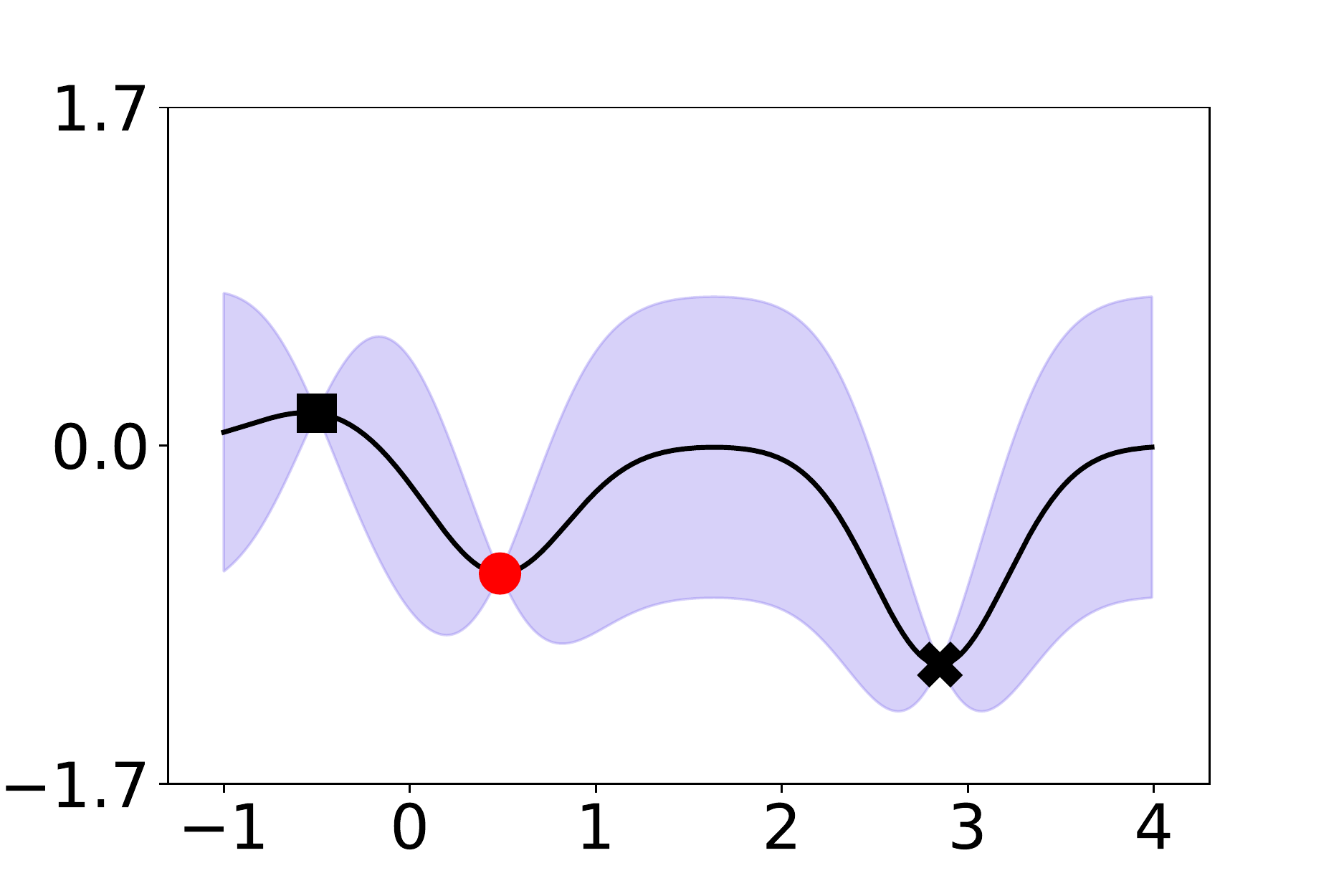}
\includegraphics[width=.16\textwidth]{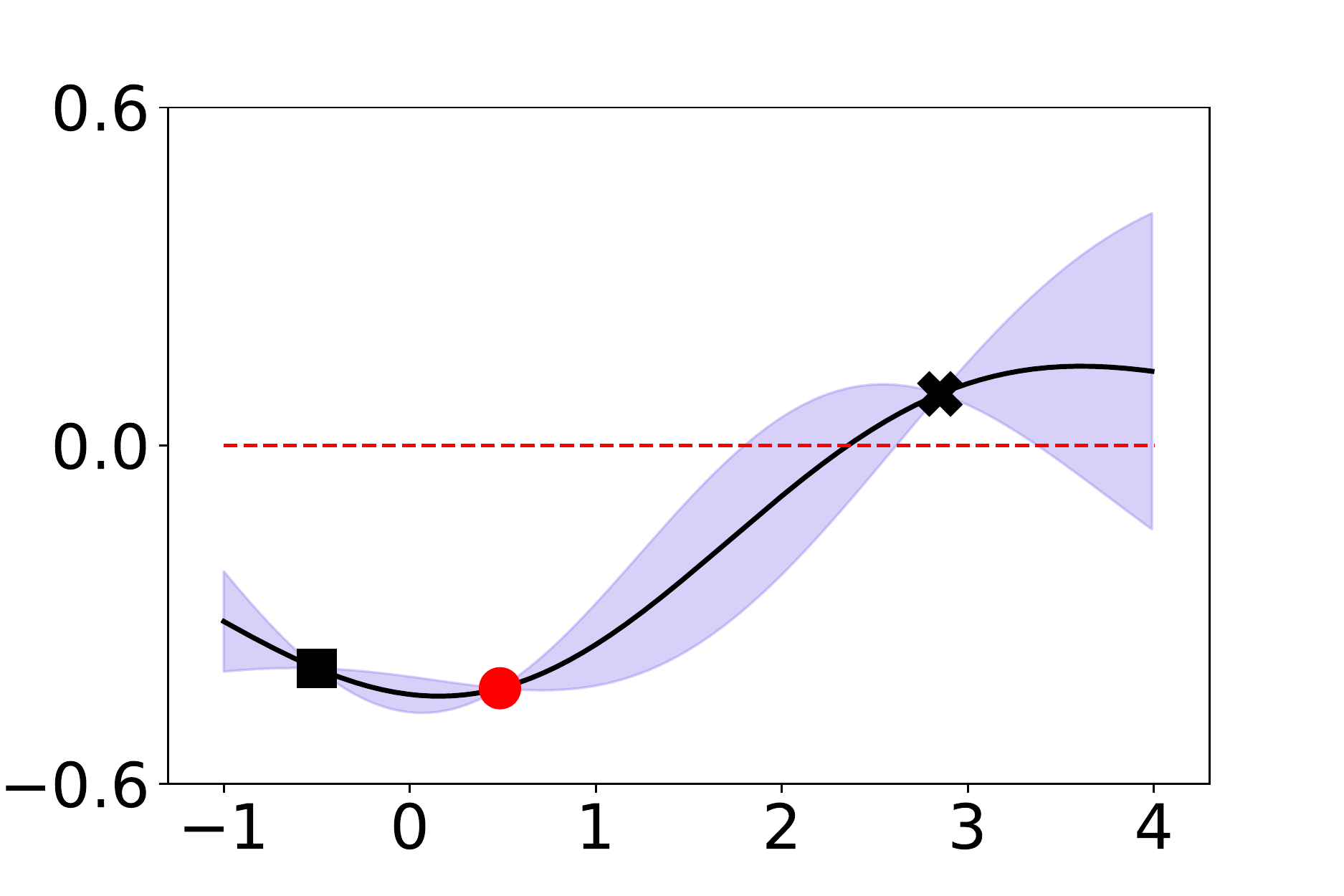}
\includegraphics[width=.16\textwidth]{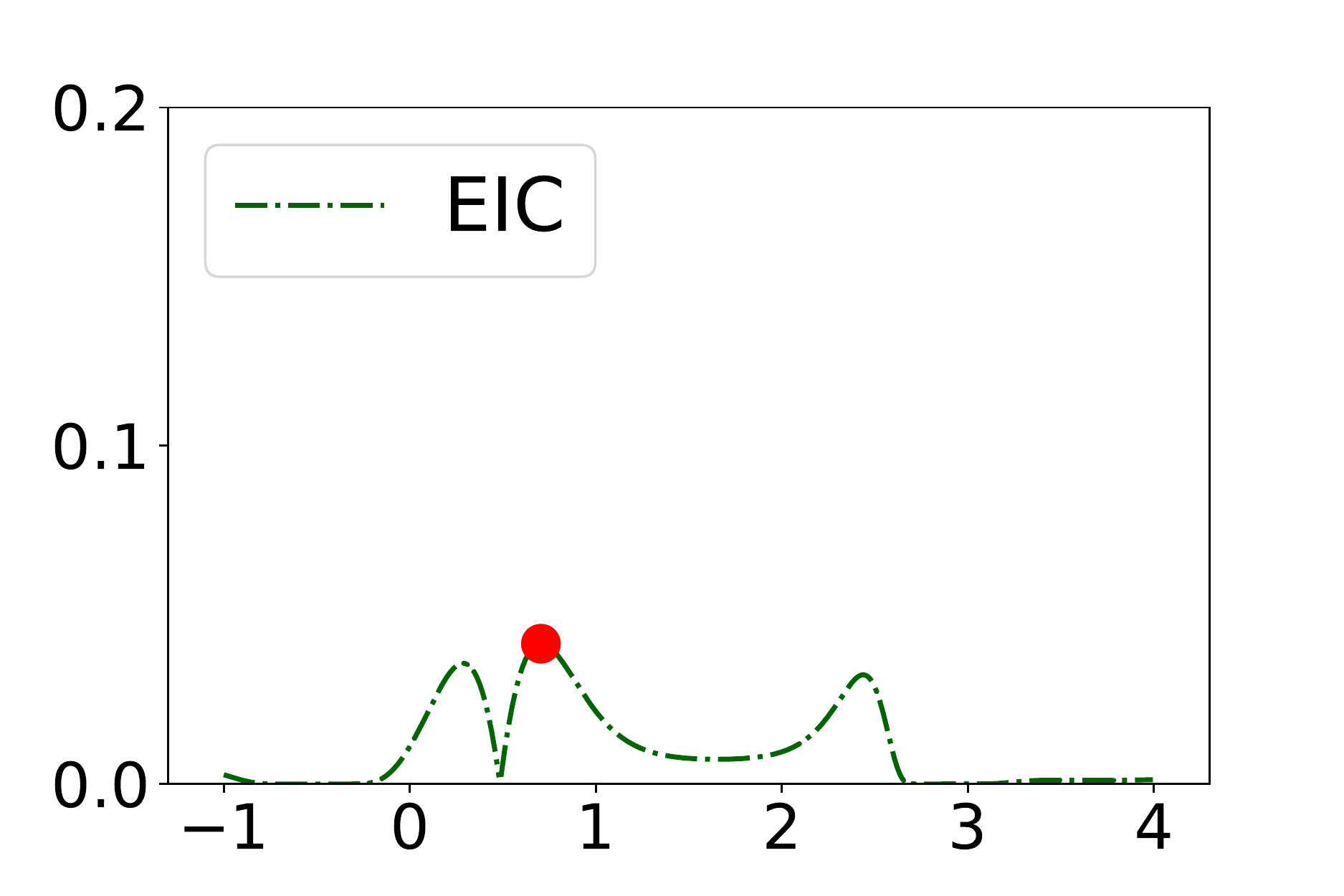}
\includegraphics[width=.16\textwidth]{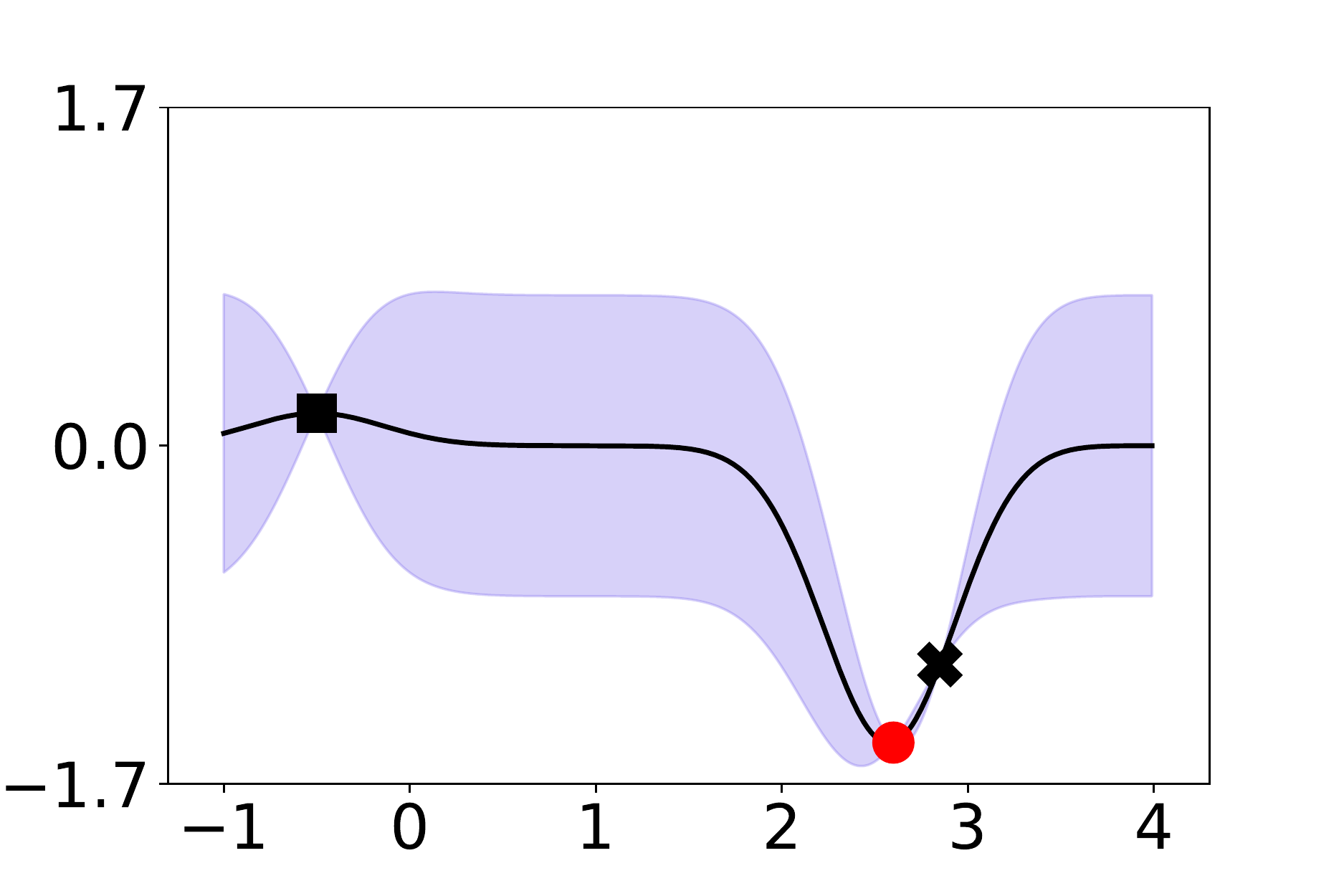}
\includegraphics[width=.16\textwidth]{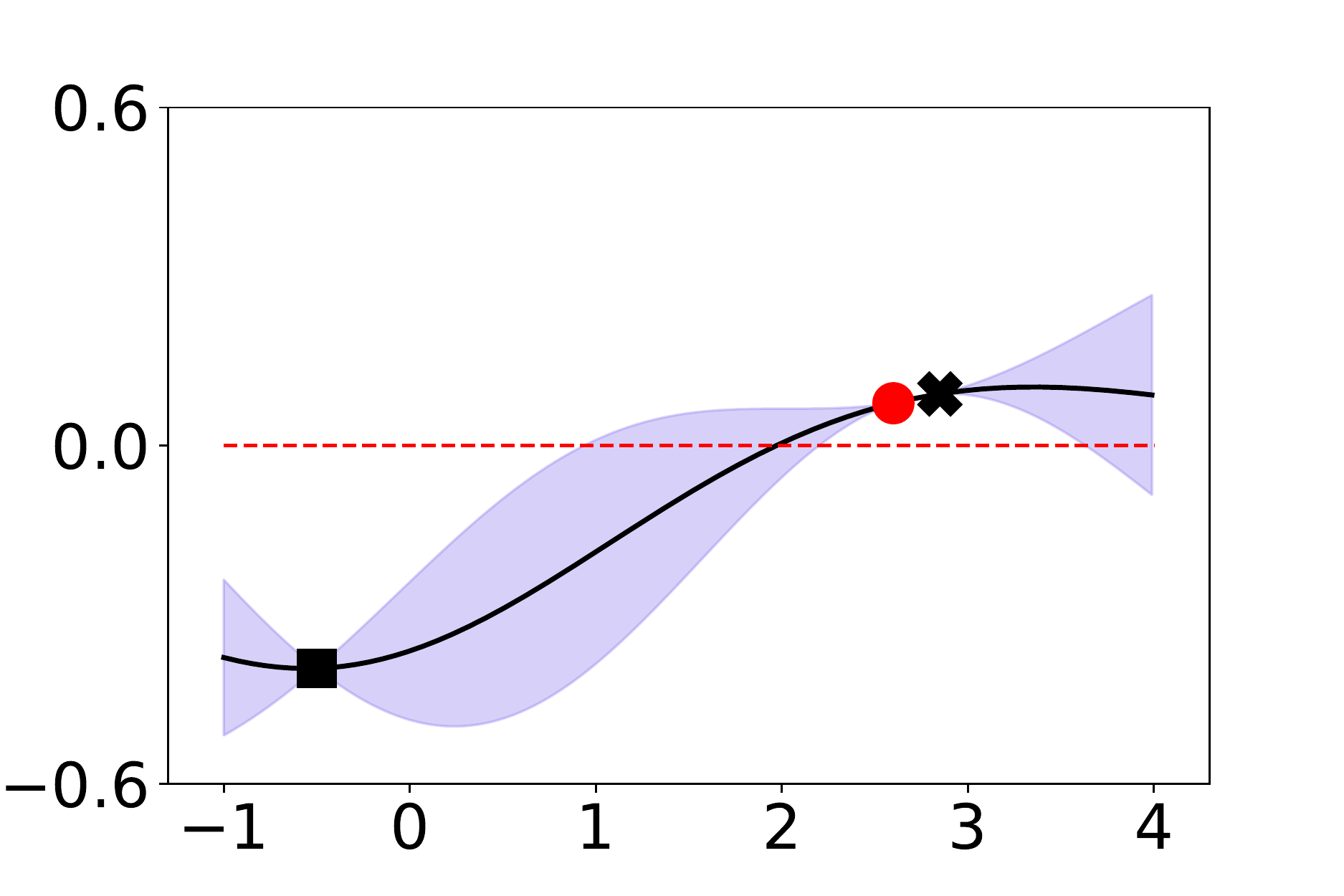}
\includegraphics[width=.16\textwidth]{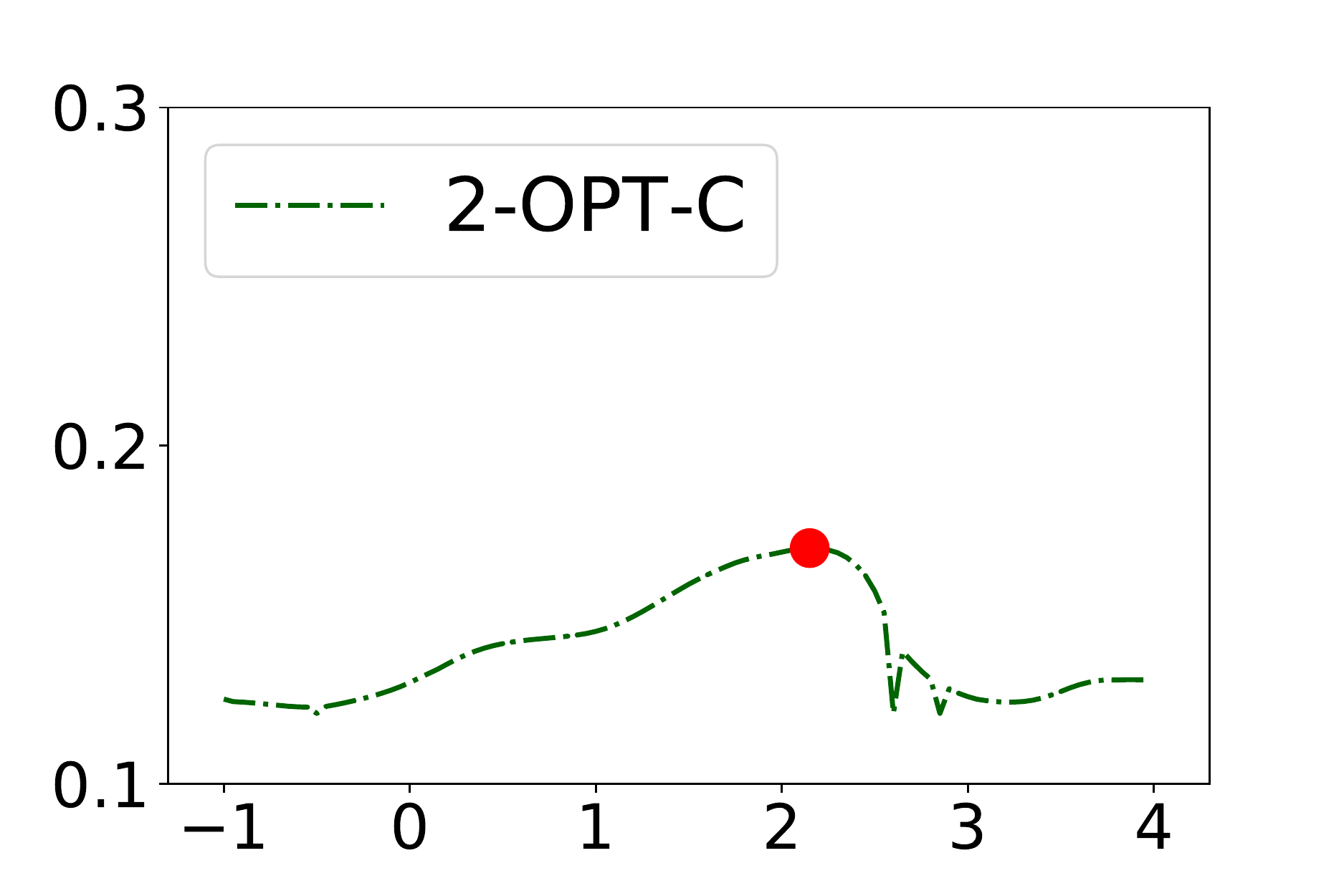}
\includegraphics[width=.16\textwidth]{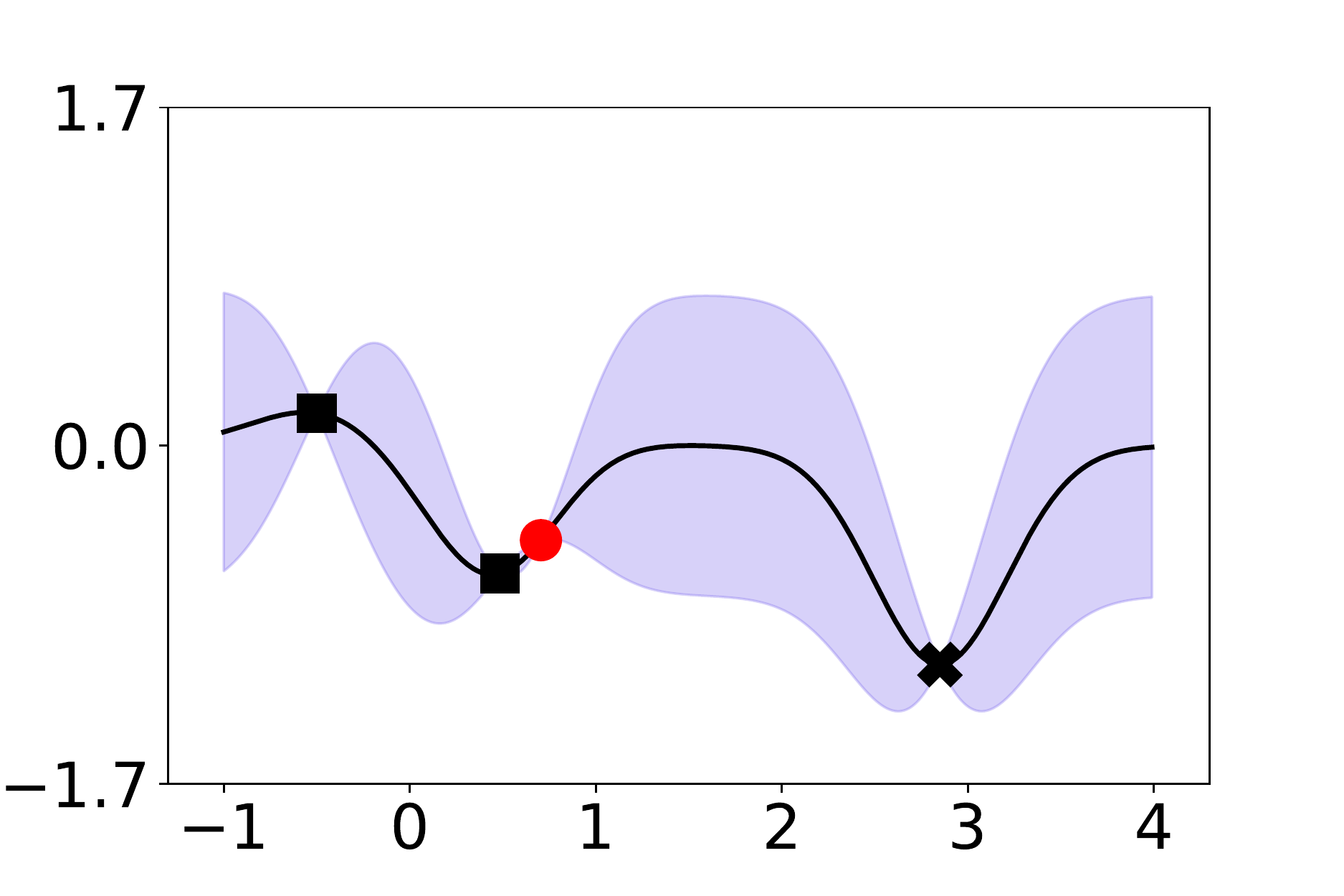}
\includegraphics[width=.16\textwidth]{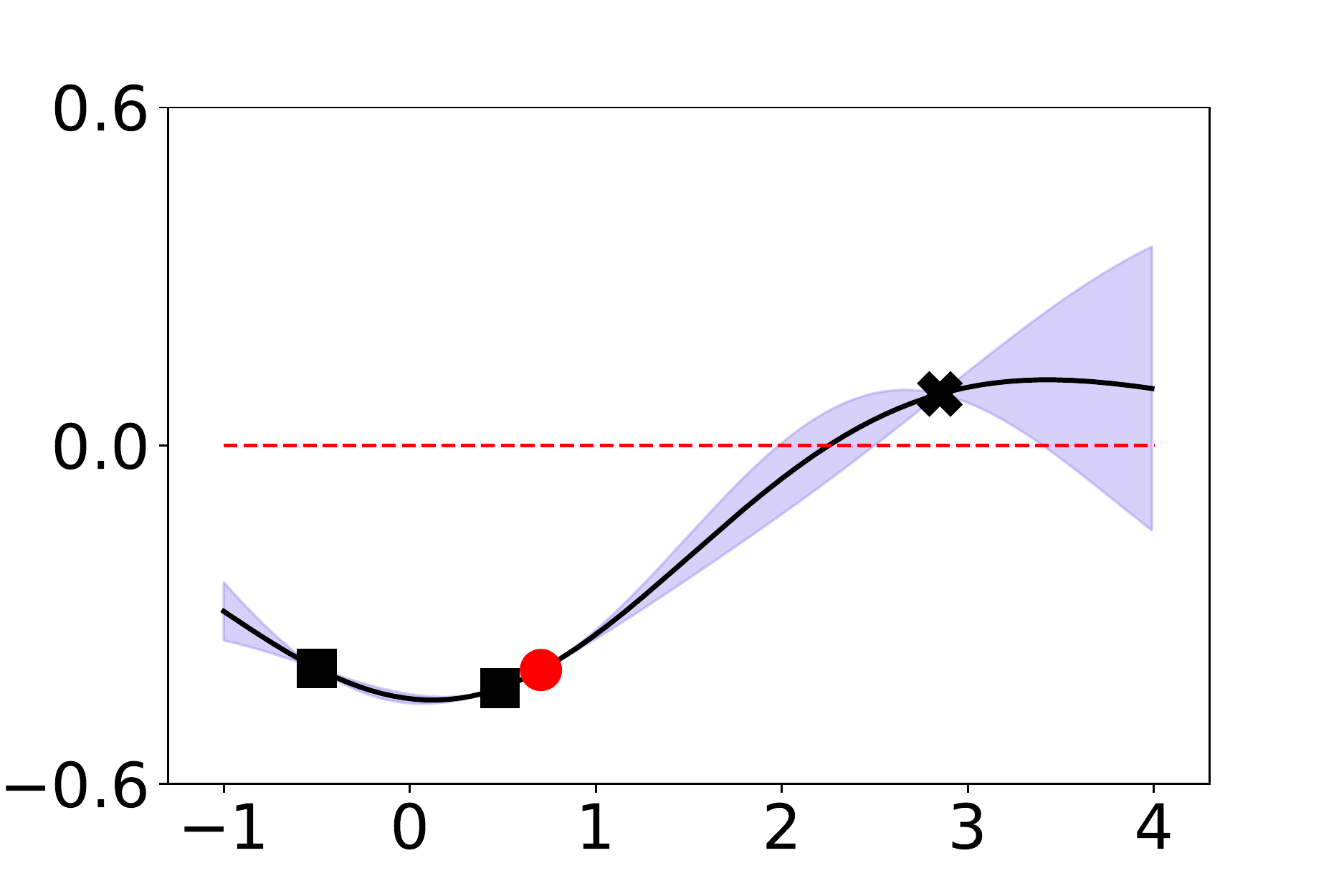}
\includegraphics[width=.16\textwidth]{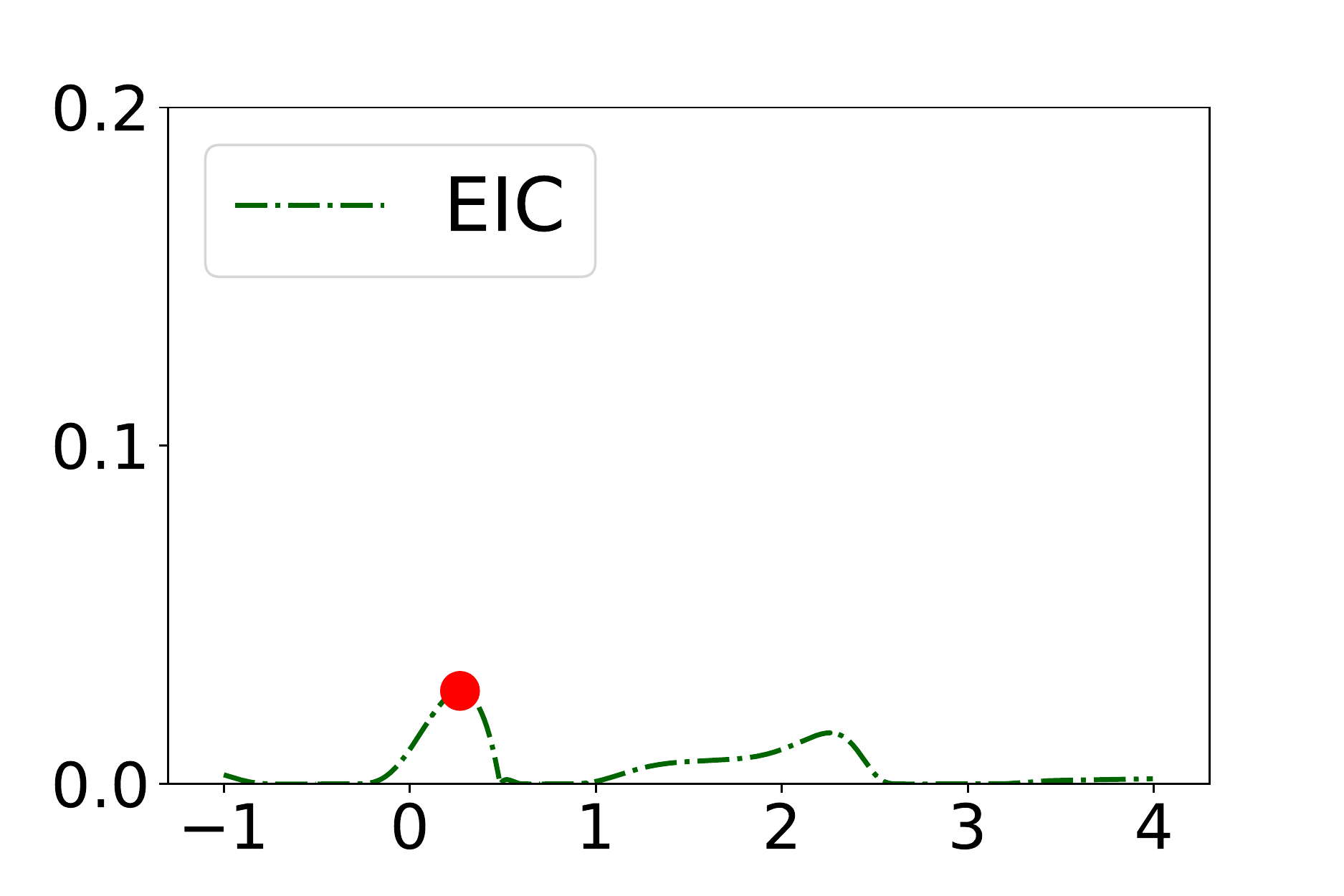}
\includegraphics[width=.16\textwidth]{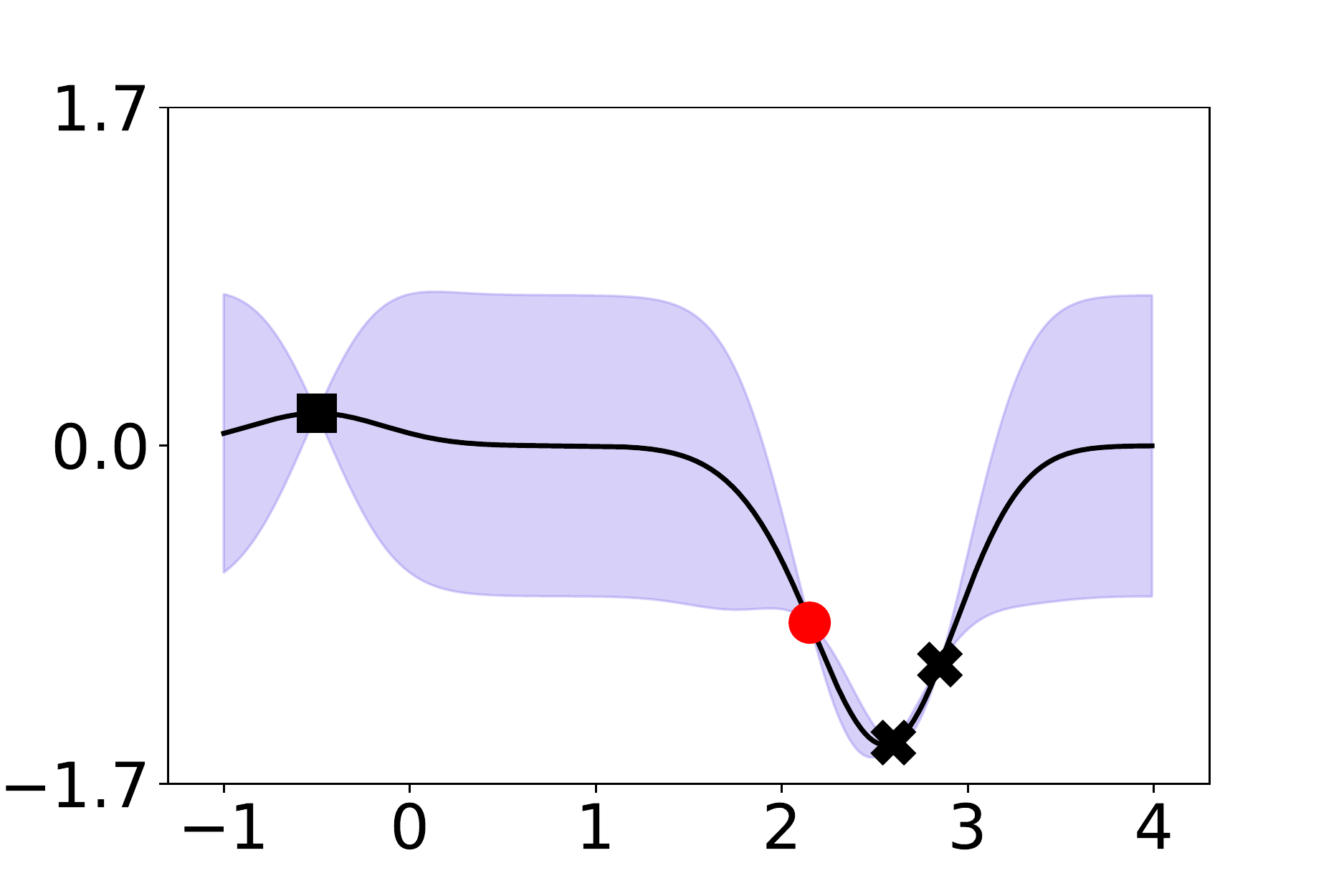}
\includegraphics[width=.16\textwidth]{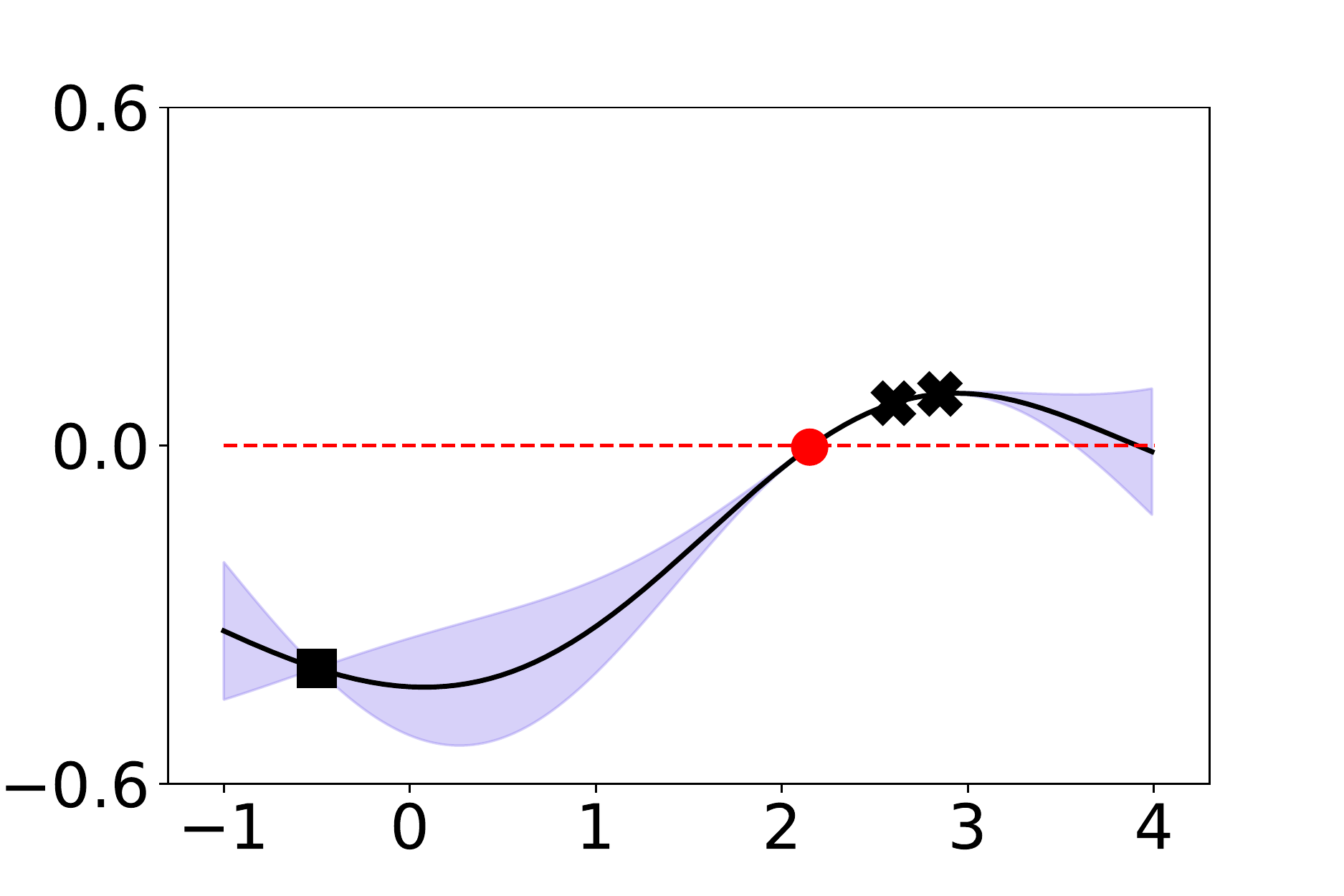}
\includegraphics[width=.16\textwidth]{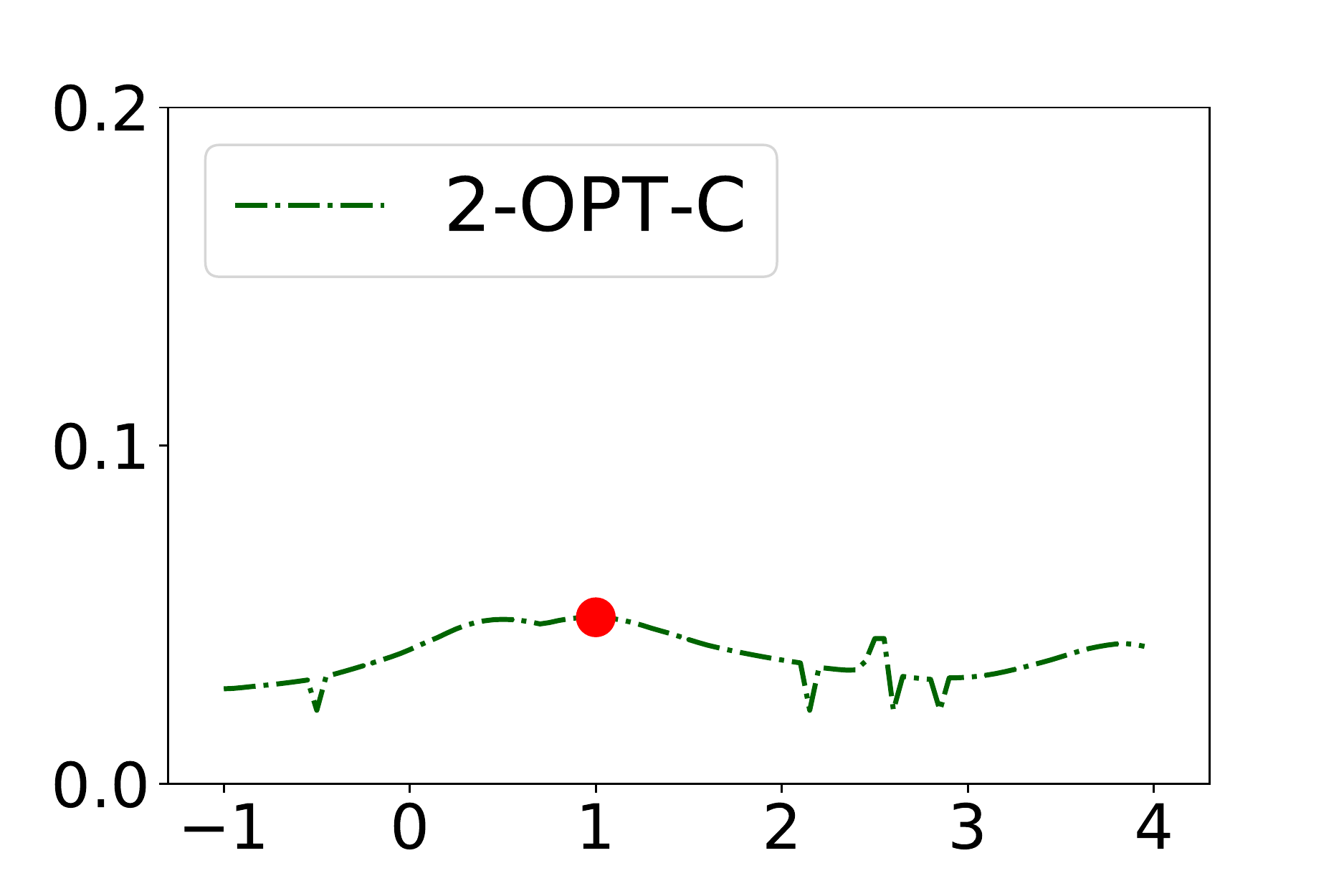}
\caption{EIC and $\TOPTC$ optimizing a 1-d synthetic function. The first three graphs of each row are the posterior on $f$, the posterior on $g$, and the acquisition function under EIC, and the last three graphs are the corresponding quantities under $\TOPTC$.  Each row shows one iteration, proceeding from the first iteration at the top to the last iteration at the bottom. 
In the plots of the posterior on $f$ and $g$, ``x'' denotes infeasible points and a square denotes feasible points. The constraint threshold (0) is plotted as a red line.  The red dot is the point newly sampled in the current iteration. $\TOPTC$ explores potentially infeasible points more aggressively and discovers a better feasible point.
\label{two_vs_eic}
}
\end{center}
\end{figure*}

Before presenting out non-myopic acquisition function, $\TOPTC$, we give an intuition for why being non-myopic is even more important in constrained problems than it is in unconstrained ones. Suppose the constraint function value of the (unobserved) best feasible point is equal or close to the constraint threshold. To find this point with as few function evaluations as possible, the algorithms need to efficiently learn the feasible region's boundaries. This happens most quickly when evaluating points close to this boundary. Since the boundary is uncertain, this requires an algorithm to be willing to evaluate points that may be infeasible. However, the value of such points to myopic CBO algorithms like EIC is significantly reduced because the probability of infeasibility is high --- typically near 50\%, lowering the $\text{PF}(x)$ term in (\ref{eq:1}). Thus, myopic CBO algorithms may insufficiently explore the boundary, instead sticking well inside the feasible region.

Figure~\ref{two_vs_eic} illustrates this with an example. On the first iteration, EIC evaluates a point in a region around an existing feasible point that is likely feasible, getting a small one-step improvement but little information on $f$ and $g$ for future evaluations. However, $\TOPTC$ (defined below) explores more aggressively. The point it evaluates is not feasible but provides a significant amount of information about the boundary between the feasible and infeasible region. As a result, after progressing two steps, $\TOPTC$ locates a new feasible point ($g(x) = -0.02$) with a much lower objective function value while EIC remains stuck near the previous feasible point.

\subsection{Definition of \texorpdfstring{$\TOPTC$}{Lg}}
We define our two-step optimal acquisition function for CBO, called $\TOPTC$, for both sequential and batch settings, and then discuss its computational optimization in \S\ref{sec:computation}. As in other CBO methods, we sequentially evaluate f and g next at the point that maximizes our acquisition function. This point is optimal to measure now assuming we will be able to measure one additional point before being awarded the maximum value found over the two new measurements and in our previously collected data. In other words, $\TOPTC$ looks two steps ahead instead of the one-step ahead considered by myopic methods. 
For clarity, we assume a single constraint and treat multiple constraints in the supplement.

In the batch setting, we find the batch of $q$ points to measure next assuming we will then measure one additional point before being awarded the best value found. We assume only one additional point rather than a full additional batch for computational tractability. We argue that even the single additional point allows $\TOPTC$ to think non-myopically and to value information about the feasible region that will only be acted on later.


We first define notation. We use subscripts $\{0, 1, 2\}$ to denote stages. Stage $0$ is the current stage when we will evaluate a batch of points. In stage $1$, we will evaluate one additional point based on the results of stage $0$ and then in stage 2 will be judged based on the best of these points evaluated.
We put independent GP priors on $f$ and $g$.
Let $D$ be the collection of data points that we have observed so far. 
Then given the information in $D$ and their corresponding objective $f(D)$ and constraint $g(D)$ values, the posterior distribution on the objective $f$ is a Gaussian process with a mean function $\mu_0$ and kernel $K_0$. Similarly, the posterior distribution on the constraint $g$ is also a Gaussian process with a mean function $\mu_0^c$ and kernel $K_0^c$.
$\mathbb{E}_0$ denotes taking the expectation with respect to the posterior distributions given $D$. 

Let $f^{*}_0$ be the best evaluated point satisfying the constraint so far, i.e. $f^{*}_0 = \min_{x \in D, g(x) \leq 0} f(x)$. Let $X_1$ be the set of $q$ candidate points that we consider evaluating at the first stage and let $Y_f = \{ f(x): x\in X_1\}$ and $Y_g = \{ g(x): x\in X_1\}$ be the sets of corresponding objective function values and constraint values respectively. 
We let $\mu_1$, $\mu_1^c$ and $K_1$, $K_1^c$ denote the mean function and kernel for the posterior distributions of $f$ and $g$ respectively given $D$ and $X_1$.
Let $\sigma_1(x) = \sqrt{K_1(x,x)}$
and $\sigma_1^c(x) = \sqrt{K_1^c(x,x)}$.
Let $\mathbb{E}_1$ indicate the expectation with respect to the corresponding Gaussian processes given both $D$ and $X_1$. 
Finally, we use $x_2$ to denote a single point to be evaluated in the second stage, based on the results of the first.

Let $f^{*}_1$ and $f^*_2$ be the best evaluated feasible point by the end of the first and second stage respectively, 
$f_1^* = \min\{f_0^*, \min_{x \in X_1, g(x) \le 0} f(x)\}$ and
$f_2^* = \min\{f_1^*, f(x_2)\}$ if $g(x_2) \le 0$ and $f_2^* = f_1^*$ if not.



Following the principle of dynamic programming, our goal is to choose $X_1$ to minimize the overall expected objective $\mathbb{E}_1(f^{*}_2)$
. This is done under the assumption that $x_2$ will be chosen optimally leveraging observations of $X_1$.
Equivalently, our goal is to maximize 
\begin{equation*}
    \mathbb{E}_0\left[\max_{x_2} \left[f_0^* - f_2^*\right] \right] = {\mathbb{E}}_0\left[f_0^* - f_1^* + \max_{x_2} {\mathbb{E}}_1[f_1^* - f_2^*]\right],
\end{equation*}
over $X_1$ chosen in the first stage. $\mathbb{E}_1[f_1^* - f_2^*]$ depends implicitly on the information obtained from $X_1$, which is $Y_f$ and $Y_g$, and is included in the posterior over which $\mathbb{E}_1$ is taken.

Thus, we define the constrained two-step acquisition function:
\begin{align*}
    \TOPTC(X_1) 
:= {\mathbb{E}_0}\left[f_0^* - f_1^* + \max_{x_2 \in A(\delta)}{\mathbb{E}_1}\left[ f_1^* - f_2^* \right]\right],
\end{align*}
where $A(\delta)$ is a compact subset of $A$ consisting of points at least $\delta$ away from sampled points in $D \cup X_1$. With $\delta = 0$, $A = A(\delta)$. We introduce the parameter $\delta \ge 0$ purely to overcome a technical hurdle in our theoretical analysis: that the standard deviation of the posterior distribution is not smooth at sampled points. 
We believe $\delta$ can be set to $0$ in practice. Indeed, the theoretical analysis (Theorem \ref{th:1}) allows setting $\delta$ at any arbitrary small positive value.

For use in \S\ref{sec:computation}, we derive a more directly computable expression for $\TOPTC(X_1)$.
We rewrite
\begin{equation*}
    \TOPTC(X_1) = {\mathbb{E}_0}\left[\max_{x_2 \in A(\delta)}{ [f_0^* - f_1^* + \mathbb{E}_1}\left[ f_1^* - f_2^* \right]]\right] = {\mathbb{E}_0}\left[\max_{x_2 \in A(\delta)}\alpha (X_1, x_2, Y) \right],
\end{equation*}
where $Y = (Y_f, Y_g) \sim p(y;X_1)$ and 
$p(y;X_1)$ is the distribution of $Y$ given $f(D)$ and $g(D)$, specified explicitly as
\begin{equation}\label{eq:p}
p(y; X_1) =  \mathcal{N}\Bigg{(}
\begin{bmatrix}
      \mu_0(X_1)\\[0.5em]
     \mu_0^c(X_1) 
   \end{bmatrix}, 
   \begin{bmatrix}
     K_0(X_1,X_1) & 0  \\[0.5em]
     0 & K_0^c(X_1,X_1)
   \end{bmatrix}
   \Bigg{)}.
 \end{equation}

Then, $\alpha (X_1, x_2, Y)$ can be written in closed form:
\begin{align*}
    \alpha (X_1, x_2, Y) 
    = f_0^* - f_1^*  + &\text{EI}(f_1^{*} - \mu_1(x_2), \sigma_1(x_2)^2) \cdot \text{PF}(\mu_1^c(x_2), (\sigma_1^c(x_2))^2),
\end{align*}
where $\text{EI}(m, v) =m \Phi(m / \sqrt{v})+\sqrt{v} \varphi(m / \sqrt{v})$ and $\text{PF}(m^c, v^c) = \Phi\left(-m^c / \sqrt{v^c}\right)$. 

\section{Discontinuities Hinder Optimization via the Reparameterization Trick}\label{sec:discontinuity}
Using $\TOPTC$ for CBO requires us to evaluate and optimize the $\TOPTC$ acquisition function. This acquisition function cannot be evaluated exactly, and must instead be evaluated via Monte Carlo. 
Derivative-free optimization of such Monte Carlo acquisition functions, e.g., using CMA-ES, is typically computationally expensive and requires a substantial number of acquisition function simulation replications to optimize well.

In response to this challenge, there are two widely-used special-purpose methods for optimizing Monte Carlo acquisition functions: infinitesimal perturbation analysis (IPA), and sample average approximation (SAA).
IPA efficiently estimates the gradient of the acquisition function and then uses this gradient estimate within multistart stochastic gradient ascent.
SAA uses Monte Carlo samples to create a deterministic and differentiable approximation to the acquisition function, which it then optimizes using derivative-based deterministic optimization.
The first papers known by the authors using IPA and SAA for optimizing BO acquisition functions are \citep{qEI} (which was available earlier on arxiv \citep{qEI-old}) and \citep{botorch} respectively.

As we explain in detail below,  both IPA and SAA rely on a Monte Carlo approximation to the acquisition function simulated using the reparameterization trick. In particular, they rely on this approximation being smooth in the input to the acquisition function.
However, in {\it constrained} BO and as we argued in the main paper, this approximation is not smooth and indeed has discontinuities.
This prevents the efficient use of IPA and SAA for optimizing $\TOPTC$ and necessitates the development of our new likelihood ratio approach.

In this section, we give background on IPA and SAA and then describe in detail these discontinuities and why they prevent the use of IPA and SAA for optimizing $\TOPTC$.

\subsection{Background on IPA and SAA}

For our discussion of IPA and SAA we define generic notation. Let $V(x,\theta)$ be a function of two variables, one that will be a control vector $x$ and the other that will be a random input $\theta$. 
Critically, for both IPA and SAA, the distribution of $\theta$ does not depend on $x$. This contrasts with the likelihood ratio method described in \S \ref{sec:computation}, in which the distribution of $\theta$ {\it can} depend on $x$.

Here, we will discuss the generic use of IPA and SAA for solving $\max_x E[V(x,\theta)]$ where the expectation is over $\theta$. 
In the next section, where we focus on the difficulties arising when optimizing $\TOPTC$ using IPA and SAA, $x$ will be replaced by the batch of points that we consider evaluating in the first stage and $\theta$ will be a vector of two standard normal random variables used in the reparameterization trick.
 
\paragraph{IPA}
We first review the concept of IPA, which is a method for estimating the gradient $\nabla_{x} E[V(x,\theta)]$. Once this gradient is estimated, it can be used within multistart stochastic gradient ascent. 

IPA estimates this gradient as follows. 
Under some regularity conditions \citep{l1990unified}, the gradient operator and the expectation can be swapped, 
\begin{equation}
\nabla_{x} E[V(x,\theta)] = E[\nabla_{x} V(x,\theta)].
\label{eq:IPA}
\end{equation}
 
One can estimate the the right-hand side using Monte Carlo.
This is accomplished by 
generating i.i.d. samples $\theta_i$ from the distribution of  $\theta$ and then using the estimate 
$\frac{1}{N}\sum_{i = 1}^{N} \nabla_x V(x,\theta_i)$
This is an unbiased estimator of the right-hand side of \eqref{eq:IPA}.
When the equality in \eqref{eq:IPA} holds, this is also an unbiased estimator of $\nabla_x \mathbb{E}[V(x,\theta)]$.

\newcommand{\dx}{\frac{\mathrm{d}}{\mathrm{d} x}}

Unfortunately, however, \eqref{eq:IPA} does not always hold. To illustrate, suppose $\theta$ is a standard normal random variable, $\mu(x)$ and $\sigma(x)$ are two continuous functions, 
and $V(x,\theta)=1\{\mu(x) + \sigma(x) \theta \le 0\}$. 
Observe that $V(x,\theta)$ can be  discontinuous in $x$ for any given value of $\theta$.

Then the left-hand side of \eqref{eq:IPA} is 
$\dx E[V(x,\theta)] 
= \dx \Phi(-\mu(x) / \sigma(x))$
where $\Phi$ is the standard normal cumulative distribution function.
The right-hand side of \eqref{eq:IPA} is $E[\dx V(x,\theta)] = 0$, 
since $\dx V(x,\theta)$ is 0 at all $\theta$ except where $\theta = -\mu(x) / \sigma(x)$, and this $\theta$ occurs with probability 0.
Thus, in this example, the left-hand side of \eqref{eq:IPA} is different from the right-hand side.

The essential difficulty in this example is the discontinuity in $V(x,\theta)$ as we vary $x$.
This allows the derivative of $V(x,\theta)$ to be 0 for a set of $\theta$ with probability 1 while the derivative of $V(x,\theta)$ is nevertheless non-zero.
This failure of IPA caused by $V(x,\theta)$ being discontinuous in $x$ is similar to its failure in our setting.
While we have a closed form analytic expression for $\mathbb{E}[V(x,\theta)]$ in this simple example, in situations like $\TOPTC$ where we do not, this issue prevents the use of IPA for estimating derivatives.

\paragraph{SAA}

We now briefly review SAA.
As above, we want to maximize 
$\mathbb{E} [V(x, \theta)]$.
In SAA, we first sample $\theta_{1}, \theta_{2}, \ldots, \theta_{M}$  (referred to as base samples) i.i.d. from the distribution of $\theta$. Then to optimize $\mathbb{E} [V(x, \theta)]$, SAA applies a deterministic gradient-based optimization algorithm 
to optimize $\widehat{V}_{M}(x)$, where
$\widehat{V}_{M}(x)=\frac{1}{M} \sum_{m=1}^{M} V\left(x, \theta_{m}\right)$.
More details about SAA can be found in \citep{saa_shane}.

As we demonstrate later in the section, writing our acquisition function $\TOPTC(X_1)$ in this form using the reparameterization trick results in a $V_M(x)$ that is discontinuous in $x$, where the number of discontinuities grows with $n$. 
This can also be seen by applying SAA to the example above, $V(x,\theta)= 1\{\mu(x) + \sigma(x) \theta \le 0\}$.  In this example, 
$\widehat{V}_{M}(x)=\frac{1}{M} \sum_{m=1}^{M} 
1\{\mu(x) + \sigma(x) \theta_m \le 0\}$, which has jumps of size $1/M$ at $x$ where $\mu(x) + \sigma(x) \theta_m = 0$ for any $m$.
Discontinuous functions are difficult to optimize well, especially when there are many discontinuities. 
Moreover, seeking to improve the accuracy of $\widehat{V}_M(x)$ by increasing the number of samples $M$ also 
increases the number of points $x$ with a discontinuity.
This makes it difficult to use SAA for such discontinuous functions.

\subsection{Discontinuities with the Reparameterization Trick}
 

We now show how discontinuities arising from the reparameterization trick prevent the use of IPA and SAA for optimizing $\TOPTC$. Using the reparameterization trick, we first write $Y_f$ and $Y_g$ as deterministic functions of normal random variables:
$Y_f = \mu_0(X_1) + R_0(X_1)Z_f$, $Y_g = \mu_0^c(X_1) + R_0^c(X_1)Z_g$, where $Z_f$, $Z_g$ are two $q$-dimensional independent standard normal random variables and $R_0(X_1)$, $R_0^c(X_1)$ are the Cholesky decompositions of $K_0(X_1,X_1)$ and $K_0^c(X_1,X_1)$. 


\newcommand{\E}{\mathbb{E}}

To optimize $\TOPTC$,  we would replace the values of $Y_f$ and $Y_g$ in the definition of  $\TOPTC(X_1)$ with these expressions. We focus on the term $\mathbb{E}_0[f^*_0 - f^*_1]$, 
since this is where the difficulty lies, and focus on the non-batch case for simplicity. We have: 
\begin{align*}
\E_0[f^*_0 - f^*_1]
&=\E_0[
(f_0^* - f(X_1))^+ \cdot \mathbbm{1}\{g(X_1) \leq 0 \}]\\
&= \E_0[ (f_0^* - \mu_0(X_1) - R_0(X_1)Z_f)^+ \cdot \mathbbm{1}\big{\{} \mu_0^c(X_1) + R_0^c(X_1)Z_g \leq 0 \big{\}}].
\end{align*}

The key difficulty for both IPA and SAA arises because 
the term 
$\mathbbm{1}\Big{\{} \mu_0^c(X_1) + R_0^c(X_1)Z_g \leq 0 \Big{\}}$
has a discontinuity as we vary $X_1$, at points where 
$\mu_0^c(X_1) + R_0^c(X_1)Z_g$ is 0.

We describe this difficulty, beginning with IPA.
To use IPA to estimate the gradient of $\TOPTC$ in support of a gradient-based optimization method, one would first sample $Z_f$ and $Z_g$ and then use the gradient with respect to $X_1$ of the expression inside the expectation, 
$ V(X_1, \theta) = (f_0^* - \mu_0(X_1) - R_0(X_1)Z_f)^+ \cdot  
\mathbbm{1}\big{\{} \mu_0^c(X_1) + R_0^c(X_1)Z_g \leq 0\}$, 
as an estimator of $\nabla_{X_1} \mathbb{E}_0[f^*_0 - f^*_1] = \nabla_{X_1} \mathbb{E}_0[V(X_1, \theta)]$, where $\theta = (Z_f, Z_g)$.

Unfortunately, however, the term 
$\mathbbm{1}\big{\{} \mu_0^c(X_1) + R_0^c(X_1)Z_g \leq 0 \big{\}}$
is discontinuous, taking two different constant values for different values of $X_1$. While modifying $X_1$ affects the probability of feasibility, the derivative of this term with respect to $X_1$ will be $0$ (as long as $X_1$ avoids the boundary between feasible and infeasible regions). This causes the gradient estimator to be biased, preventing its use for optimizing the acquisition function.




\begin{figure*}[t]
\begin{center}
\includegraphics[height=4cm,width=7cm]{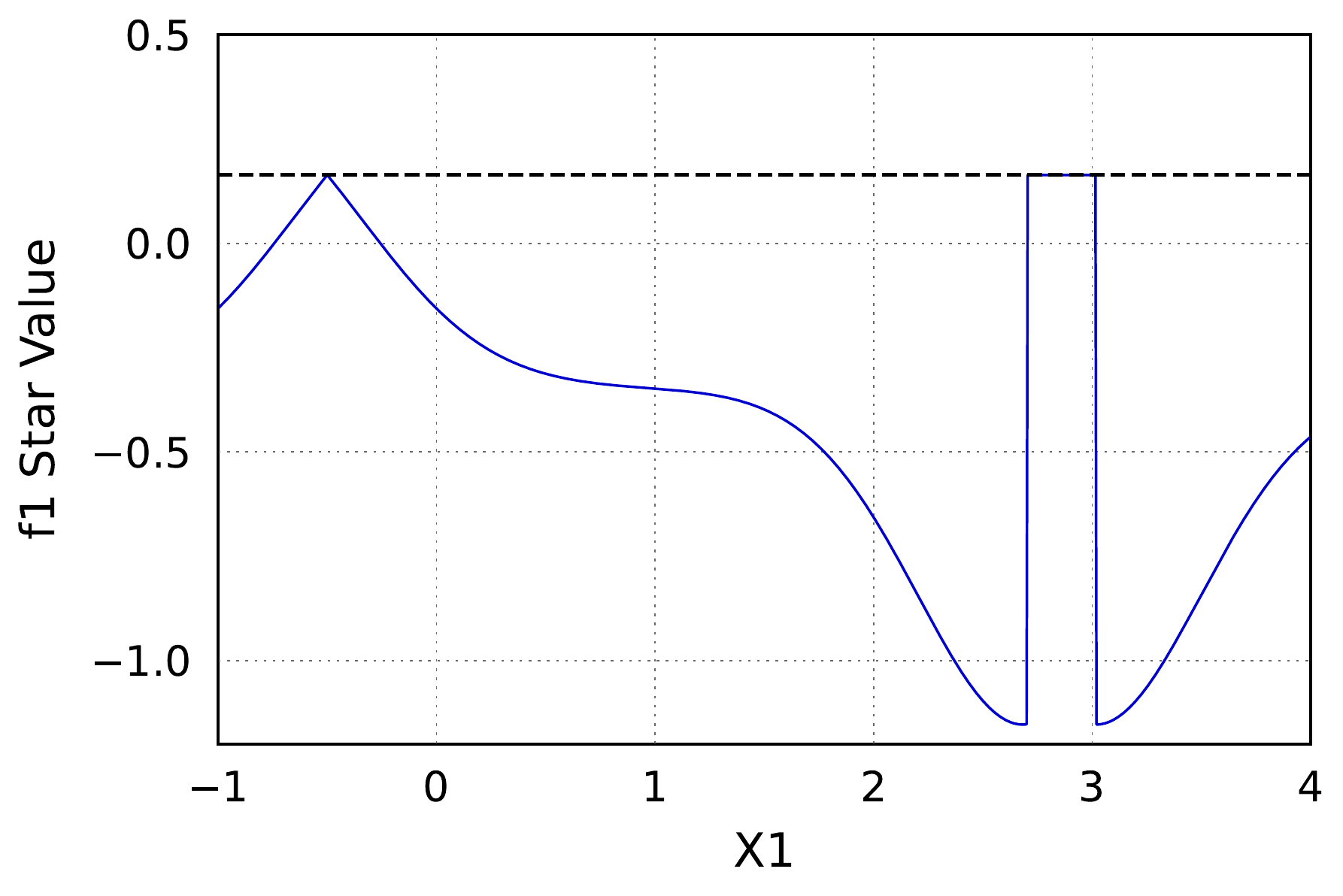}
\includegraphics[height=4cm,width=7cm]{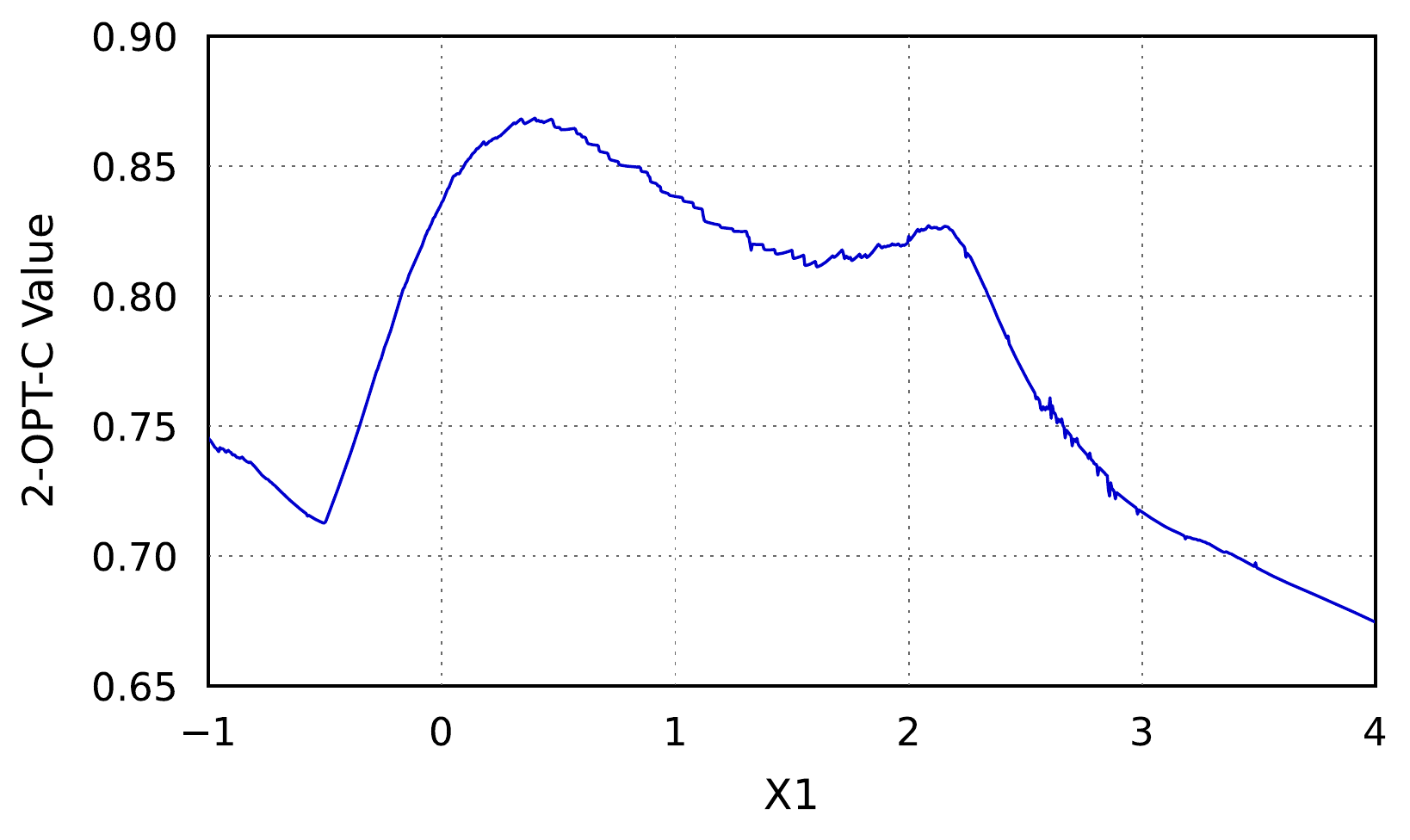}
\caption{Left: 
SAA approximation to $f_1^*$ using one base sample vs $X_1$ in a constrained optimization problem with a 1-dimensional input and one constraint. This SAA approximation is  
$ f_1^*(X_1) = f_0^* - [f_0^* - f(X_1)]^+ \cdot \mathbbm{1} \{ g(X_1) \leq 0\}$. It has discontinuities or jumps in $X_1$ when $g(X_1) = 0$. The black horizontal dashed line is drawn at $f^*_0$. Right: SAA approximation to $\TOPTC(X_1)$ using 256 base samples vs $X_1$ in the same problem.
    As sample size increases, jumps shrink in size but grow in number. Computation also grows as dimensions, constraints, or samples are added.
    The surface's roughness caused by these many small jumps (zoom in to see them better) prevents efficient optimization with more than a few dimensions or constraints.}
\end{center}
\end{figure*}




The discontinuity in 
$\mathbbm{1}\big{\{} \mu_0^c(X_1) + R_0^c(X_1)Z_g \leq 0 \big{\}}$
also creates a problem for 
SAA.  
In our context, to optimize our acquisition function $\mathbb{E}_0[f^*_0 - f^*_1]$, SAA would 
sample standard normal random vectors $Z_{f,m}$ and $Z_{g,m}$ of appropriate dimension and then optimize the following over $X_1$:
\begin{equation}
    \frac{1}{M} \sum_{m = 1}^M \bigg{[}(f_0^* - \mu_0(X_1) - R_0(X_1)Z_{f,m})^+ \cdot  \mathbbm{1}\big{\{} \mu_0^c(X_1) + R_0^c(X_1)Z_{g,m} \leq 0 \big{\}} \Big{].} 
    \label{eq:SAA}
\end{equation}

Unfortunately, the discontinuity in 
$\mathbbm{1}\big{\{} \mu_0^c(X_1) + R_0^c(X_1)Z_g \leq 0 \big{\}}$ causes \eqref{eq:SAA} to be discontinuous,
as shown in Figure 2,
making it difficult to optimize.
To combat these discontinuities, one approach is to average over a larger number of samples $M$, paying a higher computational cost.
This shrinks the jumps but increases their number. Many jumps (even small ones) make gradient-based methods fail. 
Instead, SAA for constrained problems requires derivative-free methods. Derivative-free methods are much slower for acquisition-function optimization than derivative-based ones (\citep{botorch} Appendix F, \citep{qEI} Figs 3 \& 4), especially for the problems with higher dimensions or more constraints. 
In numerical experiments in \S5.1, we compare with an SAA-based implementation of two-step lookahead and find that it requires more computation and offers degraded query efficiency compared to our method.

\section{Optimizing \texorpdfstring{$\TOPTC$}{Lg} Using Likelihood Ratios}
\label{sec:computation}

Evaluating $\TOPTC(X_1)$ requires performing a simulation where each replication samples $Y$ and then evaluates $\max_{x_2} \in A(\delta) \alpha(X_1,x_2,Y)$.  Averaging these replications gives a Monte Carlo estimate of $\TOPTC(X_1)$. Optimizing $\TOPTC(X_1)$ using such Monte Carlo estimates is difficult because of noise from simulation and because there is not a straightforward way to obtain derivatives.

Indeed, we show in \S\ref{sec:discontinuity} that the two widely used approaches to efficiently optimizing Monte Carlo acquisition functions, IPA and SAA, fail to optimize $\TOPTC$ well because constraints cause discontinuities in the surface they sample using the reparameterization trick.

Here, we develop a novel approach for optimizing Monte Carlo acquisition functions like $\TOPTC$ without using the reparameterization trick, overcoming the challenges created by these discontinuities. 
This approach uses the likelihood ratio method to derive an unbiased estimator of the gradient for $\TOPTC$. 
We then use this novel estimator in multistart stochastic gradient ascent to maximize $\TOPTC$. (For details on multistart stochastic gradient ascent for maximizing a Monte Carlo acquisition function, see \citep{qEI}.)
To the best of our knowledge, we are the first to demonstrate the benefits of the likelihood ratio method for acquisition function gradient estimation in BO.

\subsection{Background on the Likelihood Ratio Method}

We first give background on the likelihood ratio method using generic notation before describing how we use it in our setting.
Given a generic random variable $\theta(x)$ whose distribution depends on a control vector $x$ with density $p(\theta; x)$ and a function $V(x, \theta(x))$, our goal is to solve 
$\max_x \mathbb{E}[V(x, \theta(x))]$.
To do this, we estimate the gradient of 
$\mathbb{E}[V(x, \theta(x))]$ for use within multistart stochastic gradient.
\savespace{Note that, unlike IPA and SAA, in the likelihood ratio method the distribution of the random variable (here $\theta(x)$) may depend on $x$.}

To provide this gradient estimator,
we first choose a density $\tilde{p}(\theta)$ that does not depend on $x$ and for which
$\{\theta : p(\theta,x) > 0 \} \subseteq  
\{\theta : \tilde{p}(\theta) > 0 \}$ for all $x$.
Using this density, we construct the likelihood ratio, $L(\theta; x) = p(\theta; x) / \tilde{p}(\theta)$.
We then have that 
\begin{equation*}
\mathbb{E}[V(x,\theta(x))]
= \int V(x,\theta)\,p(\theta;x)\,d\theta
=
\int V(x,\theta)\,L(\theta; x) \tilde{p}(\theta)\,d\theta.
\end{equation*}
This is referred to as importance sampling and $\tilde{p}$ is referred to as the importance sampling distribution.

The likelihood ratio method uses this expression to construct a gradient estimator.
Under regularity conditions \citep{l1990unified}, we can exchange the gradient operator and integration,
\begin{equation*}
\nabla_x
\mathbb{E}[V(x,\theta(x))]
=
\nabla_x
\int V(x,\theta)\,L(\theta; x) \tilde{p}(\theta)\,d\theta
=
\int 
\nabla_x
V(x,\theta)\,L(\theta; x) \tilde{p}(\theta)\,d\theta.
\end{equation*}

From this we can create an unbiased estimator of the gradient of $\mathbb{E}[V(x,\theta(x))]$ by sampling $\theta$ from the density $\tilde{p}(\theta)$ and returning as our estimator 
$\nabla_x V(x,\theta) L(\theta; x)$.

A natural choice for the importance sampling distribution suggested in \citep{likelihood}
is to take 
$\widetilde{p}(\theta)=p(\theta, \tilde{x})$ 
when estimating 
$\nabla_x  \mathbb{E}[V(x,\theta(x))]$ 
at $x = \tilde{x}$. We make this choice when designing an unbiased gradient estimator for $\TOPTC$. We discuss this in detail later in \S\ref{sec:optimizing}.

\savespace{To illustrate why the likelihood ratio method is able to overcome discontinuities, consider the example from our discussion of IPA and SAA.  \pfcomment{explain what LR does for this example.} 

As described above, IPA approach only allows the dependence of $\theta$ in the sample function $V(\theta)$. However, the generalized version of the likelihood ratio method allows the dependence of $\theta$ on both the sample function and the probability density. Therefore, as we will show in \S5, through the likelihood ratio method, we can decompose the dependence on $\theta$ in the sample function of the IPA approach into two parts. The first part of the dependence is still expressed in the sample function. The second part of the dependence is expressed in the probability density. As a result, we are able to shift the dependence on $\theta$ which causes the discontinuity in the sample function to the probability density. Then, we solve the discontinuity issue through the likelihood ratio. 
}

\subsection{Optimizing 2-OPT-C}
\label{sec:optimizing}
To apply the likelihood ratio method to develop a gradient estimator for $\TOPTC$,
we begin by rewriting the expression to be differentiated using importance sampling.

Mapping the notation of our generic discussion of importance sampling onto our specific problem, 
our control $x$ is the batch of points $X_1$, our random variable $\theta$ is  $Y$ (whose distribution depends on $X_1)$, and $V(x,\theta)$ is $\max_{x_2} \alpha(X_1, x_2, Y)$.
$Y$ is a multivariate normal random variable whose density we write $p(y;X_1)$, noting that its mean and covariance matrix are determined by $X_1$.
We let $p(y;\tilde{X}_1)$ be our importance sampling density for $Y$, for some fixed point $\tilde{X}_1$.
$\Xtilde_1$ is arbitrary for now but is specified below. Our likelihood ratio is then $L(y;X_1,\tilde{X}_1) = p(y;X_1) / p(y;\tilde{X}_1)$.


\savespace{
We then rewrite $\TOPTC$ using the likelihood ratio $L(y; X_1, \Xtilde_1) = p(y; X_1)/p(y; \Xtilde_1)$
as:
\begin{align*}
    \TOPTC(X_1^\epsilon) = \int \max_{x_2 \in A(\delta)} \alpha(X_1^\epsilon, x_2, y) L(y;X_1^\epsilon,\Xtilde_1) p(y;\Xtilde_1)\,dy,
\end{align*}
where $X_1^{\epsilon} = X_1 + \epsilon$, $\epsilon$ is an arbitrary vector (later, we will differentiate with respect to $\epsilon$ at $\epsilon=0$).
This integral can be understood as the expectation of 
$\max_{x_2 \in A(\delta)}\alpha (X_1^{\epsilon}, x_2, Y)L(Y; X_1^{\epsilon}, \Xtilde_1)$, where $Y$ is distributed according to $p(y;\Xtilde_1)$ rather than $p(y;X_1)$.

Critically, since $\Xtilde_1$ stays fixed as we vary $\epsilon$, the distribution of $Y$ also stays the same in this new alternate expression for $\TOPTC$. Moreover, for any given $Y$, the expression $\max_{x_2 \in A(\delta)}\alpha (X_1^{\epsilon}, x_2, Y)L(Y; X_1^{\epsilon}, \Xtilde_1)$ is continuous in $\epsilon$. This is the key property that overcomes the previously described challenge created by discontinuities.

To create an unbiased estimator of the gradient of $\TOPTC$ with this understanding and alternate expression, we first observe that Theorem~\ref{th:1} below allows us to interchange the integral and gradient:
\begin{align}
    \nabla_{\epsilon} \TOPTC(X_1^{\epsilon}) \Big|_{\epsilon = 0} &= \int \nabla_{\epsilon=0} \left[\max_{x_2 \in A(\delta)}\alpha (X_1^{\epsilon}, x_2, y)L(y; X_1^{\epsilon}, \Xtilde_1)
    p(y; \Xtilde_1)\right]  dy \notag \\
    &= \int \Gamma(X_1, \Xtilde_1, y) 
    p(y; \Xtilde_1) dy \label{eq:interchange}
\end{align}
}
As discussed above, the key step in the likelihood ratio method is to interchange the integral and gradient operator. This is justified in our setting by Theorem \ref{th:1} below. The proof is provided in the supplement. As a result, we have
\begin{align}
    \nabla_{X_1} \TOPTC(X_1) 
    &= \int \Gamma(X_1, \Xtilde_1, y) 
    \,p(y; \Xtilde_1) dy,
    \label{eq:interchange}
\end{align}
where
\begin{align*}
    \Gamma(X_1, \Xtilde_1, y) 
    := &\nabla_{X_1} \left[ \max_{x_2 \in A({\delta})}\alpha (X_1, x_2, y)L(y; X_1, \Xtilde_1) \right]\\[0.25em]
    = &\left[\nabla_{X_1} \alpha (X_1, x_2^{*}, Y)\right]L(y; X_1, \Xtilde_1) + \alpha (X_1, x_2^{*}, Y)\left[\nabla_{X_1} p(y; X_1)\right] / p(y;\Xtilde_1)
\end{align*}

with $x_2^* \in \argmax_{x_2 \in A(\delta)} \alpha (X_1, x_2, Y)$. The last equality is by the envelope theorem \citep{envel},i.e., $\max_{x_2 \in A(\delta)} \alpha (X_1, x_2, Y)$ can be differentiated with respect to $X_1$ by first optimizing over $x_2$ given $X_1$, then holding $x_2^*$ fixed while differentiating with respect to $X_1$. From now on, we will drop the subscript of the differential operator $\nabla$ for simplicity.

By \eqref{eq:interchange}, $\Gamma(X_1,\Xtilde_1,Y)$ is an unbiased estimator of $\nabla \TOPTC(X_1)$ when $Y$ is drawn according to $p(y;\tilde{X}_1)$.
With this stochastic gradient estimator, we then can use stochastic gradient ascent \citep{SGA} with multiple restarts to find a collection of stationary values for $X_1$. Then we use simulation to evaluate $\TOPTC(X_1)$ at these values and select the one with the highest estimated $\TOPTC$. 
This then provides a computationally efficient algorithm for optimizing $\TOPTC$. Pseudocode for using $\TOPTC$ is provided in the supplement.

As discussed above, the key step to developing our unbiased estimator is to interchange the integral and gradient operator, which is justified by Theorem \ref{th:1} below. The proof is provided in the supplement.
Although Theorem \ref{th:1} assumes $\delta > 0$, in practice, we choose $\delta = 0$ in our stochastic gradient estimator.

\begin{theorem}\label{th:1}
    We assume:
    \begin{enumerate}
        \item The prior on the objective function $f$ is a Gaussian Process $f \sim GP(\mu_f(x), K_f(x, x'))$, and the prior on the constraints $g$ is another Gaussian Process $g \sim GP(\mu_g(x), K_g(x, x'))$. These two Gaussian processes are independent.
        \item $\mu_f$ and $\mu_g$ is continuously differentiable with $x$.
        \item $K_f(x, x')$ and $K_g(x, x')$ is continuously differentiable with $x$ and $x'$.
        \item Given $n$ different points $X = (x_1, x_2, ..., x_n)$, the matrix $K_f(X, X)$ and $K_g(X, X)$ are of full rank.
    \end{enumerate}

Then $\TOPTC(X_1)$'s partial derivatives exist almost everywhere for any $\delta > 0$. 
When $\TOPTC(X_1)$ is differentiable,
\begin{align*}
\nabla \TOPTC(X_1)
    = \int \Gamma(X_1, \Xtilde_1, y)
    p(y; \Xtilde_1) dy
\end{align*}
\end{theorem}

\textbf{Choice of Importance Sampling Distribution}
We have constructed an estimator of the gradient 
$\nabla \TOPTC$ at $X_1$.
This estimator was constructed using an importance sampling distribution parameterized by $\widetilde{X}_1$. We are free to choose this $\widetilde{X}_1$ as we wish.  We recommend setting $\widetilde{X}_1$ equal to $X_1$, as our gradient estimator takes a particularly simple form and offers robust performance. 
In particular, when $\widetilde{X}_1 = {X}_1$,
$L(y; {X}_1, {X}_1) = 1$, so our gradient estimator is
\begin{align*}
    \Gamma(X_1, X_1, y) := \alpha (X_1, x_2^{*}, y)\left[\nabla p(y; X_1)\right] / p(y;X_1) + 
    \left[\nabla \alpha (X_1, x_2^{*}, y)\right],
\end{align*}
where $x_2^* \in \argmax_{x_2} \alpha (X_1, x_2, y)$. 
 


\section{Numerical Experiments}
This section numerically investigates our algorithms on problems widely used as benchmarks in the constrained BO literature. Benchmarks demonstrate that $\TOPTC$ usually provides significant improvements in query efficiency over state-of-the-art methods. In the main paper, we focus on query efficiency for the sequential setting. The supplement includes additional experiments and discussions of batch evaluations and $\TOPTC$'s computational overhead. 



The benchmark problems include three synthetic problems from \citep{Willcox}, named P1, P2, and P3, and two real-world problems, portfolio optimization and robot pushing. Detailed descriptions are in the supplement. We use these five problems in comparisons with myopic methods in \S\ref{sec:myopic}. 
In \S\ref{sec:non-myopic} we compare with the non-myopic method from \citep{Willcox}. Since code for \citep{Willcox} was not available, we compare $\TOPTC$ against the results previously published in that paper, consisting only of P1, P2, and P3.

\subsection{Experiment Setup}

\paragraph{Evaluation Metrics}We follow \citep{pesc} in our evaluation methodology.
Along with each evaluation $n$ each algorithm makes a ``recomendation'', which is the point we would evaluate if it were our last evaluation before being scored by the best point evaluated.
From this recommended point, we compute a score $f^{**}_n$ for this algorithm in this timestep $n$.
If the recommended point is feasible, then $f^{**}_n$ is its objective function value. If not, $f^{**}_n$ is the best observed feasible value so far before the recommended point.
Following \citep{pesc},
the point recommended is the one with the lowest posterior mean objective value, among those whose probability of satisfying each constraint is 0.975 or better. 
We then report the utility gap, 
    $\epsilon_n = | f_n^{**} - f^*|$,
which is the difference between this score and the global constrained optimum $f^*$. We report the log10 median utility gaps for P1, P2, P3, and the robot pushing problem in the main paper. Mean utility gap results are provided in the supplement. In the portfolio optimization problem, the optimum is unknown so we report the mean annualized return rate instead of the utility gap. 

\paragraph{Setup for \S\ref{sec:myopic}} The $\TOPTC$ implementation uses GPs with a constant zero-mean prior and ARD square-exponential kernels for both objectives and constraints. GP hyperparameters are obtained by maximizing the marginal likelihood using GPy \citep{gpy2014}. SAA+CMA\_ES is implemented similarly to $\TOPTC$ and all the hyperparameters are the same or obtained in the same way as $\TOPTC$. 




For all three synthetic problems (P1, P2, and P3), we run 150 experiment replications for all algorithms. For the two real-world problems (portfolio optimization and robot pushing), we run 50 experiment replications. For the initialization of each experiment, we randomly sample three points with at least one feasible point from a Latin hypercube design. We run $N = 40$ function evaluations for P1 and P2, $N = 60$ for P3, $N = 30$ for the portfolio optimization problem, and $N = 50$ for the robot pushing problem. We use a batch size of 1 for all five experiments.

\paragraph{Setup for \S\ref{sec:non-myopic}} The setup for \S\ref{sec:non-myopic} is nearly the same as for \S\ref{sec:myopic}, with three key differences. These arose from the need to replicate the experimental setup from \citep{Willcox}.

First, we run 500 experiment replications. Second, rather than 3 initial points per problem, we use 1. 
Third, the evaluation method differs slightly.
If the recommended point is infeasible, then rather than setting $f^{**}_n$ to the value of the best previously observed feasible point, we set it to the maximum of the objective function over all points in the domain, thus enforcing a substantial penalty.

\subsection{Comparison with Myopic Methods}
\label{sec:myopic}
\begin{figure*}[tb]
\begin{center}
\includegraphics[width=13em, height = 9em]{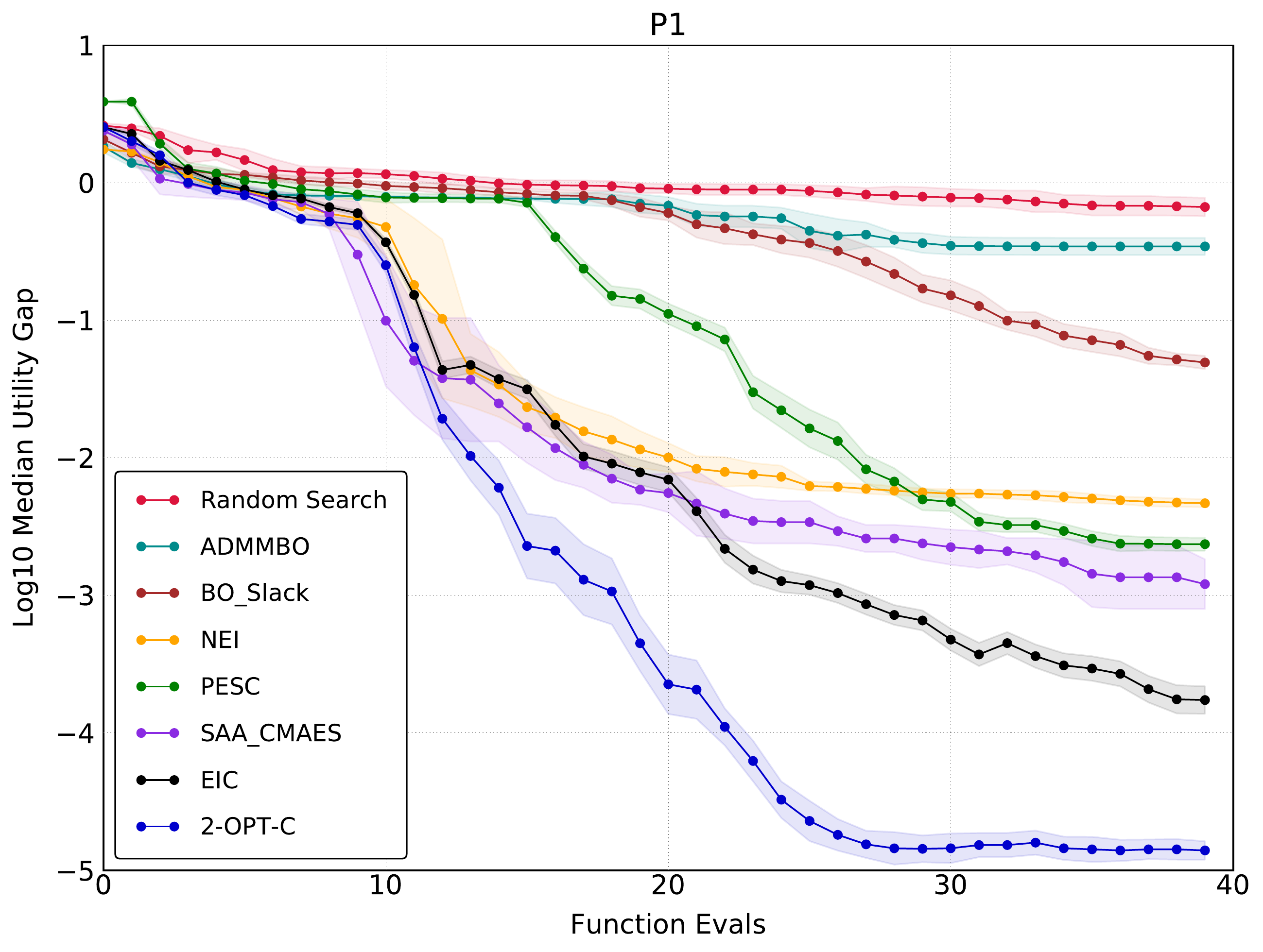}
\includegraphics[width=13em, height = 9em]{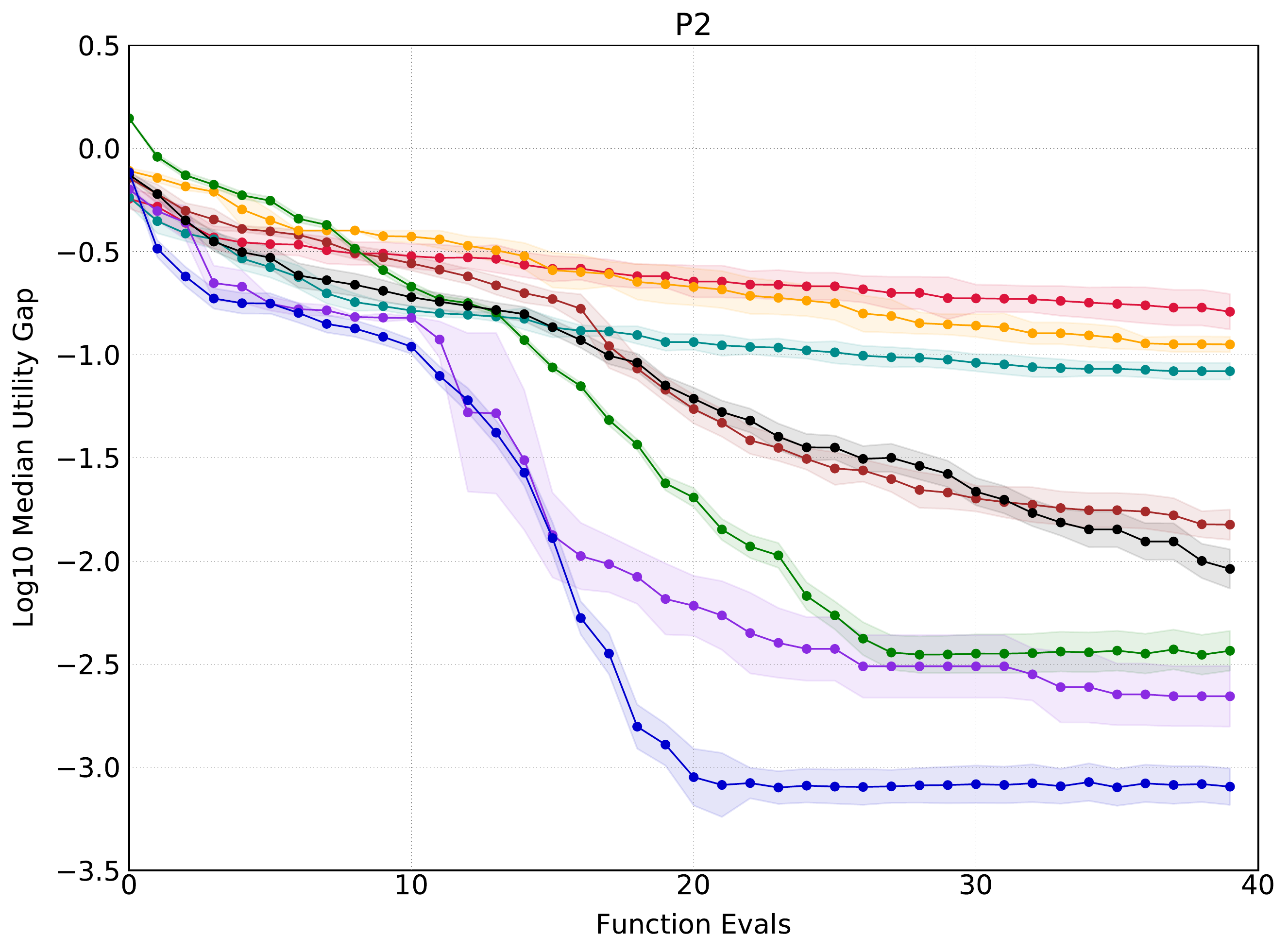}
\includegraphics[width=13em, height = 9em]{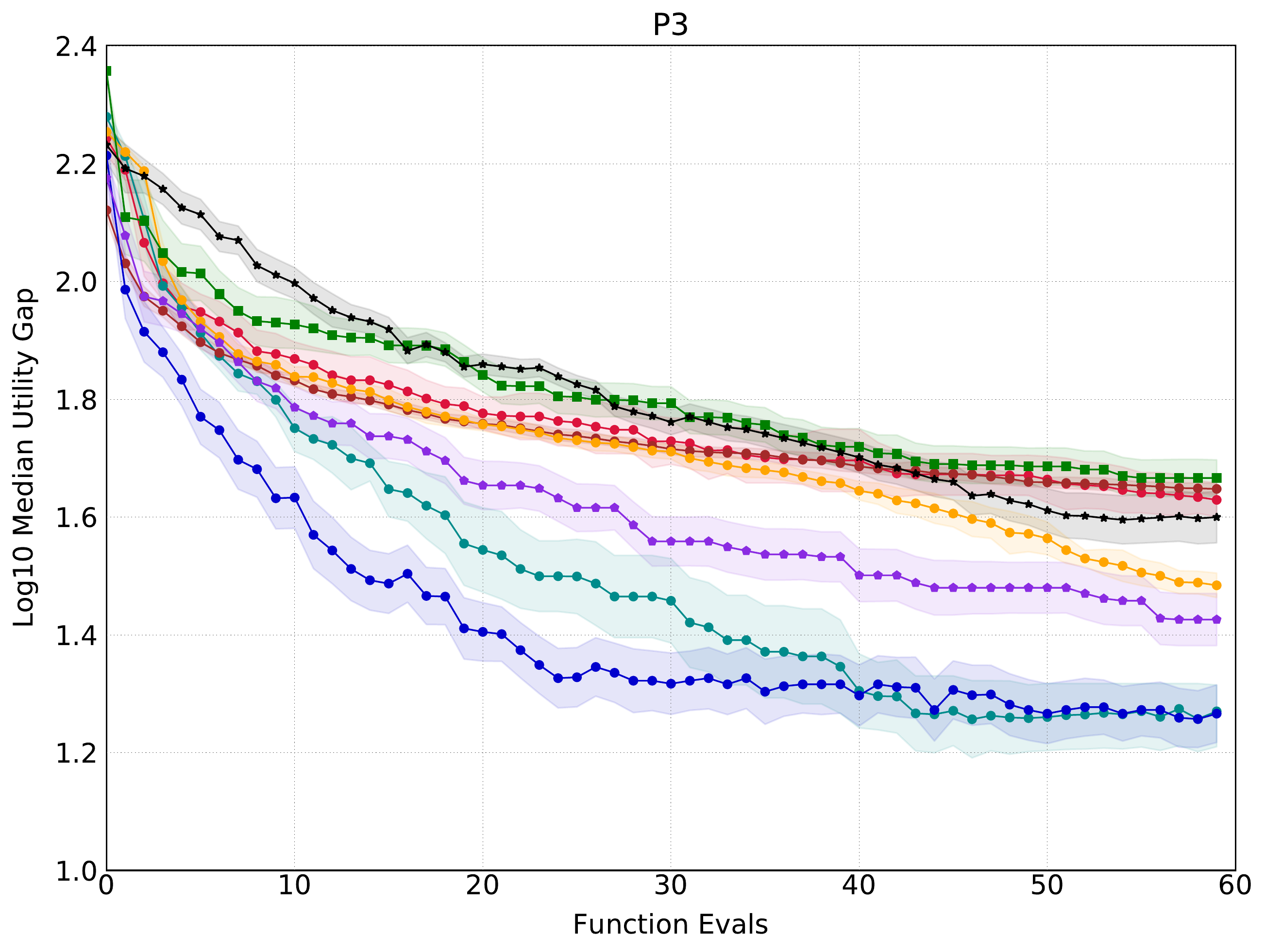}
\caption{Log10 median utility gap of Random Search, ADMMBO, BO\_Slack, NEI, EIC, PESC, SAA\_CMAES, and $\TOPTC$ with 95\% confidence intervals for problems P1, P2, and P3.}
\label{icml-historical} 
\end{center}
\vskip -1em
\end{figure*}

\begin{figure*}[tb]
\begin{center}
\includegraphics[width=13em, height = 9em]{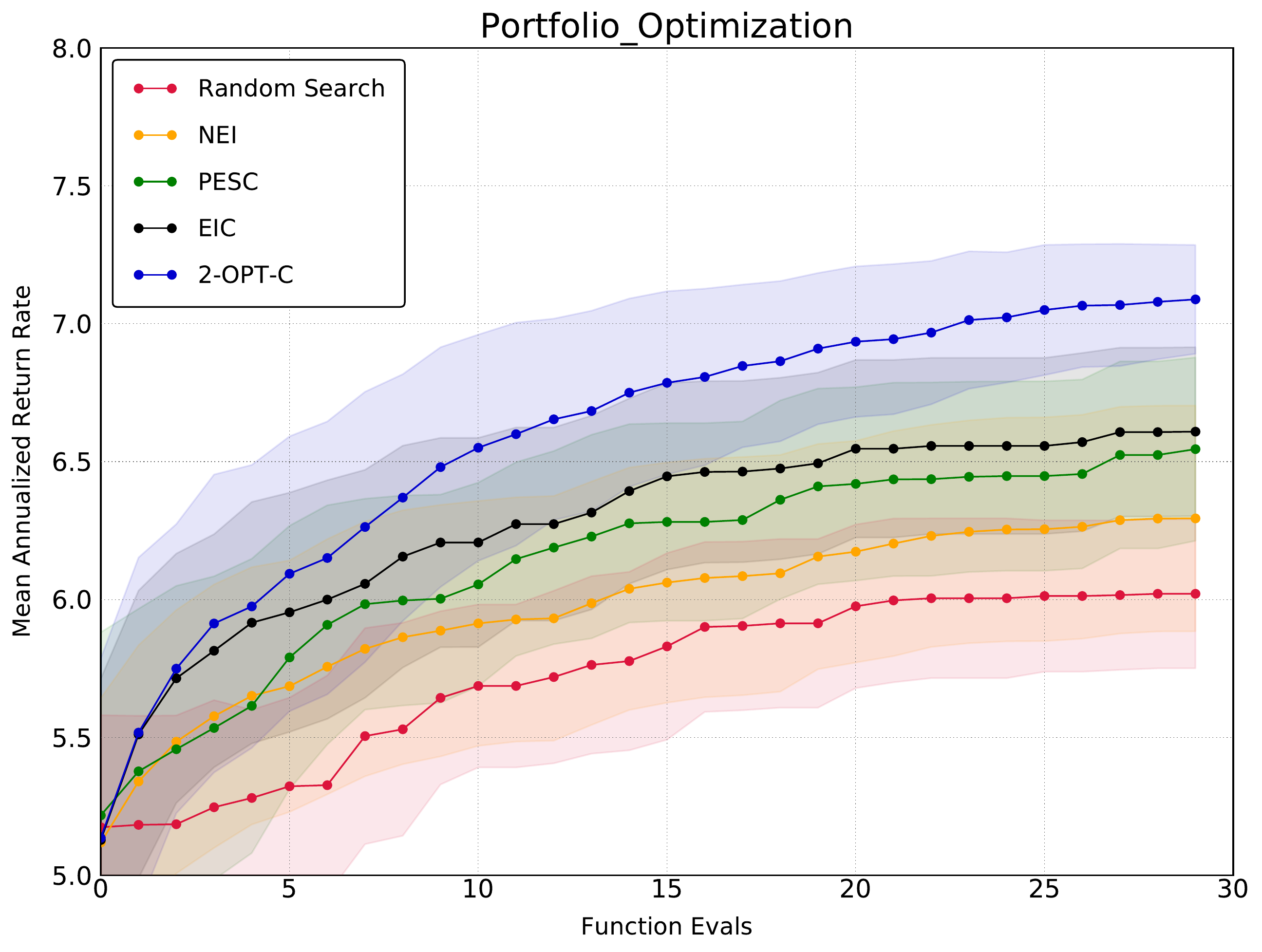}
\includegraphics[width=13em, height = 9em]{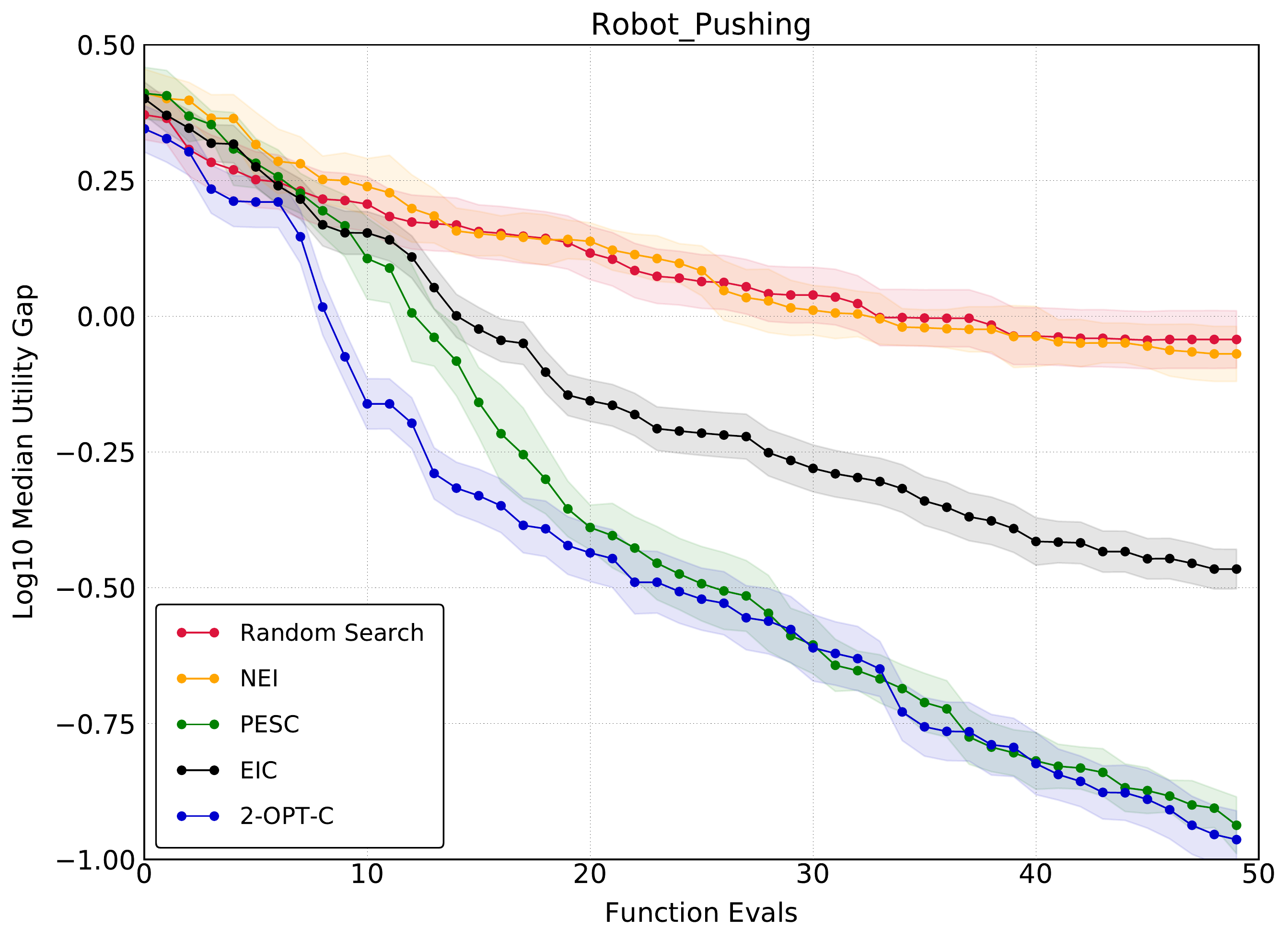}
\caption{Real-world experiment results on Random Search, NEI, EIC, PESC, and $\TOPTC$. We report the mean annualized return rate for the portfolio optimization problem and the log10 median utility gap for the robot pushing problem.}
\label{real_world} 
\end{center}
\vskip -1em
\end{figure*}
Figure~\ref{icml-historical} first compares the myopic methods ADMMBO \citep{ADMMBO}, BO\_Slack \citep{BO_slack}, NEI \citep{NEI}, EIC \citep{Gardner14, EIC_1998} and PESC \citep{pesc} on three synthetic problems.  
In general, $\TOPTC$ offers significantly lower median utility gaps than the myopic methods across all three problems. 
For example, in P1, $\TOPTC$ quickly converges to a utility gap $10^{-5}$ (in 27 evaluations), while the best of the other benchmarks (EIC) has a utility gap of $10^{-3}$ at this many evaluations.
There is only one method and problem, ADMMBO on problem P3, in which a myopic method achieves an optimality gap comparable to $\TOPTC$, but $\TOPTC$ reaches this optimality gap in fewer iterations.


\savespace{
In general, $\TOPTC$ offers significantly lower median utility gaps than the myopic methods across all three problems.  For example, in P1, $\TOPTC$ quickly converges to a utility gap $10^{-5}$ (in 27 evaluations), while the best of the other benchmarks (EIC) has a utility gap of $10^{-3}$ at this many evaluations. Moreover, while EIC continues to improve its utility gap (down to $10^{-4}$) with more iterations, PESC seems to reach a plateau after roughly 35 function evaluations.  Comparisons between $\TOPTC$ are qualitatively similar in P2, with $\TOPTC$ outperforming the best of the other benchmarks, offering a utility gap that is roughly $10^{0.5}=2$ times lower than the best of the myopic methods (PESC). For P3, $\TOPTC$ also significantly improves query efficiency compared to all the myopic methods except ADMMBO. Although ADMMBO performs similarly to $\TOPTC$, $\TOPTC$ is able to reach the same utility gap with fewer iterations.  
}


Figure~\ref{icml-historical}
additionally compares with an SAA implementation of $\TOPTC$ that uses CMA-ES \citep{CMA} to optimize the SAA to the acquisition function on P1, P2, and P3, denoted as SAA\_CMAES. We see that SAA\_CMAES generally performs better than myopic policies on all three problems (except EIC on P1), supporting the value of two-step lookahead for CBO over myopic approaches. However, due to the discontinuity introduced by SAA, it underperforms $\TOPTC$ on all three synthetic problems. It also underperforms EIC in P1 and ADMMBO in P3. In addition, it requires substantially more computation. This is because, as we note in \S \ref{sec:discontinuity}, discontinuities in the SAA to the acquisition function make it extremely difficult to optimize. We provide additional experiments to illustrate this in the supplement. 

Figure~\ref{real_world} compares $\TOPTC$ with the mostly widely-used myopic methods, NEI, PESC, and EIC, on real-world problems. As in the synthetic problems, $\TOPTC$ outperforms the competing benchmarks.
In the portfolio optimization problem, $\TOPTC$ provides roughly $0.5\%$ more annualized return than the best myopic method (EIC). In the robot pushing problem, $\TOPTC$ outperforms EIC and NEI over the full range of function evaluations; PESC eventually matches $\TOPTC$ but there is a range of function evaluations where $\TOPTC$ performs strictly better.

\interfootnotelinepenalty=10000

\subsection{Comparison with the Non-myopic Method}
\label{sec:non-myopic}
We compare $\TOPTC$ with the non-myopic rollout algorithm of \citep{Willcox} on the three synthetic problems based on the results reported in their paper.
We use the terms Rollout-1, Rollout-2, and Rollout-3 to denote this algorithm with horizons of 1, 2, and 3, respectively. 

To compare with the rollout algorithm, we extract experimental data from \citep{Willcox} and replicate their experimental setup, adding $\TOPTC$'s results to their Figures 2 and 3.

\begin{figure}[ht!]
\begin{center}
\includegraphics[width=13em, height = 8em]{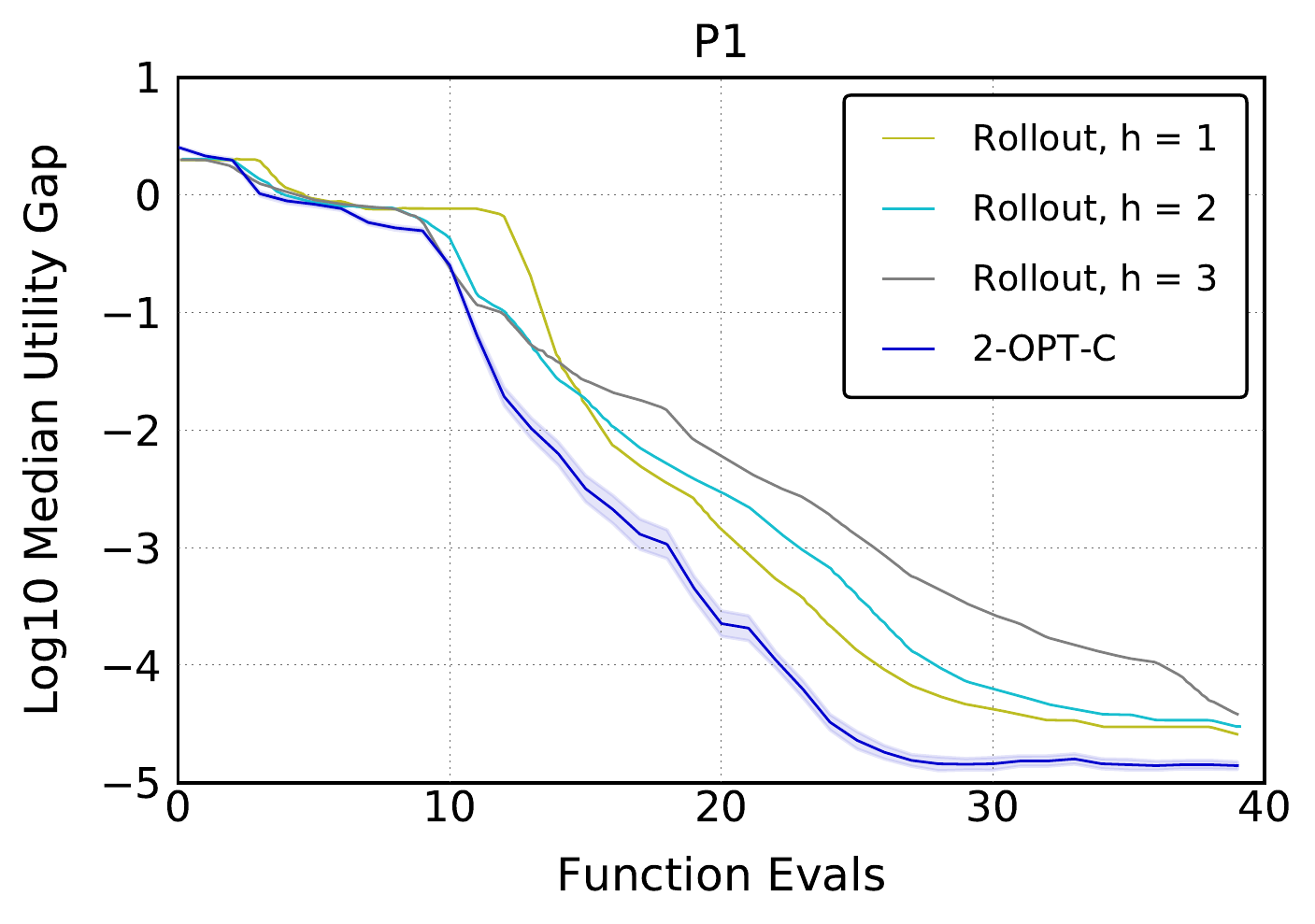}
\includegraphics[width=13em, height = 8em]{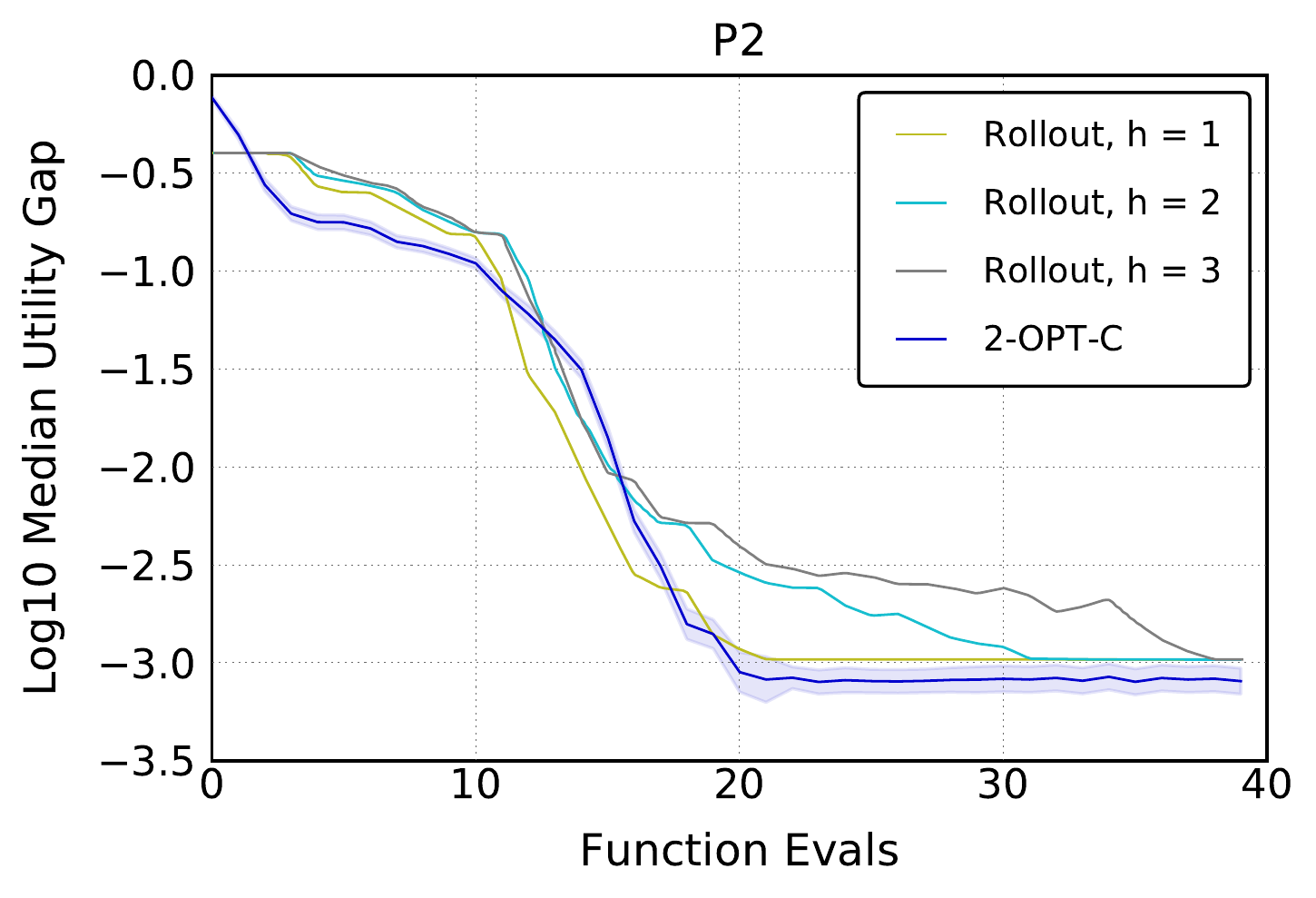}
\includegraphics[width=13em, height = 8em]{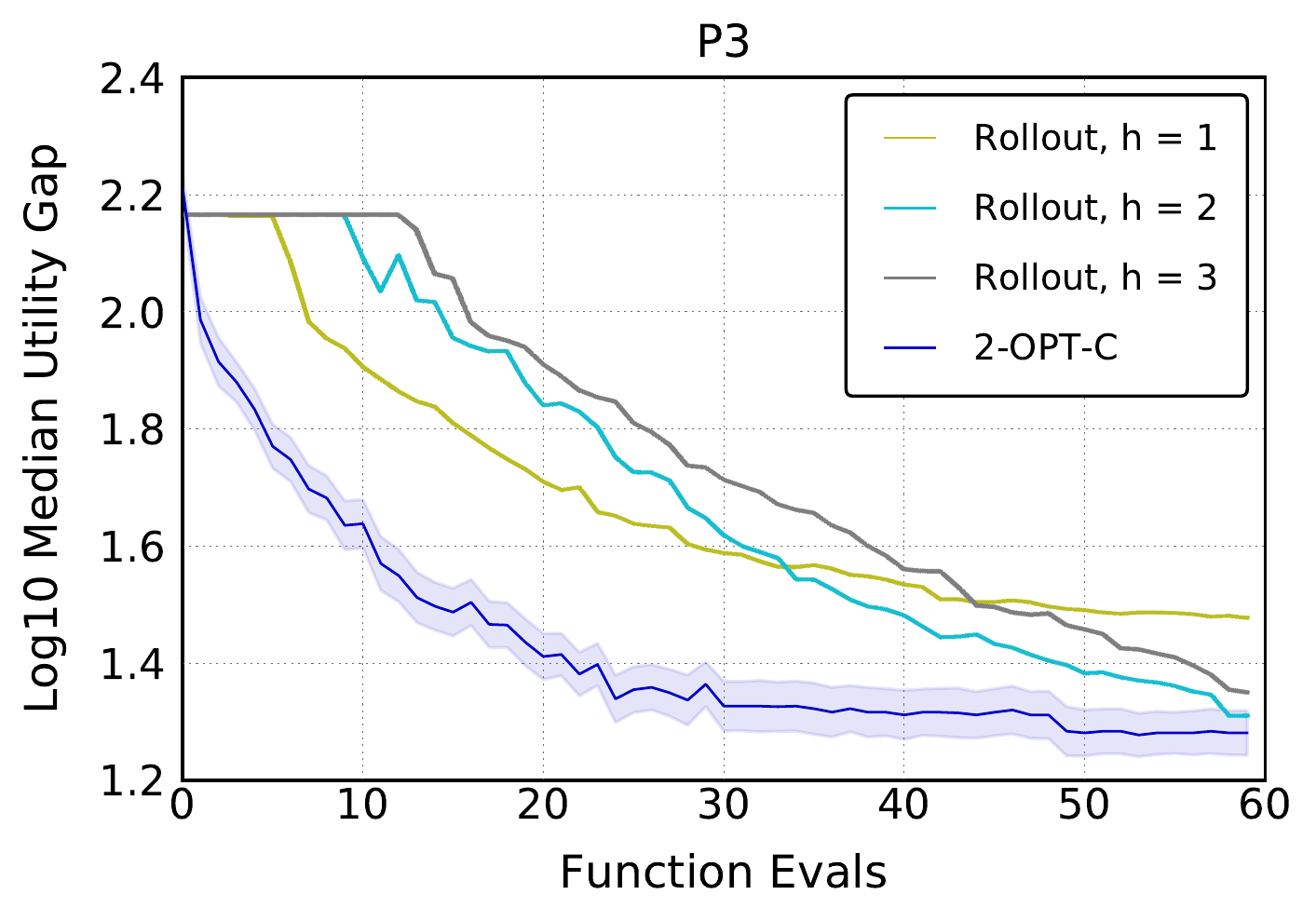}
\caption{Log10 median utility gap of $\TOPTC$ and the existing non-myopic constrained BO algorithm with different rollout horizons on P1, P2, and P3. The three lines of the rollout algorithm are extracted from Figure 3 in \citep{Willcox}. Additional information is provided in Table 1 in the supplement.
}
\label{two_step_nonmyopic}
\end{center}
\end{figure}


Figure~\ref{two_step_nonmyopic} reports the results of these comparisons. It shows that $\TOPTC$ provides query efficiency that is at least as good as the non-myopic rollout methods with different horizons on all three problems and is sometimes substantially better. In detail, for P1, $\TOPTC$ increases solution quality by more than a factor of 2 ($10^{-4.59} / 10^{-4.92}$) compared to the best rollout strategies, i.e., Rollout-1. For P3, $\TOPTC$ is able to improve solution quality over the best competing rollout method by roughly a factor 2 when the evaluation budget is tight (after 30 function evaluations). 

$\TOPTC$ also uses less computation than the rollout algorithm \citep{Willcox}. Since the code from \citep{Willcox} is not publicly available, we refer to discussion in \citep{practical} of the computational time of an unconstrained version of the rollout algorithm, which we understand from private communication has a similar cost to \citep{Willcox}. This states that the unconstrained rollout strategies require from 10 minutes to 1 hour to evaluate a single point even on a low-dimensional problem. This may be even larger in constrained settings, since we need a separate GP to model the constraint function. $\TOPTC$ is more computationally efficient. On P3, for instance, $\TOPTC$ optimizes a batch of 5 points in 25-30 minutes. 

Rollout-1 is actually attempting to optimize the same theoretical acquisition function used by $\TOPTC$. Indeed,  \citep{Willcox} uses EIC as its rollout strategy, which is optimal over a horizon of $h=1$. Rollout-1 is then trying to identify the point to evaluate now that would be optimal if one additional point were evaluated after using EIC. If this point could be found exactly, it would be 2-step optimal. 

Despite attempting to optimize an acquisition function that is conceptually the same as $\TOPTC$, the performance of Rollout-1 is substantially worse (the utility gap is roughly 2x larger over all budgets), despite requiring more than double the computation. This is because \citep{Willcox} optimizes this two-step acquisition function without the type of derivative estimates provided by our likelihood ratio method.

\section{Conclusion}\label{sec:conclusion}
This paper presents a non-myopic CBO algorithm, supporting both batch and sequential settings, that substantially improves both query efficiency and computation time over the one previous method focused on this class of problems for both sequential and batch settings. 
\savespace{
This method overcomes a challenge in using the reparameterization trick, arising from discontinuities created by constraints. In spite of this challenge, it efficiently optimizes a two-step optimal acquisition function using a novel likelihood-ratio-based unbiased estimator of its gradient. Through numerical experiments, we demonstrate that the proposed algorithm substantially improves query efficiency compared with existing myopic and non-myopic methods. 
}

While our method offers significant improvements in query efficiency compared to the state-of-the-art that more than offset its computational expense, there are likely more opportunities for reducing its computational cost. This, as well as a further theoretical investigation regarding our method providing such significant improvements in query efficiency compared to other methods, presents valuable directions for future work.

Faster constrained optimization of time-consuming-to-evaluate functions with our methodology enables creation of new engineering systems, better supervised learning methods, and lower-cost business operations. 
While we believe technological innovation is on balance good, new technology created using our methods also has the potential for negative effects. Understanding the societal effects of improved optimization capabilities is an important area for future research.




 
\clearpage

\title{Two-step Lookahead Bayesian Optimization with Inequality Constraints Supplement}

\begin{center}
\par\noindent\rule{\textwidth}{1.0pt}\\[2.0em]
\textbf{\Large Two-step Lookahead Bayesian Optimization with Inequality Constraints Supplement}\\[1.4em]
\par\noindent\rule{\textwidth}{1.0pt}
\end{center}







\renewcommand\thesection{\Alph{section}}




\vskip 2em

The supplement is organized as follows. 

\begin{itemize}
\item \S\ref{sec:A} presents detailed descriptions of the experiment problems from the main paper, and additional experiment results for the batch setting and computational overhead.
\item \S\ref{sec:C} presents a more general version of $\TOPTC$ that handles multiple constraints. 
\item \S\ref{sec:algo} provides detailed pseudocode for optimizing $\TOPTC$ using SGD and our likelihood ratio estimator.
\item \S\ref{sec:proof} proves Theorem 1 in the main paper. 
\item For completeness, \S \ref{sec:B} presents a more detailed treatment of batch constrained expected improvement 
defined in \S \ref{sec:EIC_def} of the main paper. 
\end{itemize}

\setcounter{section}{0}

\section{Additional Experiment Details and Results}\label{sec:A}

Here, we present detailed descriptions of the experiment problems from the main paper and additional experimental results. Specifically, we present:
\begin{itemize}
\item Details of the benchmark problems from the main paper (\S\ref{sec:problems}).
\item Additional experimental results for the batch setting and investigating the computational overhead associated with acquisition function optimization (\S\ref{sec:batch}).
\item Additional experiments studying the mean utility gap rather than the median reported in the main paper (\S\ref{sec:mean}).
\item Additional experimental results for non-myopic problems (\S\ref{sec:nonmyopic}).
\end{itemize}

\subsection{Benchmark Problems} \label{sec:problems}
Here we give details about the problems used as benchmarks in the main paper.

$\textbf{P1}:$ The first synthetic problem is a multi-modal objective with single constraint introduced in \cite{Gardner14}, which minimizes the objective function $f$ on the design space $A = [0, 6]^2$ subject to one constraint:
\begin{align*}
    f(x) &= \cos(2x_1) \cos(x_2) + \sin(x_1), \\[0.25em]
    g(x) &= \cos(x_1) \cos(x_2) - \sin(x_1) \sin(x_2) + 0.5.
\end{align*}

$\textbf{P2}:$ The second synthetic problem has a linear objective with multiple constraints and is discussed in \cite{BO_AL} and \cite{pesc}. The design space is a unit square $A = [0, 1]^2$ and the objective function and the constraints are given by:
\begin{align*}
    f(x) &= x_1 + x_2,\\
    g_1(x) &= 0.5\sin(2\pi(2x_2 - x_1^2)) - x_1 - 2x_2 + 1.5,\\
    g_2(x) &= x_1^2 + x_2^2 - 1.5.
\end{align*}

$\textbf{P3}:$ The third synthetic problem is a multi-modal 4-d objective with a single constraint proposed in \cite{Willcox}. The design space is $A = [-5, 5]^4$. The objective function and the constraint are defined as:
\begin{align*}
    f(x) &= \frac{1}{2} \sum_{i = 1}^4 (x_i^4 - 16x_i^2 + 5x_i), \\
    g(x) &= -0.5 + \sin(x_1 + 2x_2) - \cos(x_3)\cos(2x_4).
\end{align*}

$\textbf{Portfolio Optimization}:$ The first real-world problem is a portfolio optimization problem based on \cite{boyd2017}, as studied in a BO context by \cite{cvxpl}. In this problem, we optimize an algorithmic trading strategy to maximize a portfolio’s mean annualized return rate while constraining the portfolio’s risk, as measured by the standard deviation of the return rate. The portfolio simulation uses CVXPortfolio and real-world market data, as described in \cite{boyd2017}. A single function evaluation using CVXPortfolio takes between 10 and 15 minutes. The constraint value we choose for this problem is $2(\%)$.

$\textbf{Robot Pushing}:$ The second real-world problem a robot pushing problem based on \cite{max_entropy}, where the robot pushes an object from the origin, i.e., $L_{\text{object}} = (0,0)$, to an unknown target location $L_{\text{target}} \in [-5,5]^2$. The parameters to be optimized are the location of the robot, i.e., $L_{\text{robot}} = (x_{\text{robot}}, y_{\text{robot}})$ and the duration of the push $t \in [1, 30]$. Therefore, the decision variables to optimize are $(x_{\text{robot}}, y_{\text{robot}}, t)$. The objective is to minimize the distance between the location of the object after being pushed and the target location, namely the $L_2$ norm of  $(L_{\text{object after push}} -  L_{\text{target}})$. The cost function is the energy used by the robot for pushing the object which is $|| L_{\text{object after push}}|| + \epsilon$, where $\epsilon$ is some noise. The constraint value we use for this problem is 6.

\subsection{Batch Evaluations, Overhead of Acquisition Function Evaluation}
\label{sec:batch}
Here, we study two additional questions not studied in the main paper: performance in the batch setting;
and the overhead required to optimize the acquisition function. The experiment setup is the same as that of \S \ref{sec:myopic} in the main paper. 



While $\TOPTC$ is more query efficient, i.e., uses fewer evaluations of the objective function and the constraint than all other methods, it requires more computation time than myopic methods to decide where to evaluate. (It requires less computation than other non-myopic methods, discussed below.)
Thus, whether it reduces the {\it total} computation time required to solve an optimization problem depends on how time-consuming the objective and constraints are to evaluate.

If each evaluation is extremely fast, e.g., milliseconds, then even substantially better query efficiency is not enough to overcome a small elevation in overhead. If each evaluation is slow, however, e.g., minutes or hours, then the time saved via improved query efficiency makes up for a modest increase in overhead. Indeed, BayesOpt is almost always applied in the latter setting, when evaluations are slow.

We present experimental results on one of the example problems from the main paper, P3, showing that $\TOPTC$ solves optimization problems more quickly than myopic constrained BayesOpt methods for ranges of objective / constraint evaluation time typical to BayesOpt. 
We use batch sizes of $q=1$ (i.e., sequential), $q=5$ and $q=10$. All the experiments are run on Amazon Web Services c4.4xlarge instances.

Although the actual benchmark problems are fast to evaluate, we can easily simulate what wall-clock times would result given different amounts of time required to compute the objective function and constraints. We do this by simply adding the actual wall-clock time needed to perform acquisition function optimization to the simulated wall-clock time associated with the number of objective/constraint evaluations performed. We can then calculate the utility gap versus simulated wall-clock time.
We do this for four different levels of simulated wall-clock required for an objective / constraint evaluation: 0 minutes; 20 minutes; 40 minutes; and  60 minutes.

\begin{figure}[tb]
\subfloat[Evaluation time per point: 0 min.]{\includegraphics[scale = 0.3]{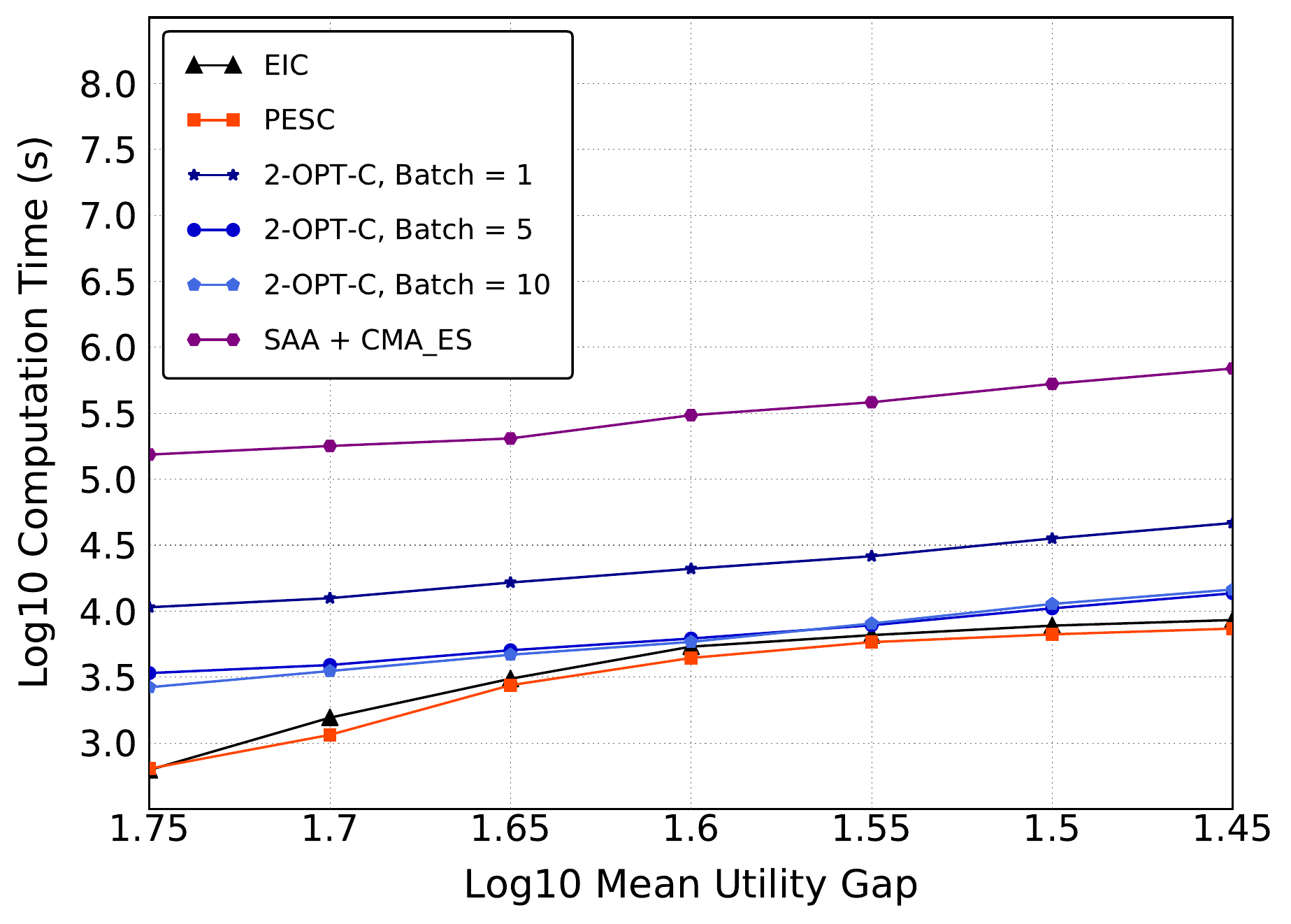}}
\subfloat[Evaluation time per point: 20 min.]{\includegraphics[scale = 0.3]{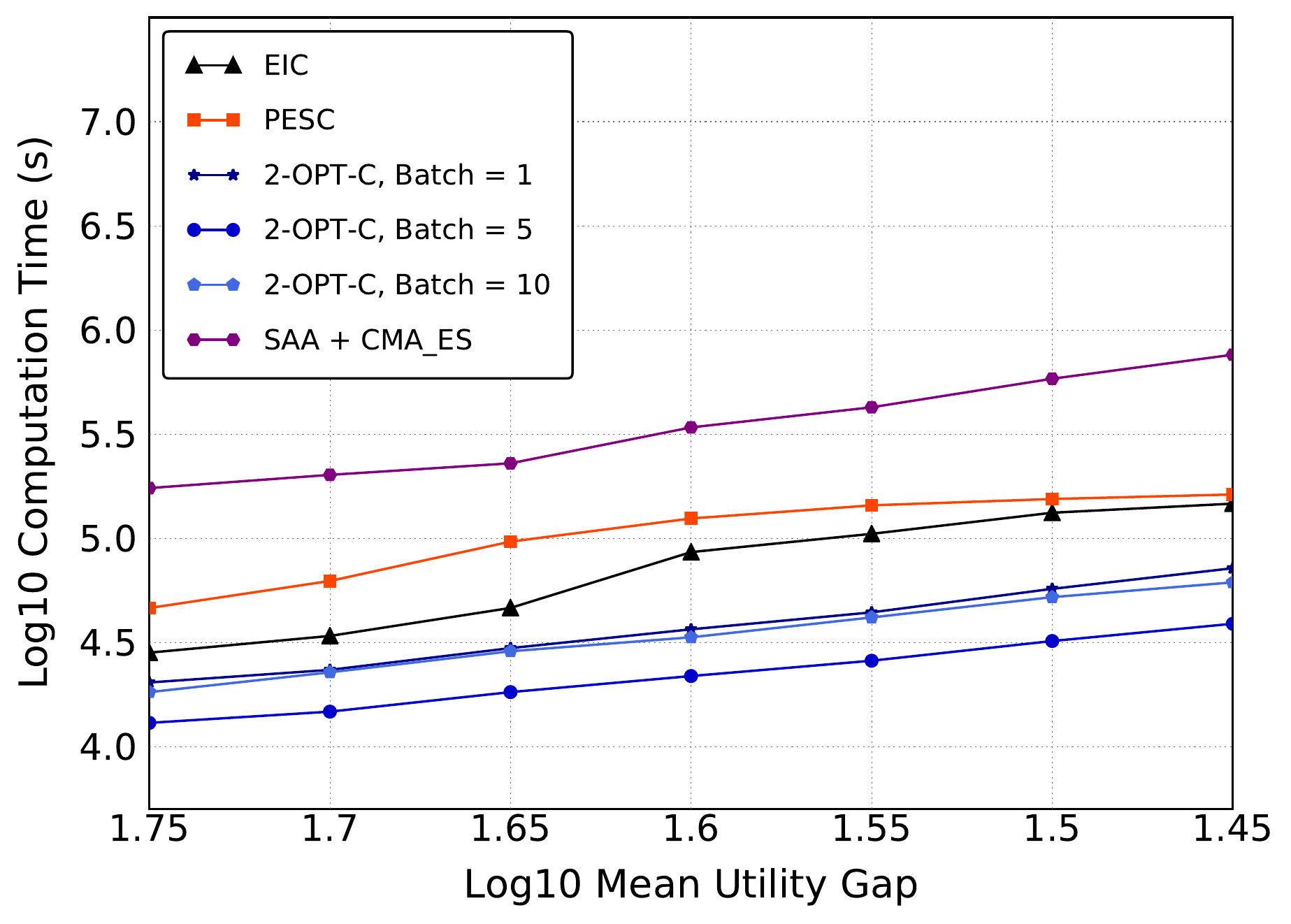}}
\\
\subfloat[Evaluation time per point: 40 min.]{\includegraphics[scale = 0.3]{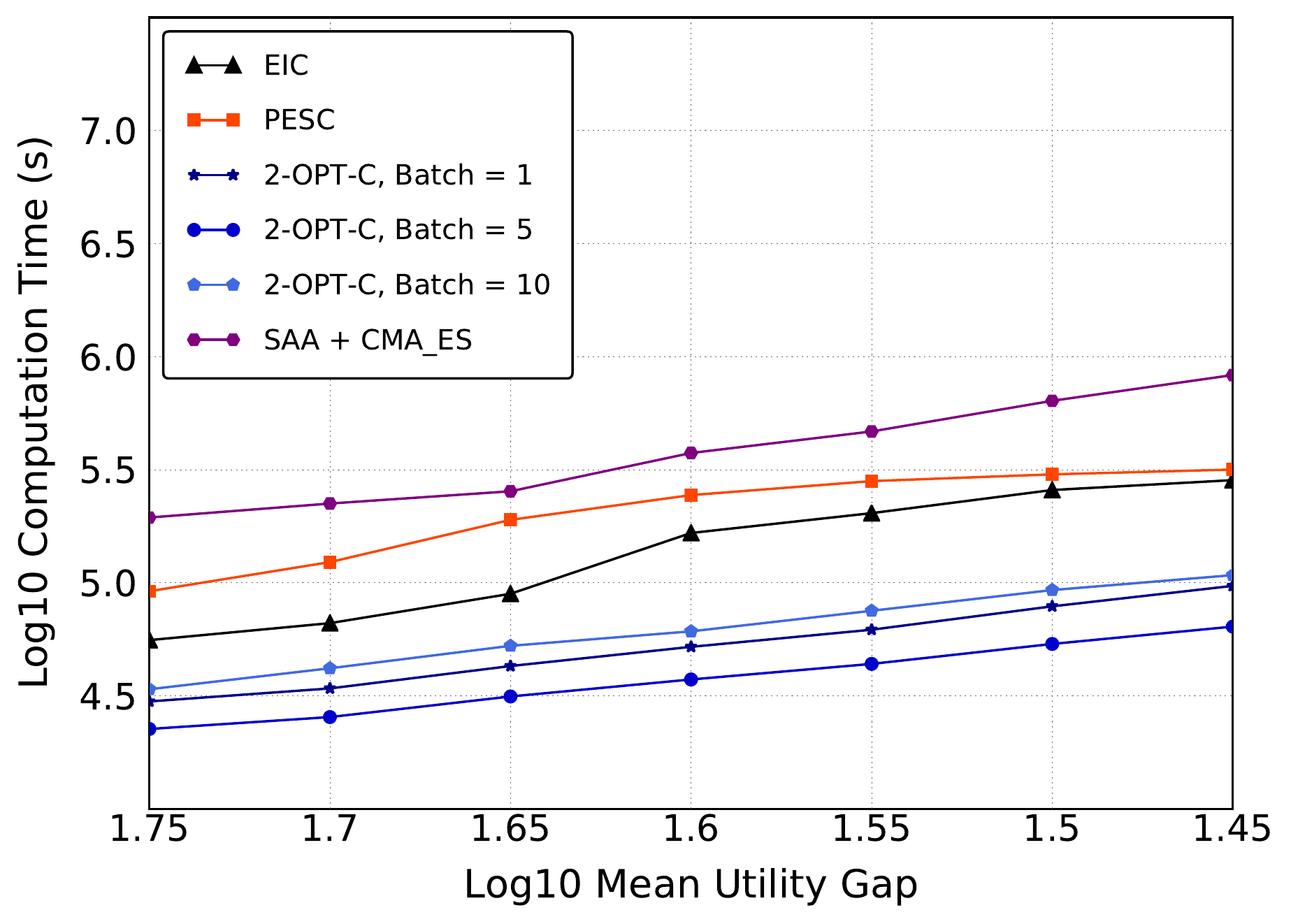}}
\subfloat[Evaluation time per point: 60 min.]{\includegraphics[scale = 0.3]{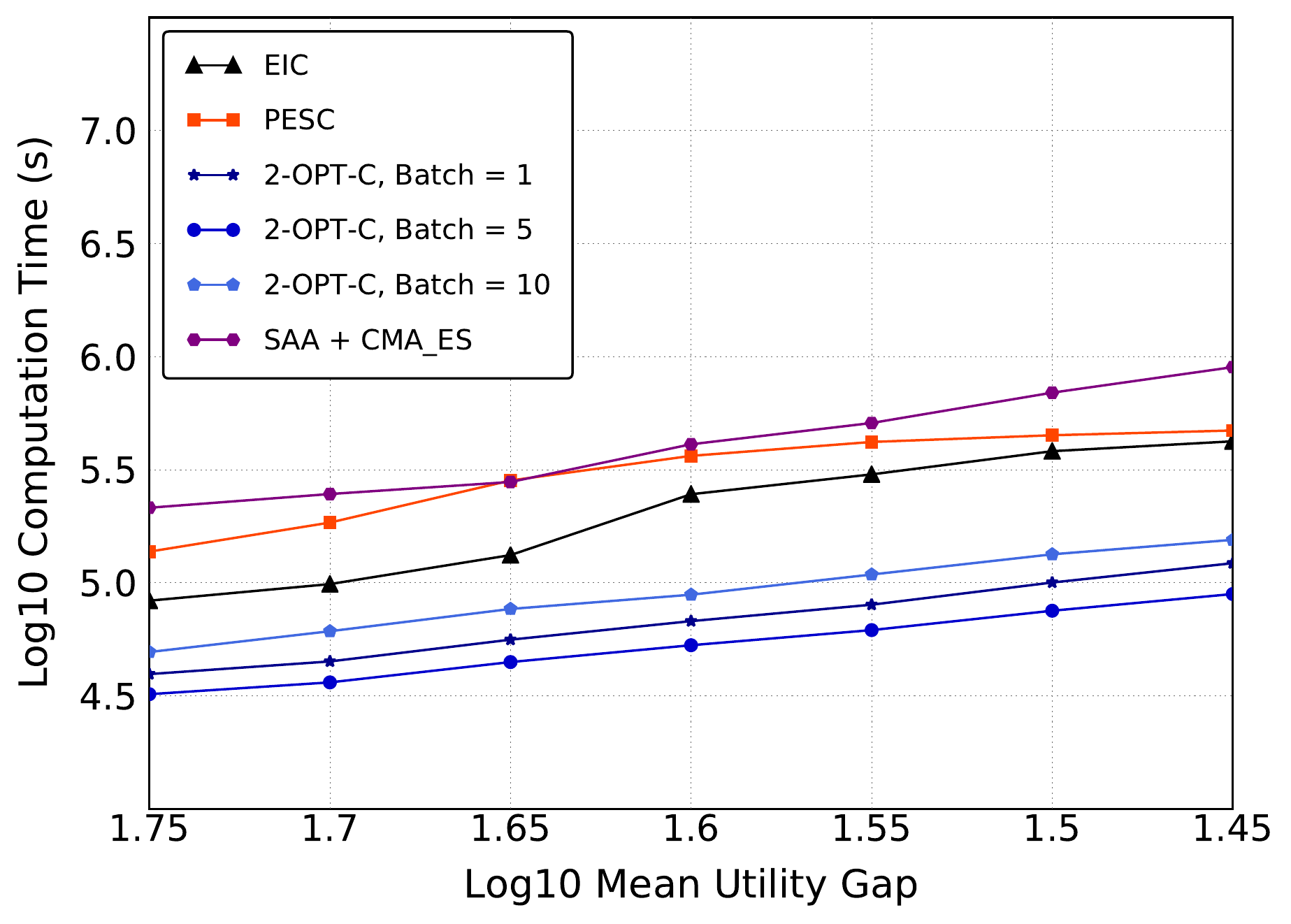}}
\caption{Runtime comparison for EIC, PESC, SAA+CMA\_ES and $\TOPTC$. Each point represents the log10 computational time required for the algorithm to achieve a certain level of log10 mean utility gap. The time needed per evaluation for objective / constraint functions ranges from 0 min (Figure a) to 20 min (Figure b), 40 min(Figure c) and 60 min (Figure d).
\label{fig:batch-overhead}
}
\end{figure}

Figure~\ref{fig:batch-overhead} shows the results of these experiments, showing total computation time to reach a given mean utility gap under $\TOPTC$ with $q=1$, $q=5$, and $q=10$ as compared with myopic BO methods. 
As the time needed per evaluation for objective / constraint functions varies from 0 to 60 minutes, we plot the total computation time v.s. the log10 utility gap in Figure~\ref{fig:batch-overhead}(a-d).

We make several observations.
First, when the objective and constraints are time-consuming-to-evaluate, $\TOPTC$ performs much better than the other methods. Although its computational overhead is higher, the time saved by its substantially better query efficiency more than makes up for overhead. 
When the objective is fast-to-evaluate (0 minutes), $\TOPTC$'s computational overhead causes it to underperform other methods. In this regime, however, one would not use Bayesian optimization --- even the relatively mild overhead of a method like constrained EI would be undesirable and it would be better to use, e.g., random search or a genetic algorithm.

Second, the batch version of $\TOPTC$ is particularly effective. This is for two reasons. First, $\TOPTC$ is optimized via a gradient-based method that retains its effectiveness in the higher-dimensional optimization problems created by batch acquisition function optimization. Thus, the overhead required to optimize the acquisition function does not grow substantially as the batch size $q$ grows:  optimizing the multiple-point acquisition function for a batch size of $q$ is faster than optimizing the single-point acquisition function $q$ times. This is because the bottleneck in optimizing $\TOPTC$'s acquisition function is the ``inner'' problem, i.e., finding the optimal $x_2^*$ given $X_1$. The time spent solving this inner problem does not grow substantially with $q$. With larger $q$, this overhead is amortized over a larger number of function evaluations.
Second, surprisingly, there is little loss in query efficiency associated with moving to the batch setting in this problem, which is evidence that $\TOPTC$ effectively optimizes its batch acquisition function. Note that one can use batch $\TOPTC$ even when evaluations must be sequential: choose $q$ points to evaluate simultaneously with one large acquisition function optimization, then evaluate them one at a time without updating the GPs until the full batch finishes. When function evaluations are less expensive, this is an appealing way to retain query efficiency while saving computational overhead.

Third, SAA+CMA\_ES performs poorly, despite the fact that it is optimizing the same theoretical acquisition function as $\TOPTC$. This points to the value of our novel approach to optimizing this acquisition function. Indeed, although Figure 1 shows that the SAA+CMA\_ES approach has better query efficiency compared to myopic methods (EIC and PESC), the discontinuity issue discussed in the main paper causes the computational overhead to be considerably larger than for both myopic methods and $\TOPTC$, while also degrading query efficiency. This in turn causes SAA+CMA\_ES to be the worst of the methods in nearly all settings considered.

Fourth, $\TOPTC$ does better when one desires a better solution quality (lower mean utility gap). This is because the convergence of myopic methods is slower: the multiplicative gap in the number of evaluations required to reach a given optimality gap grows as the optimality gap grows small.

The above discussion compares $\TOPTC$ to the myopic methods EIC and PESC and to SAA+CMA\_ES. Here we briefly discuss computation time relative to the other non-myopic method \cite{Willcox}.
For $\TOPTC$, under the sequential setting ($q = 1$), it takes roughly 15 to 20 minutes over computational overhead per evaluation. However, \cite{Willcox} is likely to take even more time: a similar code designed for the simpler unconstrained setting requires 10 minutes to 1 hour per iteration, as reported in \cite{practical} based on private communication with the authors. Modeling the constraint in a non-myopic way requires a substantially larger scenario space and also doubles the number of GPs to be modeled. This suggests that, in the sequential setting, our method is more efficient than \cite{Willcox}. Moreover, in the batch setting, optimizing $\TOPTC$ requires only a modest amount of additional computation per batch as the batch size grows, thus resulting in substantially less overhead per point. For example, choosing a batch size of 5 with $\TOPTC$ requires 20-30 minutes of overhead, or 4-5 minutes per point. 
As noted above, this is a viable method for saving computational overhead when functions evaluations are not especially slow, even if points must be deployed sequentially. This fact further emphasizes the computational benefits of $\TOPTC$ over the method of \cite{Willcox}.


\begin{figure}[tbh]
\begin{center}
\includegraphics[scale = 0.22]{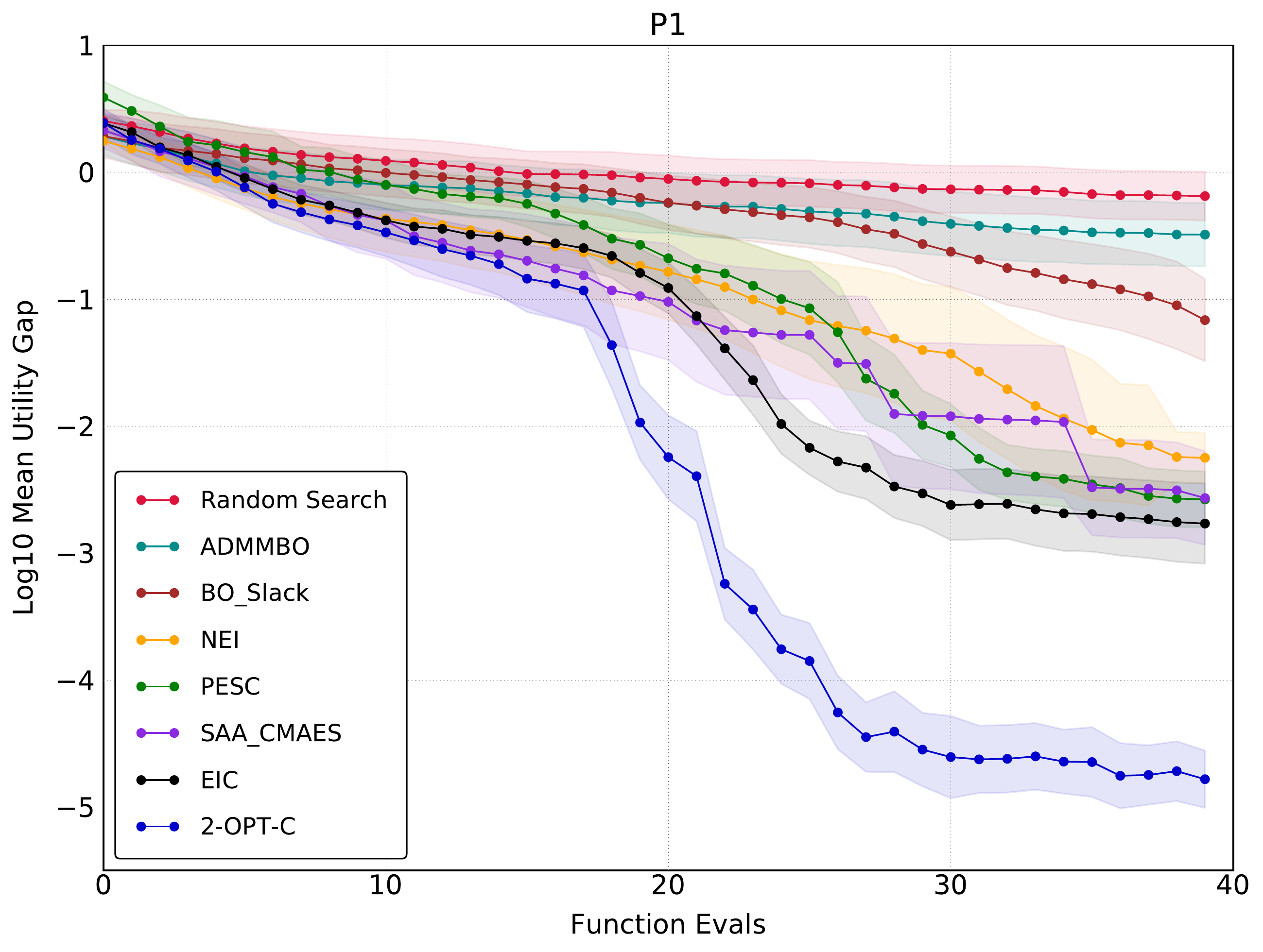}\label{marker}
\hspace{1em}
\includegraphics[scale = 0.22]{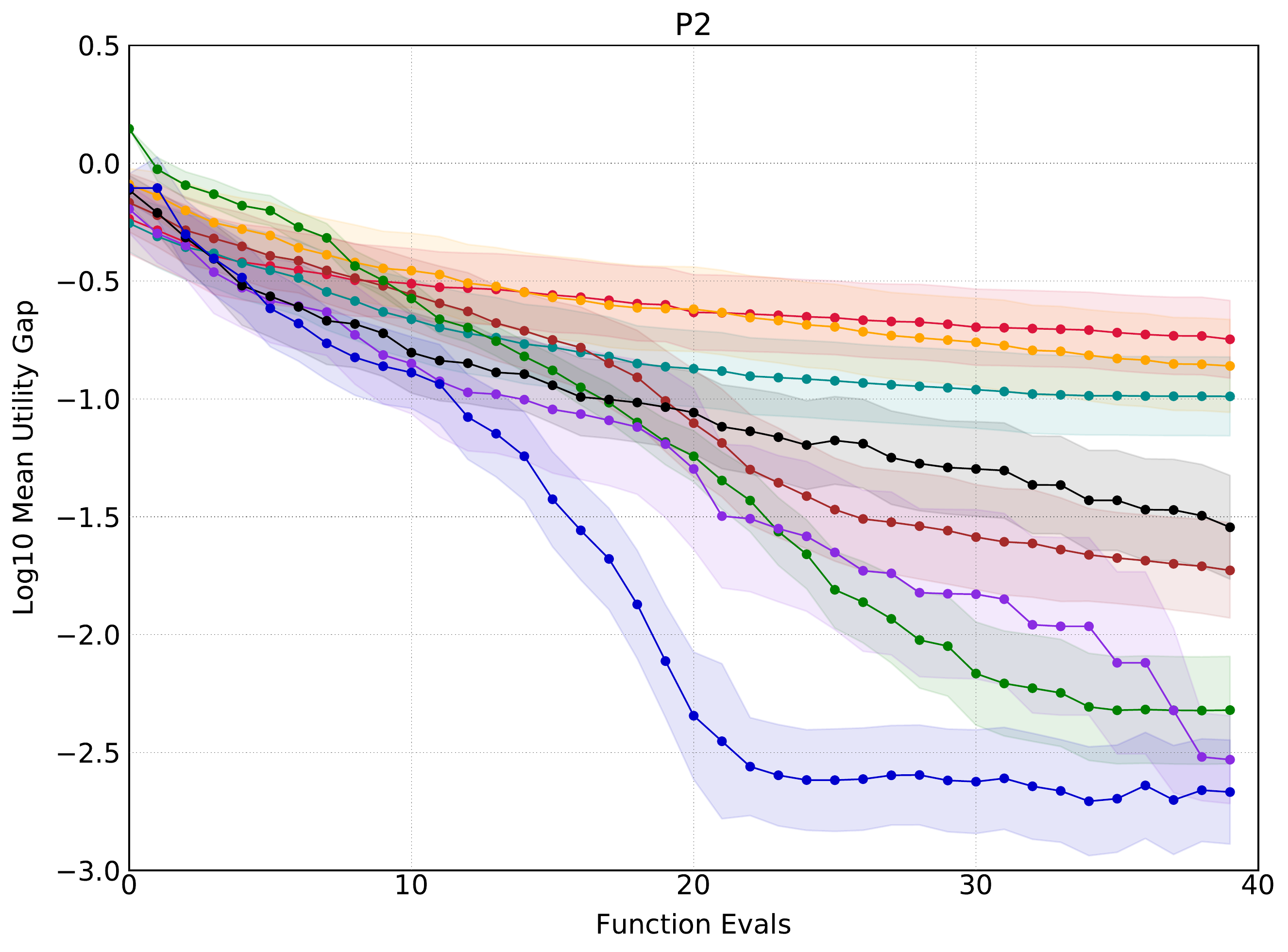}
\includegraphics[scale = 0.22]{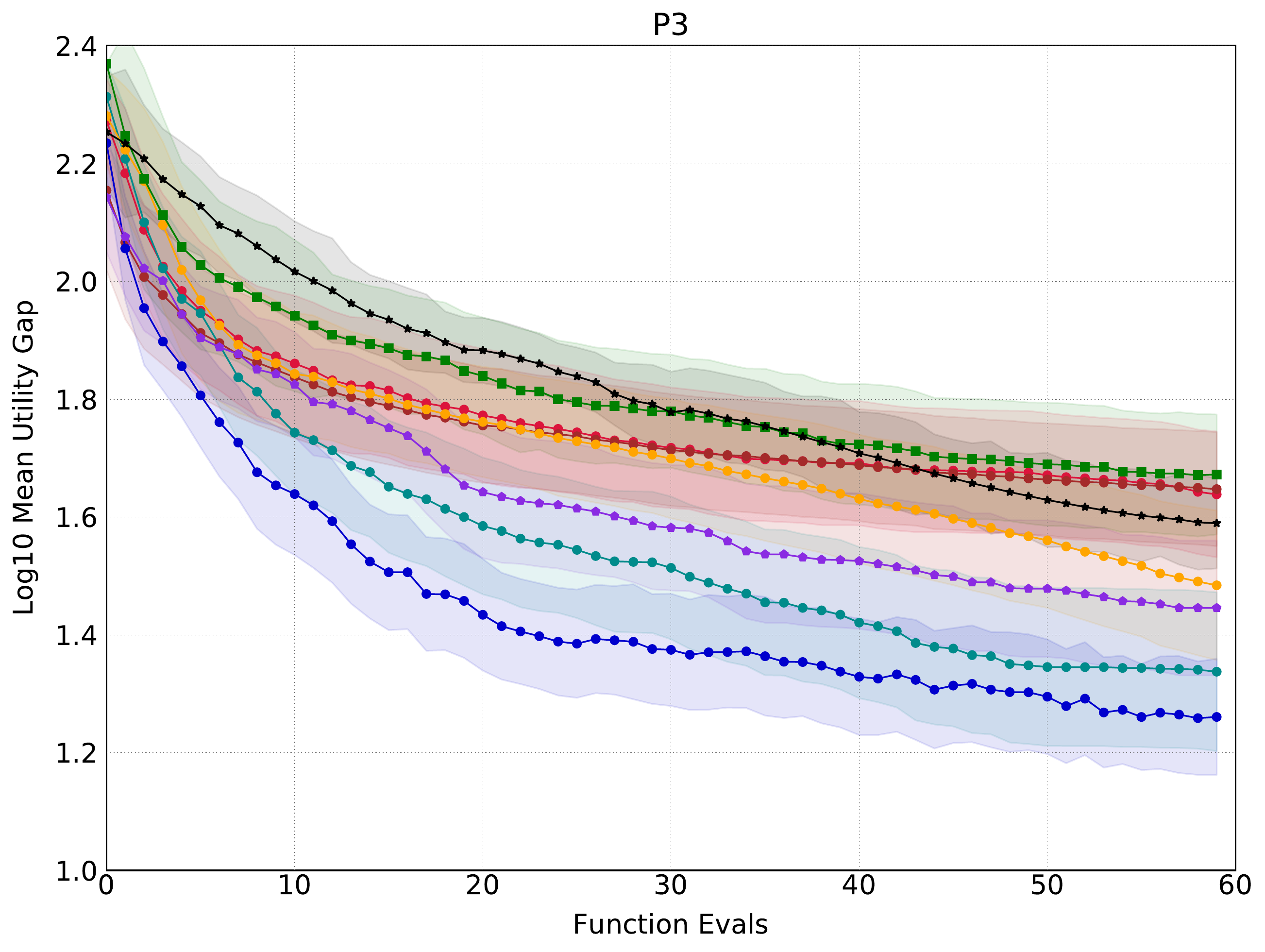}
\hspace{1em}
\includegraphics[scale = 0.22]{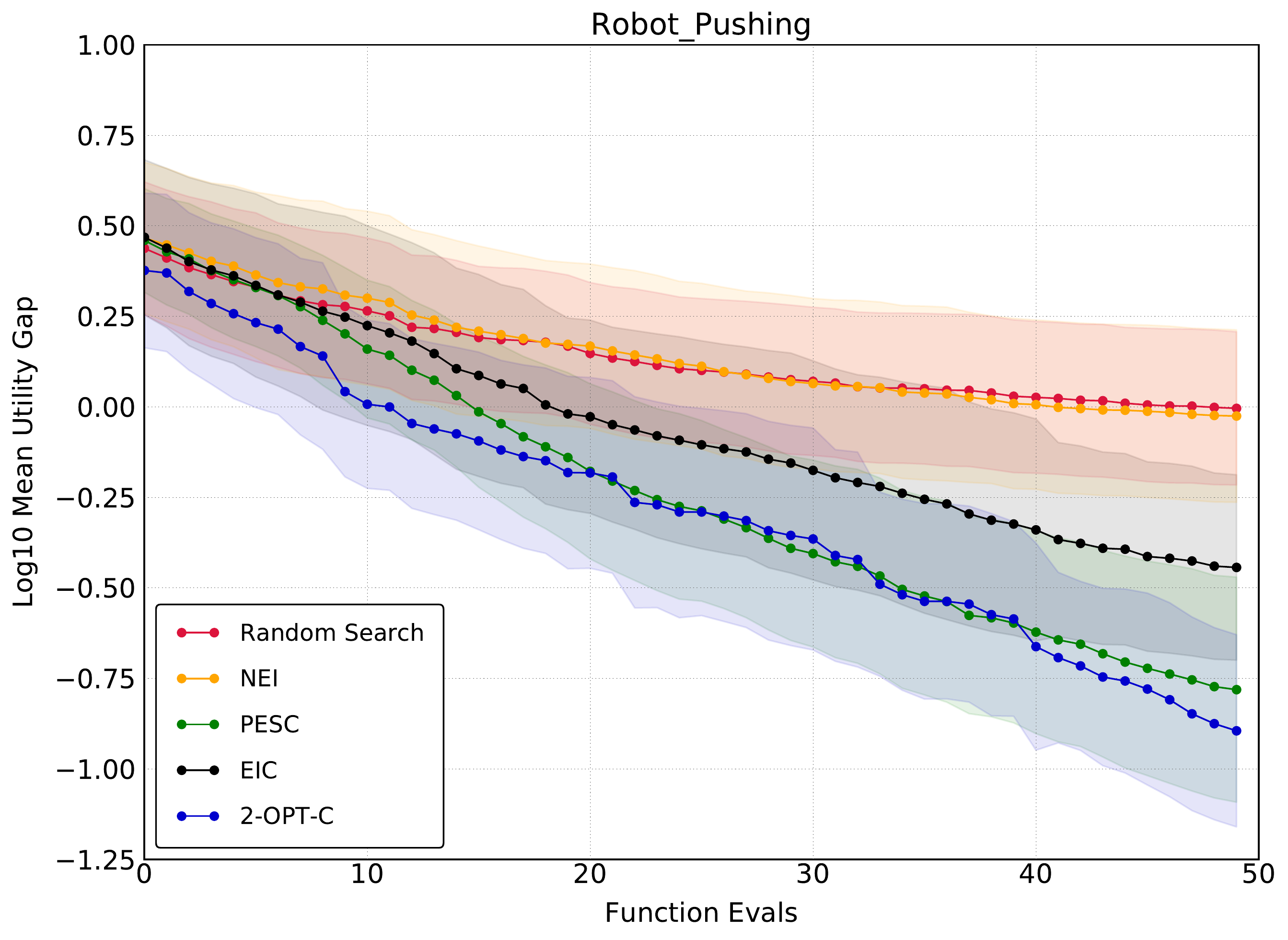}
\caption{Log10 mean utility gap of Random Search, ADMMBO, BO\_Slack, NEI, EIC, PESC, SAA\_CMAES, and $\TOPTC$ with 95\% confidence intervals for P1 (top left), P2 (top right), and P3 (bottom left). Log10 mean utility gap of Random Search, NEI, EIC, PESC, and $\TOPTC$ with 95\% confidence intervals for robot pushing (bottom right).
\label{fig:mean}
}
\label{icml-mean}
\end{center}
\end{figure}

\subsection{Experiment Results for the Mean Utility Gap}
\label{sec:mean}

We report the {\it mean} utility gap rather than the median, using the same experiments (except for the portfolio optimization problem) reported in \S\ref{sec:myopic}. As in \S\ref{sec:myopic}, we continue to assume sequential evaluations and plot versus the number of function evaluations. Figure~\ref{fig:mean} provides these additional results. Results for this alternative outcome, the mean, are very similar to those for the median.



\interfootnotelinepenalty=10000

\subsection{Additional Non-Myopic Results}
\label{sec:nonmyopic}
We provide the detailed experiment results of \S \ref{sec:non-myopic} in Table 1\footnote{The performance of EIC as reported in \cite{Willcox} differs from both our results and those in \cite{pesc}. Since the \cite{Willcox} EIC implementation leveraged complex rollout code, which may have special characteristics, we exclude those results.}. All the statistics are obtained from Table 1 of \cite{Willcox} and we add results from $\TOPTC$ to their table. 

\begin{table*}[htb]
\label{sample-table}
\vskip 0.15in
\begin{center}
\begin{small}
\begin{sc}
\setlength{\tabcolsep}{4.4pt}
\begin{tabular}{cccccccccccc}
\toprule
Problem & N  & SLSQP & MMA   & COBYLA & ISRES & PESC & PM    & R-1   & R-2   & R-3   & 2-OPTC \\
\midrule
P1   & 40 & 0.59  & 0.59  & -0.05  & -0.19 & -2.68 & 0.30  & -4.59 & -4.52 & -4.42 & \textbf{-4.92}       \\
P2   & 40 & -0.40 & -0.40 & -0.82  & -0.70 & -2.43 & -1.76 & -2.99 & -2.99 & -2.99 &    \textbf{-3.08}    \\
P3   & 60 & 2.15  & 3.06  & 3.06   & 1.68  & 1.66  & 1.79  & 1.48  & 1.31  & 1.35  &     \textbf{1.28}    \\ 
\bottomrule
\end{tabular}
\end{sc}
\end{small}
\end{center}
\caption{Log10 median utility gap. $\TOPTC$'s result is presented in the last column. All other statistics are obtained from \cite{Willcox}. The results of the non-myopic rollout method with different horizons are shown in the three columns before $\TOPTC$. }
\label{two_step_table}
\vskip -0.1in
\end{table*}

\section{\texorpdfstring{$\TOPTC$}{Lg} with Multiple Constraints}\label{sec:C}


Our approach can also be easily generalized to the case with multiple constraints. Here, we provide the form of the gradient estimator $\Gamma(X_1, \Xtilde_1, Y)$ with multiple constraints. We focus on independent constraints for simplicity. Our discussion here can be easily generalized to the dependent case, where the multivariate Gaussian probability distribution over the constraints can be calculated through numerical methods \cite{cunningham}.  

Suppose we have $M$ constraints ($M$ can be 1). Let $Y = (Y_f, \{Y_{g,m}\}_{1\leq m \leq M})$, where $Y_{g,m}$ is the $m$-th constraint function evaluated at $X_1$. Then similar to \S\ref{sec:two_step_def}, if we use $p(y; X_1)$ to denote the distribution of $Y$, i.e., $Y \sim p(y; X_1)$, then 
\begin{equation*}
p(y; X_1) =  \mathcal{N}\left(
\begin{bmatrix}
     \mu_0(X_1)\\[0.5em]
    \mu_{0,1}^c(X_1) \\[0.5em]
    \mu_{0,2}^c(X_1) \\[0.5em]
    \vdots \\[0.5em]
    \mu_{0,M}^c(X_1) \\[0.5em]
  \end{bmatrix}, 
  \begin{bmatrix}
    K_0(X_1, X_1) & 0  & 0 & \ldots  & 0 \\[0.5em]
    0 & K_{0,1}^c(X_1, X_1)  & 0 &\ldots  & 0 \\[0.5em]
    0 & 0 & K_{0,2}^c(X_1, X_1)  &\ldots  & 0 \\[0.5em]
    \vdots & \vdots & \vdots &\ddots  & \vdots \\[0.5em]
    0 & 0 & \ldots & \ldots& K_{0, M}^c(X_1, X_1) 
  \end{bmatrix}
  \right)
\end{equation*}

where $\mu_{0, m}^c$ and $K_{0, m}^c$ are the posterior mean and posterior variance of the corresponding Gaussian process modeling the $m$-th constraint function for $m = \{1,2,\cdots, M \}$ respectively. In other words, $p(y; X_1)$ now is the joint posterior distribution of $Y_f$ and $\{Y_{g,m}\}_{1\leq m \leq M}$ given the information of the past observations $D$ and their objective $f(D)$ and constraint $g_m(D)$, where $g_m$ is the $m$-th constraint function. Let $p(y; \widetilde{X}_1)$ be the importance sampling distribution defined in a similar way to the main paper but under the multiple constraint setting. Then our gradient estimator has the same form as the one in \S \ref{sec:computation}:
\begin{align*}
    \Gamma(X_1, \Xtilde_1, y) 
    := &\nabla \max_{x_2 \in A({\delta})}\alpha (X_1, x_2, y)L(y; X_1, \Xtilde_1)\\
    = &\left[\nabla \alpha (X_1, x_2^{*}, Y)\right]L(y; X_1, \Xtilde_1) + \alpha (X_1, x_2^{*}, Y)\left[\nabla p(y; X_1)\right] / p(y;\widetilde{X}_1)
\end{align*}
with $x_2^* \in \argmax_{x_2 \in A(\delta)} \alpha (X_1, x_2, Y)$. Here, $\alpha (X_1, x_2, y)$ is under the multiple constraint setting, which is explicitly written as:
\begin{equation*}
    \alpha (X_1, x_2, y) 
    = f_0^* - f_1^*  + \text{EI}(f_1^{*} - \mu_1(x_2), \sigma_1(x_2)^2) \prod_{m = 1}^{M} \text{PF}(\mu_{1,m}^{c}(x_2), (\sigma_{1,m}^{c}(x_2))^2)
\end{equation*}
where $f^{*}_0$ is the best evaluated point satisfying all the constraints so far, i.e. $f^{*}_0 = \min\{f(x) : x \in D, g_m(x) \leq 0 \text{ for } 1 \leq m \leq M\}$. Similarly, $f^{*}_1$ and $f^*_2$ are the best evaluated points satisfying all the constraints by the end of the first and second stage respectively. Moreover, $\mu_1(x)$, $\mu_{1,m}^c(x)$ and $\sigma_1(x)$, $\sigma_{1,m}^c(x)$ denote the posterior mean and posterior standard deviation of $f(x)$ and $g(x)$ given D and $X_1$.

\section{\texorpdfstring{$\TOPTC$}{Lg} Algorithm}\label{sec:algo}
The overall description of the algorithm is provided in \S \ref{sec:computation}. Here, we provide implementation details of our algorithm and present pseudo-code for $\TOPTC$. This pseudo-code sets $\widetilde{X}_1$ equal to $X_1$.

To reduce the variance of $\TOPTC$, instead of sampling the fantasy values for $Y_f = \{ f(x): x\in X_1\}$ and $Y_g = \{ g(x): x\in X_1\}$ through classic Monte Carlo (MC) methods, we apply the quasi Monte Carlo technique (QMC) \cite{qmc}, which uses a low-discrepancy sequence as its sample set. In classic MC, the sampled values tend to cluster even when the sampling distribution is uniform. Consequently, the sampled values may not cover the domain of integration particularly efficiently. However, QMC covers the domain of integration more efficiently and thoroughly. Therefore, QMC provides more robust performance using fewer samples compared to classic MC.

In addition, to reduce the computational burden of $\TOPTC$, we leverage the two time scale optimization technique in \cite{bo_risk}. The key idea of this technique is that for the nested optimization problem, the optimal solution (in our case, its $x_2^*$) obtained from the previous iteration of the inner problem should remain within a small neighborhood of a current high quality local optimal solution (or the global solution). In other words, we do not need to solve the inner optimization problem when we calculate every gradient copy for our gradient. Instead, for certain gradient copies, we can use the optimal solution previously obtained and directly calculate the gradient copy. We let $k$ denote the frequency of solving the inner optimization. $k = 2$ means that we solve the inner optimization on every other iteration when calculating the gradient.

Let subscripts $r,t,m$ denote the number of restarts, the number of gradient ascent steps, and the number of samples of the gradient. For simplicity, we drop the subscript representing the stage and use $X$ to denote $X_1$. With this notation we summarize our full algorithm for optimizing $\TOPTC$ as follows. 

\begin{algorithm}[H]
\caption{Estimating $\nabla \TOPTC(X)$ with Two Time Scale Optimization}
  \begin{algorithmic}[1]
\STATE  Initialize $\nabla \TOPTC(X) = 0$ 
\STATE Initialize $x_{2, \text{previous}}^* = 0$
      \FOR {m = 1:M}
         \STATE Use QMC sampling to get $Y_f$ and $Y_g$
         \IF {$m \neq k$} 
          \STATE Obtain $x_2^*$ by solving $\argmax_{x_2} \alpha (X, x_2, y)$
          
          \ELSE
            \STATE $x_2^* = x_{2, \text{previous}}^*$ 
          \ENDIF
          
          \STATE  Plug $x_2^*$ into $\Gamma(X, X, y)$ and compute the $m$-th gradient estimation: $\widehat{\nabla \TOPTC}(X)_{m}$.
    \ENDFOR
    \STATE return the estimated gradient $\widehat{\nabla \TOPTC}(X) = \frac{1}{M} \sum_{m = 1}^M \widehat{\nabla \TOPTC}(X)_{m}$
  \end{algorithmic}
\end{algorithm}

\begin{algorithm}[H]
  \caption{Optimization of $\TOPTC$}
  \begin{algorithmic}[1]
   \FOR{r = 1:R} 
     \STATE Draw an initial point or a batch of points $X_{r,0}$ from the Latin hypercube
     \FOR{t = 0:T-1}
        \STATE Estimate $\nabla \TOPTC(X_{r,t})$ using Algorithm 1 and store the result as $G(X_{r,t})$
        \STATE Update solution $X_{r,t+1} = (1 - \alpha_t)X_{r,t} + \alpha_t G(X_{r,t})$, where $\alpha_t$ is the step size at time t
     \ENDFOR
    \STATE Estimate $\TOPTC(X_{r, T})$ using Monte Carlo simulation and store the estimate as $\widehat{\TOPTC}(X_{r,T})$
   \ENDFOR
   \STATE return $X_{r^*,T}$ and $\widehat{\TOPTC}(X_{r^*,T})$, where $r^* = \max_r \widehat{\TOPTC}(X_{r,T})$
  \end{algorithmic}
\end{algorithm}

\section{Proof of Theorem 1}\label{sec:proof}
This section proves Theorem 1.
We first restate the theorem and detail the regularity conditions 
(Assumption~\ref{as:1}) 
it requires.

\begin{assumption}\label{as:1}
    We assume
    \begin{enumerate}
        \item\label{as:1-1} The prior on the objective function $f$ is a Gaussian Process $f \sim GP(\mu_f(x), K_f(x, x'))$, and the prior on the constraints $g$ is another Gaussian Process $g \sim GP(\mu_g(x), K_g(x, x'))$. These two Gaussian processes are independent.
        \item\label{as:1-2} $\mu_f$ and $\mu_g$ is continuously differentiable with $x$.
        \item\label{as:1-3} $K_f(x, x')$ and $K_g(x, x')$ is continuously differentiable with $x$ and $x'$.
        \item\label{as:1-4} Given $n$ different points $X = (x_1, x_2, ..., x_n)$, the matrix $K_f(X, X)$ and $K_g(X, X)$ are of full rank.
    \end{enumerate}
\end{assumption}

Assumption \ref{as:1}.\ref{as:1-1} models the objective $f$ and constraints $g$ as two independent Gaussian processes, following \cite{Gardner14}. Assumption \ref{as:1}.\ref{as:1-2} and \ref{as:1}.\ref{as:1-3} assumes the mean and the kernel functions of the two Gaussian processes are smooth. Assumption \ref{as:1}.\ref{as:1-4} assumes non-degeneracy of the kernel matrix, i.e., the posterior variance at a point is zero if and only if this point is already sampled. All the assumptions are standard in the Bayesian Optimization literature \cite{BO_tutorial}.

We now restate Theorem 1 before providing the proof: under Assumption 1, $\TOPTC(X_1)$'s partial derivatives exist almost everywhere for any $\delta > 0$. 
When $\TOPTC(X_1)$ is differentiable,
\begin{align*}
\nabla \TOPTC(X_1)
    = \int \Gamma(X_1, \Xtilde_1, y)
    p(y; \Xtilde_1) dy.
\end{align*}

To support the proof, we first define some notation. First, we use $A^c_\delta := A \backslash A_c$ to denote the complementary set of $A_\delta$; clearly, $A^c_\delta$ is an open set. Second, we use $d(X_1, \Xtilde_1) := \max_{1 \le i \le q} |x^{(i)} - \tilde{x}^{(i)}|$ where $X_1 = ( x^{(1)}, x^{(2)}, ..., x^{(q)}) $ and $\Xtilde_1 = (\tilde{x}^{(1)}, \tilde{x}^{(2)}, ..., \tilde{x}^{(q)})$ are two sets of batched points. $d(X_1, \Xtilde_1)$ measure the similarity of $X_1$ and $\Xtilde_1$, and $d(X_1, \Xtilde_1) = 0$ if and only if $X_1 = \Xtilde_1$. Finally, we define
\begin{align*}
    \TOPTC_{\delta}(X_1, \Xtilde, y) := \max_{x_2 \in A_\delta} \alpha(X_1, x_2, y) L(y; X_1, \Xtilde).
\end{align*}
With this new notation, 
\begin{align*}
    \TOPTC(X_1) = \int \TOPTC_\delta(X_1, \Xtilde, y) p(y; \Xtilde) dy.
\end{align*}

Before diving into the proof of Theorem 1, we give an overview of all the lemmas needed and their roles in the proof. Lemma \ref{le:0} bounds the posterior mean and posterior standard deviation of the objective and the constraints functions, which serves a cornerstone to bound $\nabla_{X_1} \TOPTC_{\delta}(X_1, \Xtilde, y)$ (Lemma 2). Lemma \ref{le:2} shows $\TOPTC_{\delta}(X_1, \Xtilde, y)$ is absolutely continuous with respect to $X_1$, thus combining Lemma \ref{le:1} and Lemma \ref{le:2} shows $\nabla_{X_1} \TOPTC_{\delta}(X_1, \Xtilde, y)$ is joint integratable with respect to $X_1 \times y$. Applying Fubini Theorem to interchange the integration of $X_1$ and $y$ proves Theorem 1.

\begin{lemma}\label{le:0}
Given the set of sampled points $D$, batched points $\Xtilde_1$ and $\delta' < \delta$, there exists constant $C_0, C_1$ and $C_2$, s.t. for any $y$, $x_2 \in A_\delta$ and $X_1$ s.t. $d(X_1, \Xtilde_1) \le \delta'$, the posterior mean and standard deviation of the objective $f$ and constraints $g$ at $x_2$ satisfy:
\begin{itemize}
    \item $|\mu_1(x_2)| \le C_1 + C_2 |y|$, $|\mu_1^c(x_2)| \le C_1 + C_2 |y|$;
    \item $|\nabla_{X_1} \mu_1(x_2)| \le C_1 + C_2 |y|$, $|\nabla_{X_1} \mu_1^c(x_2)| \le C_1 + C_2 |y|$;
    \item $C_0 \le |\sigma_1(x_2)| \le C_1$, $C_0 \le |\sigma_1^c(x_2)| \le C_1$;
    \item $|\nabla_{X_1} \sigma_1(x_2)| \le C_1$, $|\nabla_{X_1} \sigma_1^c(x_2)| \le C_1$.
\end{itemize}
When there are multiple constraints, $\sigma_1^c(x_2)$ is understood as the upper triangle matrix by Cholesky decomposition of the covariance matrix at $x_2$.
\end{lemma}
\textit{Proof of Lemma \ref{le:0}}
Without loss of generality, we assume the set of sampled points is an empty set. Otherwise, we could change the definition of $\mu_f$, $\mu_g$ and $K_f, K_g$ to be the posterior mean and kernel function conditioned on the sampled data $D$.

We only prove the inequalities considering the objective function $f$. The proof for inequalities considering the constraints functions $g$ is the same.

Conditioned on objective function $y_f: = f(X_1)$ evaluated at $X_1$, the posterior of $f(x_2)$ is a Gaussian distribution with mean $\mu(x_2)$ and variance $\sigma^2(x_2)$, where
\begin{align*}
\mu_1(x_2) &:= \mu_f(x_2) + K_f(x_2, X_1)K(X_1, X_1)^{-1}(y_f - \mu_f(X_1))\\
\sigma_1^2(x_2) &:= K(x_2,x_2) - K(x_2, X_1)K(X_1, X_1)^{-1}K(X_1, x_2).
\end{align*}

The bound on the posterior mean $\mu_1(x_2)$ follows from
\begin{align*}
    |\mu_1(x_2)|
    &= |\mu_f(x_2) + K_f(x_2, X_1)K(X_1, X_1)^{-1}(y_f - \mu_f(X_1))| \\
    &\le \max_{x_2 \in A_{\delta}} |\mu_f(x_2)|\\
    & \quad + \max_{x_2 \in A_{\delta}, d(X_1, X'_1)  \delta'}|K_f(x_2, X_1)| \max_{d(X_1, X'_1)  \delta'} |K(X_1, X_1)^{-1}| |y_f| \\
    & \quad + \max_{x_2 \in A_{\delta}, d(X_1, X'_1)  \delta'}|K_f(x_2, X_1)| \max_{d(X_1, X'_1) \le \delta'} |K(X_1, X_1)^{-1}| \max_{d(X_1, X'_1) \le \delta'}|\mu_f(X_1)|,
\end{align*}
where $\max_{x_2 \in A_{\delta}} |\mu_f(x_2)| < \infty$ by the smoothness of $\mu_f$ and compactness of $A_\delta$. Similarly,
\begin{align*}
    \max_{x_2 \in A_{\delta}, d(X_1, X'_1)  \delta'}|K_f(x_2, X_1)| < \infty,  \max_{d(X_1, X'_1)  \delta'} |K(X_1, X_1)^{-1}| < \infty, \max_{d(X_1, X'_1) \le \delta'}|\mu_f(X_1)| < \infty,
\end{align*}
follows from the compactness of the region $\{X_1: d(X_1, X'_1) \le \delta'\}$, the smoothness of $K_f$ and $\mu_f$, and the full-rank property of $K_f$.

The bound on $\nabla_{X_1} \mu_1(x_2)$, $\sigma_1(x_2)$ and $\nabla_{X_1} \sigma_1(x_2)$ can be deduced by following the same spirit.
\hfill$\Box$

\begin{lemma}\label{le:1}
Given the set of sampled points $D$, batched points $\Xtilde_1$ and $\delta' < \delta$, there exists constant $C_1$ and $C_2$, s.t. for any $y$, $x_2 \in A_\delta$ and $X_1$ s.t. $d(X_1, \Xtilde_1) \le \delta'$,
\begin{itemize}
    \item $|\alpha(X_1, x_2, y)| \le C_1 + C_2 |y|$,
    \item $|\nabla_{X_1} \alpha(X_1, x_2, y)| \le C_1 + C_2 |y|$,
    \item $|\nabla_{X_1} L(y; X_1, \Xtilde_1)| \le (C_1 + C_2 |y|^2) |L(y; X_1, \Xtilde_1)|$.
\end{itemize}
\end{lemma}

\textit{Proof of Lemma \ref{le:1}:}
To prove the first inequality, 
\begin{align*}
    |\alpha(X_1, x_2, y)|
    &= |f_0^* - f_1^* + \mathbb{E}_1[(f_1^* - f(x_2))^+ I(g(x_2) \le 0)]| \\
    &\le |f_0^*| + |f_1^*| + |\mathbb{E}_1[(f_1^* - f(x_2))^+ I(g(x_2) \le 0)]|\\
    &\le |f_0^*| + |y| + \mathbb{E}_1[|f_1^* - f(x_2)|]\\
    &\le |f_0^*| + |y| + |f_1^*| + \mathbb{E}_1[|f(x_2)|]\\
    &\le |f_0^*| + |y| + |y| + |\mu_1(x_2)| + |\sigma_1(x_2)|\\
    &\le C_1 + C_2 |y|,
\end{align*}
where applying Lemma \ref{le:0} to $\mu_1(x_2)$ and $\sigma_1(x_2)$ derives the last inequality.

To prove the second inequality, notice
\begin{align*}
    \alpha (X_1, x_2, Y) 
    = f_0^* - f_1^*  + &\text{EI}(f_1^{*} - \mu_1(x_2), \sigma_1(x_2)^2) 
    \text{PF}(\mu_1(x_2)^c, (\sigma_1(x_2)^c)^2),
\end{align*}
where $\text{EI}(m, v) :=m \Phi(m / \sqrt{v})+\sqrt{v} \varphi(m / \sqrt{v})$ and $\text{PF}(m^c, v^c) = \Phi\left(-m^c / \sqrt{v^c}\right)$. Notice functions $\Phi, \phi$ along with their derivatives $\Phi', \phi'$ are bounded. Thus, applying the chain rule along with Lemma \ref{le:0} proves the second inequality. 

To prove the third inequality, notice
\begin{align*}
    L(y; X_1, \Xtilde_1) = \frac{p(y; X_1)}{p(y; \Xtilde_1)}
\end{align*}
where 
\begin{equation*}
p(y; X_1) =  \mathcal{N}\Bigg{(}
\begin{bmatrix}
     \mu_0(X_1)\\[0.5em]
    \mu_0^c(X_1) 
  \end{bmatrix}, 
  \begin{bmatrix}
    \sigma_0^2(X_1) & 0  \\[0.5em]
    0 & (\sigma_0^c(X_1))^2
  \end{bmatrix}
  \Bigg{)}.
\end{equation*}
By following an approach similar to the second inequality's proof, 
it is straightforward to prove the third inequality.
\hfill $\Box$

\begin{lemma}\label{le:2}
Given $\Xtilde$, suppose $X_{1, 1}$ and $X_{1, 2}$ satisfy $d(\Xtilde_1, X_{1, 1}) < \delta$ and $d(\Xtilde_1, X_{1, 1}) < \delta$. We can construct the path map $\epsilon \rightarrow X_1(\epsilon)$, where $\epsilon \in [0, 1]$ and
$$X_1(\epsilon) = \epsilon X_{1, 1} + (1 - \epsilon) X_{1, 2}.$$ 
Then $\TOPTC_{\delta}(X_1(\epsilon), \Xtilde_1, y)$ is absolutely continuous with $X_1(\epsilon)$, i.e.
\begin{align*}
    \TOPTC_{\delta}(X_1(1), \Xtilde_1, y) = \TOPTC_{\delta}(X_1(0), \Xtilde_1, y) + \int_0^1 \frac{d}{d \epsilon} \TOPTC_{\delta}(X_1(\epsilon), \Xtilde_1, y) d \epsilon.
\end{align*}
In particular, 
\begin{align*}
    \frac{d}{d \epsilon} \TOPTC_{\delta}(X_1(\epsilon), \Xtilde_1, y) \left|_{\epsilon = \epsilon'} = \frac{d}{d \epsilon} \left[ \alpha(X_1(\epsilon), x_2(\epsilon'), y) L(y; X_1(\epsilon), \Xtilde_1)\right] \right|_{\epsilon = \epsilon'}
\end{align*}
where $x_2(\epsilon')= \argmax_{x_2} \ \alpha(X_1(\epsilon'), x_2, y) L(y; X_1(\epsilon'), \Xtilde_1)$.
\end{lemma}

\textit{Proof of Lemma \ref{le:2}}
According to Corollary 4 in \cite{envel}, we only need to check the following three conditions:
\begin{itemize}
    \item $A_\delta$ is non-empty compact set.
    \item $\alpha(X_1(\epsilon), x_2, y) L(y; X_1(\epsilon), \Xtilde_1)$ is continuous in $x_2$.
    \item $\frac{d}{d \epsilon} \left[ \alpha(X_1(\epsilon), x_2, y) L(y; X_1(\epsilon), \Xtilde_1)\right]$ is continuous in $\epsilon$ and $x_2$.
\end{itemize}

The first condition holds true according to the definition of $A_{\delta}$.
Because the posterior mean and variance is a continuously differentiable function of $X_1$, the second and the third condition can be verified with the chain rule. 
\hfill$\Box$

\begin{lemma}\label{le:3}
Given $\Xtilde$, suppose $X_{1, 1}$ and $X_{1, 2}$ satisfy $d(\Xtilde_1, X_{1, 1}) < \delta$ and $d(\Xtilde_1, X_{1, 1}) < \delta$. We can construct the path map $\epsilon \rightarrow X_1(\epsilon)$, where $\epsilon \in [0, 1]$ and
$$X_1(\epsilon) = \epsilon X_{1, 1} + (1 - \epsilon) X_{1, 2}.$$

Then
\begin{align*}
    \frac{d}{d \epsilon} \TOPTC_{\delta}(X_1(\epsilon),  \Xtilde_1, y)
\end{align*}
is integrable in the product region $[0, 1] \times \textnormal{Range}(y)$ with the product measure, where the measure on $[0, 1]$ is given by the Lebesgue measure and the measure on $\textnormal{Range}(y)$ is given by the Gaussian density function
\begin{equation*}
p(y; \Xtilde_1) =  \mathcal{N}\Bigg{(}
\begin{bmatrix}
     \mu_0(\Xtilde_1)\\[0.5em]
    \mu_0^c(\Xtilde_1) 
  \end{bmatrix}, 
  \begin{bmatrix}
    \sigma_0^2(\Xtilde_1) & 0  \\[0.5em]
    0 & (\sigma_0^c(\Xtilde_1))^2
  \end{bmatrix}
  \Bigg{)}.
\end{equation*}
\end{lemma}

\textit{Proof of Lemma \ref{le:3}:}
According to Lemma \ref{le:2},
\begin{align*}
    \frac{d}{d \epsilon} \TOPTC_{\delta}(X_1(\epsilon), \Xtilde_1, y) \big{|}_{\epsilon = \epsilon'} 
    &= \frac{d}{d \epsilon} \left[ \alpha(X_1(\epsilon), x_2(\epsilon'), y) L(y; X_1(\epsilon), \Xtilde_1)\right] \Big{|}_{\epsilon = \epsilon'}\\
    &= [\frac{d}{d \epsilon} \alpha(X_1(\epsilon), x_2(\epsilon'), y)] L(y; X_1(\epsilon), \Xtilde_1) + \alpha(X_1(\epsilon), x_2(\epsilon'), y) \frac{d}{d \epsilon} L(y; X_1(\epsilon), \Xtilde_1)
\end{align*}

According to Lemma \ref{le:1}, there exists $C_1$ and $C_2$, s.t.

\begin{align*}
    |\frac{d}{d \epsilon} \alpha(X_1(\epsilon), x_2(\epsilon'), y)| 
    &\le C_1 + C_2 |y|,\\
    \alpha(X_1(\epsilon), x_2(\epsilon'), y) &\le C_1 + C_2 |y|,\\
    \frac{d}{d \epsilon} L(y; X_1(\epsilon), \Xtilde_1) &\le (C_1 + C_2 |y|^2) L(y; X_1(\epsilon), \Xtilde_1).
\end{align*}

So, there exists constant $C_1$ and $C_2$, s.t. 
\begin{align*}
    \frac{d}{d \epsilon} \TOPTC_{\delta}(X_1(\epsilon), \Xtilde_1, y)
    &\le (C_1 + C_2 |y|^3) L(y; X_1(\epsilon), \Xtilde_1).
\end{align*}
Thus, Lemma \ref{le:3} is proved. \hfill$\Box$

Now we are ready to prove Theorem 1.

\textit{Proof of Theorem 1}
Without loss of generality, we assume $d(X_1, \Xtilde_1) \le \frac{\delta}{2}$. Other cases can be proved in a straightforward fashion following our approach here.

When calculating a partial derivative at $X_1$, we choose a point $X'_1$, s.t. $d(X'_1, \Xtilde_1) \le \frac{\delta}{2}$ and $X'_1 - X_1$ gives the direction in which we are interested. Construct the path map
$X_1(\epsilon) = \epsilon X_{1} + (1 - \epsilon) X'_{1}$
where $\epsilon \in [0, 1]$.

According to Lemma \ref{le:2},
$$\TOPTC_{\delta}(X_1(\epsilon), \Xtilde_1, y) = \TOPTC_{\delta}(X_1(0), \Xtilde_1, y) + \int_0^\epsilon \frac{d}{d \epsilon} \TOPTC_{\delta}(X_1(\epsilon), \Xtilde_1, y) d \epsilon.$$

Thus, 
\begin{align*}
    \TOPTC(X_1(\epsilon))
    &= \int \TOPTC_{\delta}(X_1(\epsilon), \Xtilde_1, y)
    p(y; \Xtilde_1) dy\\
    &= \int \TOPTC_{\delta}(X_1(0), \Xtilde_1, y) p(y; \Xtilde_1) dy + \int p(y; \Xtilde_1) dy \int_0^{\epsilon} \frac{d}{d \epsilon} \TOPTC_{\delta}(X_1(\epsilon), \Xtilde_1, y) d \epsilon.
\end{align*}

According to Lemma \ref{le:3}, $\frac{d}{d \epsilon} \TOPTC_{\delta}(X_1(\epsilon), \Xtilde_1, y)$ is integratable with respect to $\epsilon \times y$. Thus, by Fubini Theorem,
\begin{align*}
    \TOPTC(X_1(\epsilon)) 
    &= \int \TOPTC_{\delta}(X_1(0), \Xtilde_1, y) p(y; \Xtilde_1) dy
    + \int p(y; \Xtilde_1) dy \int_0^{\epsilon} \frac{d}{d \epsilon} \TOPTC_{\delta}(X_1(\epsilon), \Xtilde_1, y) d \epsilon\\
    &= \int \TOPTC_{\delta}(X_1(0), \Xtilde_1, y) p(y; \Xtilde_1) dy
    + \int_0^{\epsilon} d\epsilon \int \frac{d}{d \epsilon} \TOPTC_{\delta}(X_1(\epsilon), \Xtilde_1, y) p(y; \Xtilde_1) dy.
\end{align*}

Thus,
\begin{align*}
    \frac{d}{d \epsilon} \TOPTC(X_1(\epsilon)) 
    &= \int \frac{d}{d \epsilon} \TOPTC_{\delta}(X_1(\epsilon), \Xtilde_1, y) p(y; \Xtilde_1) dy.
\end{align*}

When $\TOPTC(X_1)$ is differentiable with respect to $X_1$,
\begin{align*}
    \nabla \TOPTC(X_1) 
    &= \int \nabla\TOPTC_{\delta}(X_1, \Xtilde_1, y) p(y; \Xtilde_1) dy\\
    &= \int \Gamma(X_1, \Xtilde_1, y) p(y; X_1) dy.
\end{align*}

\section{Constrained Multi-points Expected Improvement}\label{sec:B}

Here, we formally derive the form for the constrained multi-points expected improvement defined in \S \ref{sec:EIC_def}, which allows multiple points to be evaluated at each iteration. This idea is first discussed in \cite{NEI} but not formally defined.

Let $\mathbf{X} = \{x^{(n+1)}, x^{(n+2)}, \ldots, x^{(n+q)}\}$ be the set of q candidate points that we consider evaluating next. Let $f^*$ be the best evaluated point subject to the constraint. Then the improvement provided by those points subject to the constraint, $I(\mathbf{X})$, can be written as:
\begin{align}
    I(\mathbf{X}) = &f^* - \min\left\{f^*, \min \{ f(x) : x \in \mathbf{X},  g(x) \leq 0 \} \right\}\\[0.25em]
    = &\max\left\{0, f^* - \min \{ f(x) : x \in \mathbf{X}, g(x) \leq 0 \} \right\}\\[0.25em]
    = &\max\left\{0, \max \{ f^* - f(x) : x \in \mathbf{X}, g(x) \leq 0 \} \right\}\\[0.25em]
     = & \left[\max \{ f^* - f(x) : x \in \mathbf{X}, g(x) \leq 0 \}\right]^+\\[0.25em]
    = & \max_{x \in \mathbf{X}} \hspace{0.5em} (f^* - f(x))^+ \cdot \mathbbm{1}\{g(x) \leq 0 \} 
\end{align}


where $\mathbbm{1}\{A\}$ is the indicator function that equals 1 if $A$ is true and 0 elsewhere. The last equality holds true for the following reason. If for any $x \in \mathbf{X}$, $g(x) > 0$, the inner maximization in (4) will be taken over an empty set. In other words, since any point in $\mathbf{X}$ does not satisfy the constraint, there should be no improvement in $\mathbf{X}$. Therefore, the result in (4) should be 0, which is the same as that in (5). In addition, consider the case where there exists at least one $x\in X_{1}$ such that $g(x) \leq 0$ and $f^* > f(x)$, then:
\begin{align*}
    I(\mathbf{X}) = & \left[\max \{ f^* - f(x) : x \in \mathbf{X}, g(x) \leq 0 \}\right]^+\\[0.25em]
    =& \max \{ f^* - f(x) : x \in \mathbf{X}, g(x) \leq 0 \} \\[0.25em]
    = &\max_{x \in \mathbf{X}} \hspace{0.5em} (f^* - f(x)) \cdot \mathbbm{1}\{g(x) \leq 0 \} \\[0.25em]
    = &\max_{x \in \mathbf{X}} \hspace{0.5em} (f^* - f(x))^+ \cdot \mathbbm{1}\{g(x) \leq 0 \} 
\end{align*}
Thus, the last equality holds true and we call $I(\mathbf{X}) = \max_{x \in \mathbf{X}} \hspace{0.5em} (f^* - f(x))^+ \cdot \mathbbm{1}\{g(x) \leq 0 \}$ the \textit{constrained multi-points improvement}. 

Then we define the \textit{constrained multi-points expected improvement} $\text{EIC}(\mathbf{X})$ as the expectation of the constrained multi-points improvement:
\begin{equation*}
    \text{EIC}(\mathbf{X}) := \mathbb{E}\left[ I(\mathbf{X}) \right] = \mathbb{E} \left[\max_{x \in \mathbf{X}} \hspace{0.5em}(f^* - f(x))^+ \cdot \mathbbm{1}\{g(x) \leq 0 \} \right]
\end{equation*}
where $\mathbb{E} [\cdot]$ is the expectation taken with respect to the posterior distribution given the current observations $D$ and their objective values $f(D)$ and constraint values $g(D)$.

\clearpage
\bibliographystyle{unsrt}

\end{document}